\documentclass[11pt]{article}
\usepackage[a-2u]{pdfx}  
\usepackage[margin=1in, top=1in]{geometry}
\usepackage{qpaper}
\usepackage[normalem]{ulem}
\usepackage{natbib}
\setcitestyle{numbers,square,citesep={;},aysep={,},yysep={;}}
\usepackage[hyperpageref]{backref}
\renewcommand*\backref[1]{\ifx#1\relax \else (Cited on page #1) \fi}
\hypersetup{
    colorlinks = true,
    citecolor = ForestGreen,
    linkcolor = RoyalBlue,
    urlcolor = ForestGreen
}
\usepackage{tikz}
\usetikzlibrary{arrows.meta,positioning,fit,calc,shapes.misc,backgrounds}
\usepackage{cleveref}
\usepackage{threeparttable}
\usepackage{multirow}
\usepackage{tabularx}
\usepackage{booktabs}
\usepackage[rgb]{xcolor}
\definecolor{fromHTML}{HTML}{EDDCCF}
\definecolor{fromRGB}{RGB}{237,220,207}
\usepackage{hyperref}
\usepackage{microtype}
\definecolor{bloatcolor}{RGB}{255, 220, 220}
\newcommand{\bloat}[1]{\sethlcolor{bloatcolor}\hl{#1}}
\newcommand{\bridge}[1]{\textbf{\textcolor{blue!60!black}{#1}}}

\usepackage{wrapfig}
\usepackage{soul}
\newcolumntype{Y}{>{\centering\arraybackslash}X}
\newcolumntype{Z}{>{\centering\arraybackslash}p{0.217\linewidth}}

\usepackage{amsmath,amsfonts,bm,amssymb}

\def\eqref#1{eq.~\ref{#1}}

\def\1{\bm{1}}

\def\vzero{{\bm{0}}}

\def\vs{{\bm{s}}}

\def\vx{{\bm{x}}}

\def\vz{{\bm{z}}}

\DeclareMathAlphabet{\mathsfit}{\encodingdefault}{\sfdefault}{m}{sl}
\SetMathAlphabet{\mathsfit}{bold}{\encodingdefault}{\sfdefault}{bx}{n}
\newcommand{\tens}[1]{\bm{\mathsfit{#1}}}

\def\tW{{\tens{W}}}

\usepackage{amsthm}
\usepackage{amssymb}

\theoremstyle{definition}

\theoremstyle{remark}

\def\Z{\mathbb{Z}}

\newcommand*{\prob}[1]{\mathbb{P}}

\usepackage{pifont}
\usepackage{bbding}
\usepackage{amssymb}
\usepackage{pdfrender}
\usepackage{authblk}
\usepackage{enumitem}
\DeclareMathOperator{\clip}{clip}
\newcommand{\round}[1]{\ensuremath{\left\lfloor{#1}\right\rceil}}

\DeclareMathAlphabet{\mathscompmodern}{OT1}{cmr}{bx}{n}
\SetMathAlphabet{\mathscompmodern}{bold}{OT1}{cmr}{bx}{n}
\newcommand{\transformation}[1]{\mathscompmodern{#1}}

\newcommand{\TT}{\transformation{T}}

\newcommand{\nf}[1]{{\small \textsf{#1}}}

\newcommand{\p}[1]{\left(#1\right)}

\definecolor{darkred}{HTML}{C00000}

\newcommand{\no}{\textbf{-}}
\newcommand{\na}{-}

\newcommand*{\boldcheckmark}{%
  \textpdfrender{
    TextRenderingMode=FillStroke,
    LineWidth=0.5pt, 
  }{\checkmark}%
}

\newcommand{\yes}{{\boldcheckmark}}
\definecolor{updcolor}{HTML}{d712e6}
\newcommand{\upd}[1]{{#1}}

\newcommand{\ourquant}{{$\text{FPTQuant}^{\circ}$}}
\newcommand{\boldourquant}{{$\textbf{FPTQuant}^{\boldsymbol{\circ}}$}}

\newcommand{\roleA}{Core contributors}
\newcommand{\roleB}{Contributors}
\newcommand{\roleC}{Project leads}
\newcommand*{\samethanks}[1][\value{footnote}]{\footnotemark[#1]}

\title{Efficient Reasoning on the Edge}

\author{
Yelysei Bondarenko\thanks{\roleA}
, Thomas Hehn\samethanks
, Rob Hesselink\samethanks, Romain Lepert\samethanks, Fabio Valerio Massoli\samethanks, Evgeny Mironov\samethanks, Leyla Mirvakhabova\samethanks, Tribhuvanesh Orekondy\samethanks, Spyridon Stasis\samethanks, Andrey Kuzmin\thanks{\roleB}, Anna Kuzina\samethanks, Markus Nagel\samethanks, Ankita Nayak\samethanks, Corrado Rainone\samethanks, Ork de Rooij\samethanks, Paul N Whatmough\samethanks, Arash Behboodi\thanks{\roleC}%
, Babak Ehteshami Bejnordi\samethanks
}

\setabstract{
Large language models (LLMs) with chain-of-thought reasoning achieve state-of-the-art performance across complex problem-solving tasks, but their verbose reasoning traces and large context requirements make them impractical for edge deployment. These challenges include high token generation costs, large KV-cache footprints, and inefficiencies when distilling reasoning capabilities into smaller models for mobile devices. Existing approaches often rely on distilling reasoning traces from larger models into smaller models, which are verbose and stylistically redundant, undesirable for on-device inference. In this work, we propose a lightweight approach to enable reasoning in small LLMs using LoRA adapters combined with supervised fine-tuning. We further introduce budget forcing via reinforcement learning on these adapters, significantly reducing response length with minimal accuracy loss. To address memory-bound decoding, we exploit parallel test-time scaling, improving accuracy at minor latency increase. Finally, we present a dynamic adapter-switching mechanism that activates reasoning only when needed and a KV-cache sharing strategy during prompt encoding, reducing time-to-first-token for on-device inference. Experiments on Qwen2.5-7B demonstrate that our method achieves efficient, accurate reasoning under strict resource constraints, making LLM reasoning practical for mobile scenarios. Videos demonstrating our solution running on mobile devices are available on our project page:\\ \url{https://qualcomm-ai-research.github.io/llm-reasoning-on-edge/}.
}

\begin{document}
\maketitlebis
\begingroup
  \renewcommand\thefootnote{}\footnotetext{Names within each group are sorted alphabetically.}%
\endgroup

\section{Introduction}

\begin{figure}[t]
    \centering
    \includegraphics[width=.95\linewidth]{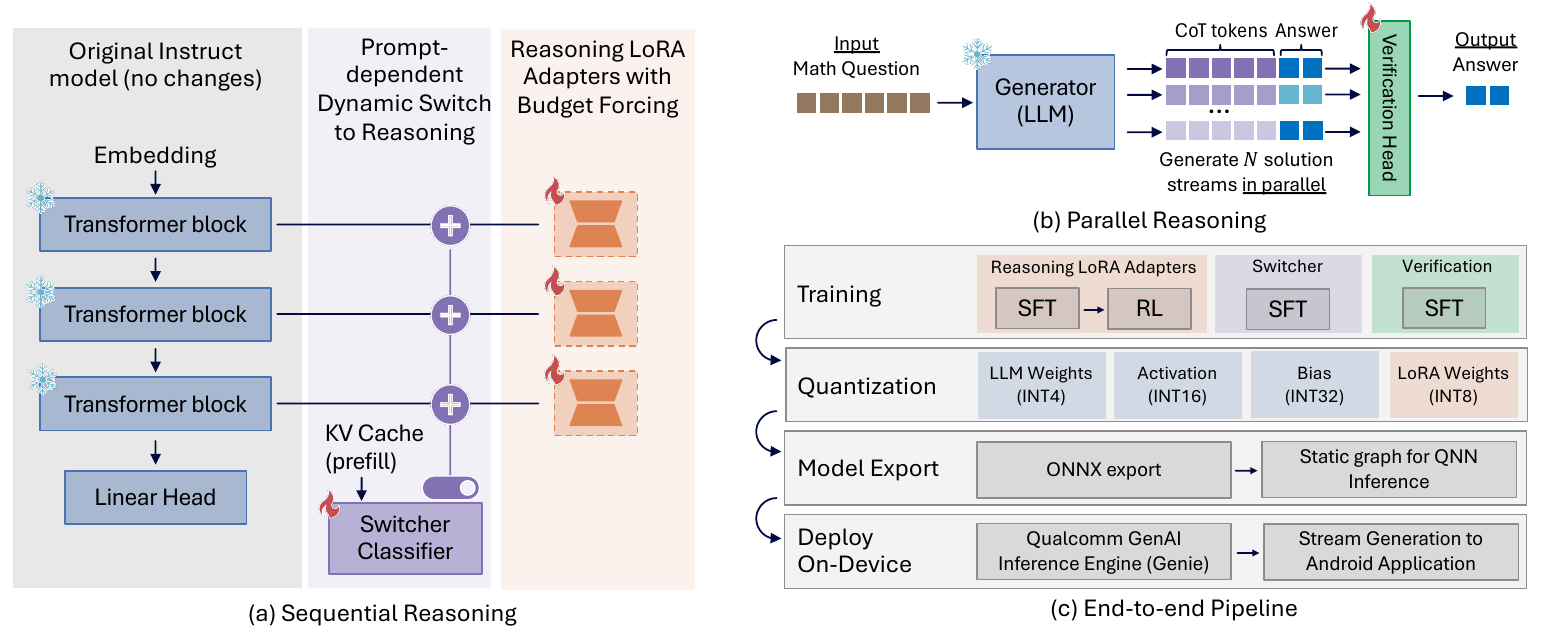}
    \caption{\textbf{Overview of the proposed efficient reasoning framework for edge devices.} (a) The model architecture utilizes parameter-efficient LoRA adapters and a lightweight switcher to dynamically route queries. This design allows the base model and the reasoning-activated mode to seamlessly share a reusable KV cache during prefill.
    (b) Parallel test-time scaling strategy, generating multiple reasoning streams concurrently to improve accuracy without severe latency penalties. 
    (c) The end-to-end deployment pipeline, illustrating the progression from multi-stage training (SFT and budget-forced RL) to quantization, model export, and final on-device execution.
    }
    \label{fig:overview}
\end{figure}

The success and impact of large language models (LLMs) continue to expand, with reasoning emerging as a fundamental component of their achievements. Commercial-grade coding agents reason about code structure, refactoring, debugging, and dependency resolution \cite{anthropic2024claude, jaech2024openai}. In scientific domains, LLMs are increasingly used to assist professional mathematicians with research level problems \cite{Abouzaid2026-first-proof, Cao2025-ai4math,Feng2026-ai4math}. On mobile devices, reasoning-capable LLMs unlock a new class of intelligent personal assistants able to plan multi-step tasks, respond contextually to user queries, and operate autonomously across apps and interfaces. Recent advances from Gemini and OpenAI demonstrate increasingly capable standalone reasoning \cite{Abouzaid2026-first-proof, Woodruff2026-gemini}, and a growing wave of agentic use cases is now emerging where models interact directly with device UIs and real-world services \cite{Yao2022-react,Zhou2025-maiui,Wang2025-uitars2,Venus-Team2026-uivenus1.5}. These achievements, however, come at the cost of generating massive numbers of tokens, with reasoning traces accounting for a large fraction of the overall computation \cite{thinking-efficiency2025}.

Deploying reasoning models on edge devices is attractive for mobile scenarios because it keeps sensitive data on-device, reduces round-trip latency, and remains available even under limited connectivity. In practice, however, on-device reasoning faces several key limitations. The first is the memory bottleneck: mobile devices, constrained by current DRAM capacities, can typically support smaller models with moderate quantization, and larger models only with more aggressive quantization schemes \cite{Alizadeh2024-llm-flash,Xiao2026-llm-mobile}. The second limitation is the cost of token generation in terms of power consumption, latency, and memory footprint. Long reasoning traces and large context lengths substantially increase KV cache size, quickly exhausting available memory. Finally, general purpose LLMs with broad capability scopes are difficult to realize within the model sizes supported by edge devices. While specialized small language models (SLMs) can match the performance of larger models on targeted tasks, model switching introduces additional memory movement overhead. Edge deployed models must therefore operate within strict memory budgets while achieving usable tokens per second (TPS) and acceptable time to response.

This work proposes an end-to-end pipeline for deploying reasoning-capable language models on edge devices under strict token, latency, and memory budgets. Our design starts from a base non-reasoning instruct model and enables a ``reasoning mode'' via LoRA adapters, so the same backbone can run either in standard chat mode (no adapters) or in reasoning mode (reasoning adapters enabled). A lightweight switcher routes each incoming query to the appropriate mode, enabling reasoning only when it is likely to help.

At training time, we use parameter-efficient fine-tuning to specialize the base model across domains while keeping deployment practical. We demonstrate that LoRA \cite{hulora} is effective for enabling domain targeted reasoning and task specialization. Selecting the LoRA rank as a function of the base model and desired performance is a central question, which we address through extensive analysis. Because LoRA adapters can be toggled at runtime, a single base model can be loaded once and then dynamically adapted to different tasks by enabling or disabling adapters. To enable KV-cache sharing between the base model and the LoRA augmented reasoning model, we propose masked LoRA training during the prefill phase, which we show has no significant impact on accuracy.

Our training recipe follows a two-stage structure commonly used for reasoning models: supervised fine-tuning (SFT) on high-quality reasoning traces, followed by a reinforcement learning (RL) phase for further alignment \cite{deepseekai2025deepseekr1}. Since, reasoning verbosity is not explicitly addressed during SFT and models often become verbose and repetitive after initial training, a key objective of our RL phase is to penalize excessively long reasoning traces via budget forcing \cite{muennighoff2025s1, li2025steering}. We apply budget forcing during RL and explore different training strategies, reward designs, prompting methods, and hard enforcement mechanisms, distilling best practices from our experiments.

At inference time, we target memory-bound decoding as a primary on-device bottleneck. We leverage parallel test-time scaling with neural verification to improve accuracy without incurring significant latency overhead, particularly in typical on-device settings. Concretely, because on-device inference splits into a compute-bound prefill phase and a memory-bound decoding phase, we can better utilize compute units (e.g., NPUs) by running parallel decoding paths with minimal incremental overhead. We further show that neural verification in an outcome reward model style can be implemented using a lightweight verifier head trained on the latent representations of the base model.

Together, these components allow us to start from a non-reasoning model and progressively harvest reasoning performance while maintaining deployability on edge devices. Figure \ref{fig:overview} provides a comprehensive overview of our proposed end-to-end framework, illustrating the complete progression from adapter training to on-device deployment.
We instantiate this pipeline on Qwen2.5 series of models, and we enable reasoning for these models using our proposed pipeline. On device deployment requires additional considerations. We discuss a range of quantization schemes supporting 4- to 8-bit weight quantization while allowing higher precision for activations to accommodate their dynamic range, minimizing quantization loss throughout the pipeline. The resulting quantized models are then exported and compiled for on-device execution. The entire workflow is implemented using Qualcomm open source tooling, including Qualcomm FastForward \cite{fastforward} and Qualcomm GENIE SDK \cite{GENIE}. In this paper, we provide deeper insights into the design choices and empirical analyses underpinning this work. We position the resulting system as a practical blueprint for deploying reasoning capable language models on resource constrained edge devices.

\section{Reasoning on Edge: System Design}
\definecolor{bg_color}{HTML}{F0F0F0}
\definecolor{orange_new}{HTML}{F0DBCD}
\definecolor{purple_new}{HTML}{BFB8D6}
\definecolor{blue_new}{HTML}{BAC6DD}
\begin{figure}[htbp]
    \centering
\begin{tikzpicture}[
  font=\sffamily,
  blockBase/.style={draw, rounded corners, thick, align=center, minimum width=2.0cm, minimum height=1.1cm},
    blockBlue/.style={blockBase, fill=purple_new!80},
  blockOrange/.style={blockBase, fill=orange!12},
  blockOrangeNew/.style={blockBase, fill=orange!20},
    blockGreen/.style={blockBase, fill=blue_new!60},
  blockPurple/.style={blockBase, fill=violet!12},
  data/.style={draw, thick, align=center, minimum width=2.2cm, minimum height=0.9cm, fill=gray!10},
  arrow/.style={-Latex, very thick},
  dashedarrow/.style={-Latex, very thick, dashed},
  group/.style={-Latex, rounded corners, dashed, inner sep=8pt, fill=bg_color, draw=none},
  note/.style={align=left, font=\footnotesize}
  every node/.style={fill opacity=1, draw opacity=1}
]

\node[blockBlue] (base) {Base LLM};

\node[blockOrangeNew, right=2cm of base] (lora) {Reasoning LoRA};
\draw[arrow] (base) -- node[midway, above] {SFT}(lora);

\coordinate (sftA) at ($(base)!0.35!(lora)$); 
\coordinate (sftB) at ($(base)!0.8!(lora)$); 
\begin{pgfonlayer}{background}
  \node[group, label={[font=\bfseries,xshift=-1.5cm]above:Section~\ref{sec:lora}}] (SFTbox)
    [fit=(lora) (sftA) (sftB)] {};
\end{pgfonlayer}

\node[blockOrangeNew, right=5cm of lora] (rl) {Efficient \\ Reasoning LoRA};
\draw[arrow] (lora)  -- node[pos=0.5, above, xshift=0.5cm] {RL + Budget Forcing} (rl);
\coordinate (rlA) at ($(lora)!0.35!(rl)$);  
\coordinate (rlB) at ($(lora)!0.8!(rl)$);  
\begin{pgfonlayer}{background}
  \node[group, label={[font=\bfseries,xshift=-2.5cm]above:Section~\ref{sec:budget_forcing}}] (RLbox)
    [fit=(rl) (rlA) (rlB)] {};
\end{pgfonlayer}

\node[blockGreen, below=1.2cm of rl] (final) {Hybrid \\ reasoning model};
\draw[dashedarrow] (rl) -- (final);

\node[blockOrangeNew, below=1.2cm  of base] (switcher) {Switcher Model\\\footnotesize reasoning-needed classifier};
\draw[arrow] (base) -- (switcher);
\coordinate (sA) at ($(base)!0.6!(switcher)$);
\begin{pgfonlayer}{background}
  \node[group, label={[font=\bfseries,xshift=-1.1cm]above:Section~\ref{sec:switcher}}] (Switchbox)
    [fit=(sA)(switcher)] {};
\end{pgfonlayer}
\draw[dashedarrow] (switcher) -- (final);

\end{tikzpicture}

\caption{\textbf{Architecture of the Hybrid Reasoning Model.} The pipeline begins with a compact base LLM, which is specialized for reasoning via LoRA-based supervised fine-tuning (SFT). To enforce concise generation and prevent excessive verbosity, these adapters undergo reinforcement learning (RL) with Budget Forcing. Finally, a lightweight Switcher module is introduced to act as a reasoning-needed classifier, creating a hybrid model that dynamically routes incoming queries to either the fast base model or the specialized reasoning adapters based on task complexity.}

\label{fig:system_design}
\end{figure}
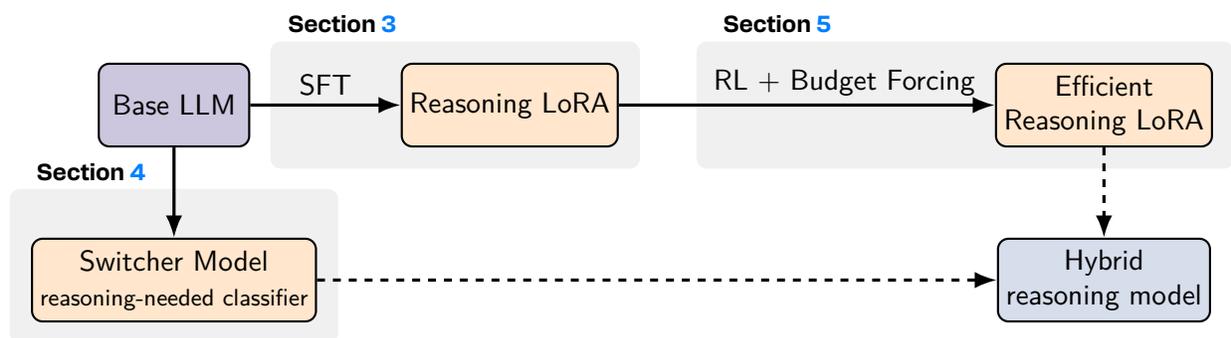

Developing compact yet capable reasoning models requires a training pipeline that can reliably elicit high-quality chain-of-thought (CoT) behavior from a relatively small base LLM while avoiding unnecessary verbosity and excessive computation. Although pretrained LLMs can sometimes function as zero-shot reasoners \citep{kojima2022large}, explicit reasoning via approaches such as scratchpads \citep{nye2022show} or supervised CoT \citep{wei2022chain} remains essential for strong performance on mathematics and coding tasks. Recent systems such as OpenAI O1 \citep{jaech2024openai} and specialized small reasoning models, including Tina \citep{wang2025tina}, Phi-4-mini \citep{xu2025phi}, and hybrid reasoning architectures \citep{jiang2025think}, illustrate that such capabilities can be distilled into relatively small models when combined with targeted fine-tuning and alignment techniques.

Our objective is to fine-tune a modest base LLM so that it performs competitively on complex reasoning tasks while producing concise outputs and minimizing token generation. To keep adaptation lightweight and deployable, we perform training primarily through Low-Rank Adapters (LoRA) \citep{hulora}, which preserve a reusable frozen backbone and enable modular reasoning specialization. The overall pipeline, presented in Figure~\ref{fig:system_design}, integrates supervised fine-tuning, reinforcement learning, and lightweight routing in a manner that preserves efficiency and supports deployment under strict memory and latency constraints.

\textbf{Supervised fine-tuning} constitutes the first stage of our pipeline and is designed to unlock the reasoning capabilities of the pretrained base LLM.  Rather than performing dense fine-tuning, we adopt parameter-efficient fine-tuning using LoRA. LoRA has been shown to match or even surpass dense fine-tuning in reasoning settings \citep{schulman2025lora}, while enabling the base model to remain frozen and reusable for multiple domains. This stage equips the model with the fundamental ability to reason through multi-step problems, but, as widely observed, also increases verbosity and can lead to unnecessarily long or repetitive traces \cite{deepseekai2025deepseekr1}.

To refine the reasoning behavior and control verbosity, we apply \textbf{Reinforcement Learning}~\citep{chen2025acereason,dang2025reinforcement} (RL) using a custom reward function tailored to two objectives: accuracy and efficiency. First, we use budget forcing~\citep{alomrani2025reasoning}, a mechanism that penalizes excessively long responses. This constraint encourages the model to produce concise reasoning traces without sacrificing correctness. Second, we incorporate an answer-based reward, which directly incentivizes the model to generate correct final answers.  For optimization, we employ the group-based relative policy optimization (GRPO) algorithm~\citep{shao2024deepseekmath}, which updates the LoRA parameters.

Finally, not all  user queries require multi-step reasoning. To avoid unnecessary computation, we introduce a lightweight \textbf{Switcher module} that predicts whether reasoning is needed based on hidden prompt representations. When reasoning is unnecessary, the model bypasses the LoRA adapters and relies on the base model directly, reducing latency and limiting KV cache growth, an important consideration for edge deployment.

In the following sections, we decompose our end-to-end pipeline into its main components, LoRA-based adaptation (Section~\ref{sec:lora}), dynamic inference-time routing using a Switcher module to activate or bypass these adapters (Section~\ref{sec:switcher}), budget-forced RL for verbosity control (Section~\ref{sec:budget_forcing}), and parallel test-time scaling (Section~\ref{sec:parallel}), and the deployment path including quantization and on-device execution (Section~\ref{sec:quantization}).  In each section, we describe the component and then quantify its impact with targeted experiments, highlighting key insights and focused ablations to isolate which design choices drive accuracy and which improve on-device efficiency.

\section{LoRA for Modular Reasoning}\label{sec:lora}

In our reasoning framework, we adopt parameter-efficient fine-tuning (PEFT) for two reasons: first, it enables scalable experimentation at low training costs, and second, it produces modular adapters that can be enabled or disabled at runtime, allowing a single base model to switch between general-purpose chat and enhanced reasoning modes. To elicit reasoning behavior, we perform SFT on datasets composed of reasoning traces generated by stronger teacher models such as DeepSeek-R1 \cite{deepseekai2025deepseekr1} and QwQ-32B \cite{QwQ-32B}.

Our experiments show that 3B and 7B models can acquire strong reasoning ability using straightforward SFT on curated trace datasets in a cost-efficient setup, reaching performance comparable to substantially larger distilled baselines (e.g., DeepSeek-R1-Distill-Qwen-7B). These results suggest that strong reasoning does not require heavy distillation pipelines or large training budgets; high-quality trace data (e.g., OpenThoughts3 \cite{guha2025openthoughts}) combined with lightweight fine-tuning is sufficient to close most of the gap. Overall, this provides a practical and scalable path to building capable reasoning models without expensive infrastructure.

\subsection{Experimental Setup}

The primary goal of our LoRA adaptation stage is to determine if small, general-purpose instruct models can acquire expert-level reasoning capabilities without full-parameter distillation. Accordingly, we adopt the Qwen2.5-3B-Instruct and Qwen2.5-7B-Instruct models \cite{qwen2025qwen25technicalreport} as our core experimental backbones. In this section, we outline the datasets and optimization strategies used to elicit reasoning behavior via LoRA, as well as the diverse suite of math, science, and coding benchmarks used to rigorously evaluate the resulting performance trade-offs. We additionally provide an extensive LoRA hyperparameter study designed to identify the most stable and compute-efficient adaptation strategies for both the 3B and 7B backbones.

\subsection{Training Details}

\paragraph{Data.}
In our experiments, we utilize two SFT datasets. The first is \textbf{Mixture of Thoughts} (MoT) \cite{openr1}, which contains 350k reasoning traces distilled from the DeepSeek-R1 model. This dataset covers three core domains: Math (93.7k traces), Code (83.1k traces), and Science (173k traces).
The second dataset is \textbf{OpenThoughts3-1.2M} (OT3) \cite{guha2025openthoughts}, comprising 850k Math questions, 250k Code questions, and 100k Science questions. The annotation traces for OT3 were generated using the QwQ-32B model.

\paragraph{Training configuration.}

All models were trained for 5 epochs using the bfloat16 data type with the DeepSpeed zero2 configuration and CPU offloading enabled. Across all configurations, we applied a cosine learning rate schedule with a warmup ratio of 0.1, and weight decay was set to 0.

For the baseline dense training on the MoT dataset, the learning rate was set to $1\mbox{e$-$}5$ for Qwen2.5-3B-Instruct and Qwen2.5-7B-Instruct models, with a global batch size of 128. Model weights were optimized using the AdamW optimizer with $(\beta_1,\beta_2) = (0.9, 0.95)$. For dense training on the OT3 dataset, we followed the recipe described in \cite{guha2025openthoughts}. Dense models were trained with a learning rate of $8\mbox{e$-$}5$ and a batch size of 512, using AdamW with $(\beta_1,\beta_2) = (0.9, 0.999)$.

In our LoRA training setup, we follow the common practice of employing relatively larger learning rates. We also find that reducing the batch size generally leads to more stable optimization and improved results. Throughout all PEFT experiments, we use a LoRA rank of $128$, set the LoRA alpha to twice this value, adopt a learning rate of $2\mbox{e$-$}{4}$, and train with a batch size of $64$.

\subsection{Evaluation Details}
\paragraph{Benchmarks.}

We assess the reasoning capabilities of our trained models using a diverse set of benchmarks that span tasks across mathematics, science, and coding domains. To evaluate multi-step mathematical problem-solving, we use challenging competition datasets including AIME 24/25~\cite{aime}, AMC23~\cite{amc23}, and MATH500~\cite{hendrycks2021measuring}. Scientific reasoning is measured using the PhD-level GPQA Diamond dataset~\cite{rein2024gpqa}. Finally, to evaluate code generation, we utilize LiveCodeBench (v2, code generation scenario only)~\cite{livecodebench} for recent competitive programming problems, alongside standard Python programming tasks from HumanEval~\cite{humaneval} and MBPP~\cite{mbpp}, including their rigorously verified EvalPlus variants (HumanEval+ and MBPP+)~\cite{evalplus}. Comprehensive descriptions of each benchmark, including problem counts and specific testing scenarios, are provided in Appendix~\ref{App:benchmark}.

\vspace{-5mm}
\paragraph{Evaluation pipeline.}

Our evaluation pipeline is structured as follows. For reasoning benchmarks including AIME24, AIME25, MATH500, GPQA, and AMC, we allow a generation length of up to 32,768 tokens, with the temperature set to 0.6 and $\mathrm{top}_{\_}\mathrm{p}$ to 0.95. For models trained on the OT3 dataset, we adopt the generation parameters recommended by the authors in \cite{guha2025openthoughts}, specifically setting the temperature to 0.7 and $\mathrm{top}_{\_}\mathrm{p}$ to 1.0 and keeping generation length up to 32,768 tokens. Also for LCB, MBPP and HumanEval (and their + counterparts) we set the generation parameters as recommended by the models' creators; we set the maximal generation length to 32768 tokens for LCB, and 1024 tokens for HumanEval and MBPP.

Given that some benchmarks contain a limited number of questions and are therefore more susceptible to accuracy variance, we perform multiple evaluation runs to ensure robustness similarly to the protocol outlined in \cite{guha2025openthoughts}. Specifically, we evaluate the AIME24, AIME25 and AMC datasets 10 times and report the averaged results. For GPQA, we conduct 4 evaluation runs and for MATH500 we run the evaluation once. For coding benchkmarks, the pass@1 score is estimated from a pool of 16 candidate solutions for LCB, and of 200 candidate solutions for HumanEval and MBPP, using the unbiased pass@$k$ estimator first proposed in \cite{humaneval}.
All evaluations are conducted using the lighteval framework \cite{lighteval} with vLLM support \cite{kwon2023efficient}, except for those on HumanEval, MBPP and their enhanced variants, for which we used the Evalplus package\footnote{\url{https://github.com/evalplus/evalplus}}.

\begin{table}[h]
\caption{Results of the main experiments. FT data indicates the finetuning dataset. The LoRA column specifies the adapter rank, or its absence for dense finetuning. LCB refers to LiveCodeBench, and HE denotes HumanEval. $\mathrm{R1}^*$ corresponds to the R1-Distill-Qwen-7B model released by DeepSeek; we evaluated this model following our standardized evaluation protocol.}
\label{tab:lora_main_results}
\vskip 0.15in
\centering
\tabcolsep=0.11cm
\scalebox{0.95}{
\begin{tabular}{l|cc|cccccccccc}

\hline
& \textbf{FT data} & \textbf{LoRA} & \textbf{AIME24} & \textbf{AIME25} & \textbf{MATH500} & \textbf{GPQA} &
\textbf{AMC23} & \textbf{LCB} & \textbf{HE} & \textbf{HE+} & \textbf{MBPP} & \textbf{MBPP+} \\
\hline

\multirow{4}{*}{\rotatebox{90}{\textbf{Qwen-3B}}}
& - & - & 0.09 & 0.03 & 0.66 & 0.31 & 0.36 & 0.23 & 0.40 & 0.64 & 0.72 & 0.61 \\
& MoT & - &  0.20 & 0.22 & 0.80 & 0.37 & 0.61 & 0.38 & 0.43 & 0.38 & 0.34 & 0.29 \\
& OT3 & - & 0.51 & 0.43 & 0.92 & 0.40 & 0.83 & 0.50 & 0.48 & 0.43 & 0.36 & 0.30 \\
& OT3 & rk 128  &  0.18 & 0.21 & 0.79 & 0.32 & 0.55 & 0.23 & 0.40 & 0.35 & 0.39 & 0.32 \\

\hline

\multirow{7}{*}{\rotatebox{90}{\textbf{Qwen-7B}}}
& - & - & 0.10 & 0.17 & 0.76 & 0.37 & 0.60 & 0.36 & 0.83 & 0.75 & 0.80 & 0.68 \\
& $\mathrm{R1}^{*}$ & - &  0.55 & 0.40 & 0.92 & 0.49 & 0.89 & 0.59 & 0.41 & 0.38 & 0.42 & 0.36 \\
& MoT & - & 0.37 & 0.28 & 0.90 & 0.48 & 0.78 & 0.55 & 0.53 & 0.47 & 0.60 & 0.50 \\
& OT3 & - & 0.61 & 0.54 & 0.95 & 0.47 & 0.89 & 0.66 & 0.58 & 0.51 & 0.56 & 0.46 \\
& OT3 & rk 64  &  0.47 & 0.39 & 0.92 & 0.45 & 0.78 & 0.48 & 0.60 & 0.54 & 0.63 & 0.52 \\
& OT3 & rk 128  & 0.56 & 0.38 & 0.93 & 0.43 & 0.82 & 0.54 & 0.60 & 0.54 & 0.57 & 0.46 \\ 
& OT3 + MoT & rk 128  & 0.44 & 0.39 & 0.93 & 0.48 & 0.83 & 0.51 & 0.59 & 0.52 & 0.60 & 0.49 \\ \hline 
\end{tabular}}
\end{table}

\subsection{Results}
As shown in \Cref{tab:lora_main_results}, we evaluate Qwen2.5-3B-Instruct and Qwen2.5-7B-Instruct models under dense and LoRA-based fine-tuning, and compare against key baselines. Finetuning on the OpenThoughts3 (OT3) dataset yields the largest and most consistent gains in reasoning performance across both backbones, improving accuracy substantially on math and science benchmarks and also boosting performance on the more reasoning-sensitive coding benchmark (LiveCodeBench). In contrast, Mixture of Thoughts (MoT) provides clear improvements over the base models, but its gains are consistently smaller than OT3. In addition, the densely trained Qwen2.5-3B model on OT3 performs on par with, or slightly above, the densely trained Qwen2.5-7B model on MoT, suggesting that higher data quality can partially compensate for smaller backbone size.

For Qwen2.5-7B, LoRA fine-tuning on OT3 recovers most of the dense OT3 improvements, and increasing the adapter rank from 64 to 128 generally narrows the gap on core reasoning benchmarks (e.g., AIME24 and LCB).  Notably, OT3 with LoRA rank 128, which requires updating only 4.24\% of the parameters compared to dense fine-tuning, reaches performance close to the R1-Distill-Qwen-7B baseline on several reasoning benchmarks, indicating that lightweight adapter training can recover much of a distilled model’s capability at significantly lower adaptation cost.  For Qwen2.5-3B, however, OT3 LoRA (rank 128) underperforms the dense OT3 model by a large margin across reasoning benchmarks, highlighting that adapter capacity and/or optimization details matter more at smaller scales and motivating our ablations in the subsequent sections.

The coding results reveal a trade-off between reasoning specialization and ``direct-answer'' code generation. While SFT consistently improves performance on LCB, it always results in some degradation for MBPP, HumanEval and their respective .+ variants. 
This pattern is consistent with a shift toward explicit multi-step reasoning: it helps on harder, reasoning-sensitive coding tasks like LCB, but can be counterproductive on benchmarks that reward short, direct code outputs. 
Concretely, our SFT stage is designed to elicit reasoning behavior into an otherwise non-reasoning model, and LCB is a setting where such explicit reasoning can be beneficial. 
In contrast, MBPP and HumanEval typically do not prompt the model to reason, and a response must be given directly\footnote{We have nonetheless confirmed that the model's answers are never cut short by the Evalplus harness.}. Interestingly, for Qwen2.5-7B, OT3 LoRA (rank 64/128) often retains stronger HumanEval/MBPP performance than dense OT3, suggesting that PEFT can partially mitigate the specialization/forgetting trade-off relative to dense fine-tuning. Prior work has similarly observed~\cite{huan2025does} that SFT on reasoning traces can result in partial forgetting of general capabilities, and that his phenomenon could potentially be mitigated by employing RL fine-tuning~\cite{lai2025reinforcement}.

We also explored a two-stage training strategy for the 7B model to see if we could combine the distinct benefits of both datasets. Our motivation was that while the OT3 dataset (generated by the smaller QwQ model) provided the strongest baseline performance, the MoT dataset might contain complementary, highly complex reasoning patterns that the model could absorb in a subsequent training phase. However, interestingly, additional training on MoT following training on OT3 results in minimal changes across benchmarks. Accuracies remain largely stable, except for a 0.12-point decrease on AIME24 and a 0.05-point improvement on GPQA.

\subsubsection{LoRA Hyperparameter Study: Rank, Learning Rate, and Batch Size}\label{sec:ablations}
To study the effect of hyperparameters on the performance of PEFT models, we consider a range of values for learning rate, batch size, and LoRA rank. We trained Qwen2.5-3B-Instruct and Qwen2.5-7B-Instruct on a subset of OT3 consisting of 50{,}000 entries for 1 epoch. For each model, we vary the values within the following ranges: learning rate in $\{1\mbox{e$-$}4, 2\mbox{e$-$}4, 5\mbox{e$-$}4\}$, batch size in $\{32, 64, 128\}$, and LoRA rank in $\{32, 64, 128, 256\}$. LoRA adapters were applied to all linear layers, with $\alpha$ set to 2$\times \text{rank}$ and dropout fixed at $0.1$. For our ablation study, we evaluate the trained models on MATH and Science benchmarks including AIME24, AIME25, MATH500, GPQA Diamond and AMC23. Similarly to the setup used in \cite{guha2025openthoughts}, we average the accuracy values over 10 runs for smaller datasets like AIME24, AIME25 and AMC23, over 4 runs for GPQA and once for MATH500. We set the temperature to 0.7 and $top\_p$ value to 1.0.
Full evaluation results for this ablation study can be found in Appendix \ref{sec:lora_ablation}.

\paragraph{Qwen2.5-3B-Instruct results.} The full performance of the models trained on a set of considered hyperparameters is summarized in \Cref{tab:ablations_qwen3b} in the Appendix. To isolate the effect of each hyperparameter, we summarize results by averaging accuracy over the other two dimensions (learning rate, batch size, and LoRA rank), highlighting the main trends. As shown in \Cref{tab:lr_summary}, learning rate has a noticeable impact on performance, with $2\mbox{e$-$}4$ providing the highest overall average accuracy across tasks. We also observe opposite sensitivities across benchmarks: AIME24 and AIME25 improve as the learning rate increases, whereas MATH500 degrades, suggesting that larger learning rates can lead to over-adaptation on more complex mathematical reasoning.

\begin{table}[ht]
\centering
\caption{Average performance grouped by learning rate for Qwen2.5-3B-Instruct.}
\vskip 0.15in

\label{tab:lr_summary}
\begin{tabular}{lcccccc}
\hline
\textbf{LR} & \textbf{AIME24} & \textbf{AIME25} & \textbf{MATH500} & \textbf{GPQA} & \textbf{AMC23} & \textbf{Avg} \\
\hline
$1\mbox{e$-$}4$ & 0.040 & 0.025 & 0.573 & 0.277 & 0.281 & 0.239 \\
$2\mbox{e$-$}4$ & \textbf{0.051} & 0.033 & \textbf{0.574} & 0.274 & \textbf{0.299} & \textbf{0.246} \\
$5\mbox{e$-$}4$ & \textbf{0.051} & \textbf{0.038} & 0.556 & \textbf{0.280} & 0.275 & 0.240 \\
\hline
\end{tabular}
\end{table}

As presented in \Cref{tab:rank_summary}, LoRA rank shows the most consistent positive influence on aggregate performance. Increasing rank generally improves accuracy, with rank $256$ achieving the highest overall average. However, rank $128$ already performs strongly on most benchmarks, making it a practical operating point when adapter memory is constrained. Across tasks, AIME25 is particularly sensitive to rank, while MATH500 remains comparatively stable. Overall, higher ranks are preferable when resources permit, but rank $128$ offers a favorable accuracy--efficiency trade-off for edge deployment.

\begin{table}[ht]
\centering
\caption{Average performance grouped by LoRA rank for Qwen2.5-3B-Instruct. \%TP denotes the percentage of trainable parameters.}
\vskip 0.15in
\label{tab:rank_summary}
\begin{tabular}{lccccccc}
\hline
\textbf{Rank} & \textbf{\%TP} & \textbf{AIME24} & \textbf{AIME25} & \textbf{MATH500} & \textbf{GPQA} & \textbf{AMC23} & \textbf{Avg} \\
\hline
32  & 1.94\% & 0.047 & 0.020 & 0.568 & 0.270 & 0.284 & 0.238 \\
64  & 3.88\% & 0.040 & 0.023 & 0.570 & \textbf{0.283} & 0.284 & 0.240 \\
128 & 7.76\% & \textbf{0.054} & 0.038 & 0.560 & 0.272 & \textbf{0.287} & 0.242 \\
256 & 15.52\% & 0.048 & \textbf{0.048} & \textbf{0.573} & 0.282 & 0.284 & \textbf{0.247} \\
\hline
\end{tabular}
\end{table}

Finally, batch size had a relatively minor effect compared to learning rate and rank as shown in \Cref{tab:bs_summary}. The best overall performance was observed at batch size set to 64, though the differences were small. MATH500 benefited slightly from larger batches like 128, while AIME25 peaked at 64. These results indicate that batch size can be chosen primarily based on computational constraints without significant impact on overall performance.

\begin{table}[ht]
\centering
\caption{Average performance grouped by batch size for Qwen2.5-3B-Instruct.}
\vskip 0.15in
\label{tab:bs_summary}
\begin{tabular}{lcccccc}
\hline
\textbf{Batch size} & \textbf{AIME24} & \textbf{AIME25} & \textbf{MATH500} & \textbf{GPQA} & \textbf{AMC23} & \textbf{Avg} \\
\hline
32  & \textbf{0.054} & 0.030 & 0.562 & 0.277 & \textbf{0.288} & 0.242 \\
64  & 0.048 & \textbf{0.038} & 0.566 & \textbf{0.278} & 0.287 & \textbf{0.243} \\
128 & 0.040 & 0.028 & \textbf{0.575} & 0.277 & 0.281 & 0.240 \\
\hline
\end{tabular}
\end{table}

\paragraph{Qwen2.5-7B-Instruct results.} \Cref{tab:qwen7_ablations} in the Appendix summarizes the Qwen2.5-7B-Instruct LoRA ablations over learning rate, batch size, and adapter rank. In contrast to the 3B setting, where learning rate affects performance but the averages vary only modestly across the tested values, the 7B backbone exhibits a narrower stable learning-rate range (Table~\ref{tab:lr_summary_second_model_full}): LR=$1\mbox{e$-$}4$ and $2\mbox{e$-$}4$ trains reliably, whereas LR=$5\mbox{e$-$}4$ is often unstable and can lead to collapsed runs. Overall, a practical guideline is to use a lower learning rate for stable training on larger backbones, and tune the remaining hyperparameters within that stable regime.
For the individual tables presented below, we account for this distinction explicitly. The table reporting averages across different learning rate settings includes all runs, highlighting the instability introduced by higher learning rates. In contrast, the tables reporting averaged accuracies for batch size and LoRA rank exclude runs with a learning rate of $5\mbox{e$-$}4$, as those configurations contain diverged results.

\begin{table}[ht]
\centering
\caption{Average performance grouped by learning rate for Qwen2.5-7B-Instruct.}
\vskip 0.15in

\label{tab:lr_summary_second_model_full}
\begin{tabular}{lcccccc}
\hline
\textbf{LR} & \textbf{AIME24} & \textbf{AIME25} & \textbf{MATH500} & \textbf{GPQA} & \textbf{AMC23} & \textbf{Avg} \\
\hline
$1\mbox{e$-$}4$ & 0.158 & 0.132 & 0.775 & \textbf{0.366} & 0.533 & 0.393 \\
$2\mbox{e$-$}4$ & \textbf{0.170} & \textbf{0.148} & \textbf{0.779} & 0.354 & \textbf{0.543} & \textbf{0.399} \\
$5\mbox{e$-$}4$ & 0.148 & 0.112 & 0.635 & 0.350 & 0.439 & 0.337 \\
\hline
\end{tabular}
\end{table}

As shown in Table~\ref{tab:rank_summary_second_model_full}, LoRA rank has a measurable but relatively small impact on Qwen2.5-7B performance: the average score improves from 0.388 (rank 32) to 0.402 (rank 128), while rank 256 is comparable at 0.397.
Overall, ranks 64--128 form a tight trade-off region, with rank 128 best on average but only marginally better than lower ranks. Compared to the 3B model, the 7B results are more tightly clustered across ranks, indicating that rank is a weaker lever here than it is for smaller backbones.

\begin{table}[ht]
\centering
\caption{Average performance grouped by LoRA rank for Qwen2.5-7B-Instruct. \%TP denotes the percentage of trainable parameters.}
\vskip 0.15in

\label{tab:rank_summary_second_model_full}
\begin{tabular}{lccccccc}
\hline
\textbf{Rank} &\textbf{\%TP} & \textbf{AIME24} & \textbf{AIME25} & \textbf{MATH500} & \textbf{GPQA} & \textbf{AMC23} & \textbf{Avg} \\
\hline
32  & 1.06\% & 0.152 & 0.119 & 0.776 & 0.357 & 0.534 & 0.388 \\
64  & 2.12\% & 0.159 & 0.150 & 0.779 & 0.355 & 0.535 & 0.396 \\
128 & 4.24\% & \textbf{0.178} & 0.141 & \textbf{0.784} & \textbf{0.367} & \textbf{0.541} & \textbf{0.402} \\
256 & 8.48\% & 0.165 & \textbf{0.152} & 0.771 & 0.359 & 0.536 & 0.397 \\
\hline
\end{tabular}
\end{table}

Similarly, Table~\ref{tab:batch_summary_second_model_full} shows that batch size has a negligible effect on Qwen2.5-7B performance in our sweep. The average accuracy varies only from 0.394 to 0.398. This mirrors the 3B trend, suggesting batch size can largely be chosen based on computational constraints once learning rate is in a stable regime.

\begin{table}[ht]
\centering
\caption{Average performance grouped by batch size for Qwen2.5-7B-Instruct.}
\vskip 0.15in

\label{tab:batch_summary_second_model_full}
\begin{tabular}{lcccccc}
\hline
\textbf{Batch size} & \textbf{AIME24} & \textbf{AIME25} & \textbf{MATH500} & \textbf{GPQA} & \textbf{AMC23} & \textbf{Avg} \\
\hline
32  & 0.154 & \textbf{0.148} & \textbf{0.781} & 0.355 & 0.530 & 0.394 \\
64  & 0.164 & 0.146 & 0.774 & \textbf{0.366} & 0.538 & \textbf{0.398} \\
128 & \textbf{0.172} & 0.127 & 0.776 & 0.358 & \textbf{0.541} & 0.395 \\
\hline
\end{tabular}
\end{table}

As a result of the hyperparameter study, it follows that the setup with learning rate value of $2\mbox{e$-$}4$, batch size of 64, and LoRA rank set to 128 results in consistently good results while providing a good compromise between efficiency and training stability.

\section{Dynamic LoRA Routing via the Switcher Module}\label{sec:switcher}

While reasoning models excel at complex problem-solving, not every user query requires an exhaustive, multi-step CoT. For standard conversational prompts or straightforward factual questions, generating long reasoning traces incurs unnecessary latency and computational overhead which can be a critical bottleneck for edge devices. To address this, we introduce a lightweight \textit{Switcher} module that enables dynamic adapter routing. By analyzing the user prompt, the switcher decides whether to bypass or activate the reasoning-specific LoRA adapters. When disabled, the system operates as the highly efficient original instruct model for regular conversation; when activated, it seamlessly transitions into a specialized reasoning engine.

Architecturally, the switcher serves as an auxiliary classification head on top of the base LLM and operates during the prefilling stage. It processes the hidden states from the final transformer layer and computes an averaged sequence representation. Based on this representation, the switcher performs binary classification to determine whether the input sequence corresponds to a reasoning-oriented task. If classified as such, the reasoning LoRA adapters are activated for subsequent decoding.
 
The switcher head is implemented as a lightweight multilayer perceptron (MLP) with a single hidden dimension of $8$, a ReLU activation function, and a dropout rate of $p = 0.2$. This compact architecture ensures negligible overhead for efficient on-device inference while maintaining sufficient expressive capacity for sequence-level classification.
 
On edge devices, the prefill phase is heavily compute-bound. Processing a long input prompt in a single pass can incur prohibitive computational overhead. To mitigate this, practical on-device implementations typically divide the prefill sequence into smaller, discrete chunks. The switcher module is explicitly designed to support this chunked prefill strategy. Rather than buffering the hidden states of the entire prompt to compute a global average, the switcher updates its sequence representation on the fly. Specifically, we compute a running exponential moving average of the hidden states across these chunks. In our setup, we use a chunk size of $128$ tokens with a smoothing coefficient of $\alpha = 0.5$. To enhance robustness to quantization artifacts, we inject independent Gaussian noise with zero mean and standard deviation $\sigma = 0.5$ into the averaged representation during training.

\paragraph{Masked LoRA training for KV-cache reuse.}  A major challenge with dynamically activating LoRA adapters at inference time is KV-cache compatibility. Under standard LoRA training, the model expects the KV cache for the prompt tokens (the prefill phase) to be generated with the LoRA adapters fully active. If a query is routed to the reasoning mode after the base model has already encoded the prompt, the system would typically need to re-encode the entire prompt with the LoRA adapters activated to generate a compatible KV cache. On edge devices, this re-encoding incurs a severe latency and compute penalty. To eliminate this inefficiency, we introduce a \textit{masked LoRA} training strategy. During the fine-tuning of the reasoning adapters, we mask (disable) the LoRA weights during the forward pass of the prompt tokens, activating them only for the generation of the response tokens. This forces the LoRA adapters to adapt to the prompt KV cache generated strictly by the base model. Empirically, we observe that this strategy incurs no drop in reasoning accuracy, while allowing the base model and reasoning mode to seamlessly share a single prefill KV cache, entirely obviating the need to re-encode prompt tokens when switching.

\subsection{Training details}
To train the switcher, we constructed a small dataset that combines both straightforward conversational or knowledge based queries and complex queries, enabling the model to learn how to distinguish when reasoning is required. The dataset is structured as follows: each entry consists of a question prompt paired with a label 0 or 1, indicating low or high complexity, respectively. Complexity is determined based on the source of the dataset from which the prompt was sampled. We included prompts from datasets covering math and non-math domains to reduce the risk that the switcher overfits to domain-specific cues (e.g., assuming all math questions are inherently more difficult).

The final dataset contains approximately 2k samples. Easy queries were randomly drawn from the SQuAD2.0 dataset (600 questions) \cite{rajpurkar2018knowdontknowunanswerable}, which consists primarily of general-knowledge comprehension questions, and from the MMLU math subset (419 questions) \cite{hendrycks2021measuring}, which includes straightforward mathematical problems. Hard prompts were sourced from a subset of the S1K dataset (500 questions) \cite{muennighoff2025s1}, encompassing challenging questions spanning math, science, and crosswords, as well as from StrategyQA (500 questions) \cite{geva2021didaristotleuselaptop} covering reasoning questions from non-scientific domains.

\subsection{Results}

The primary motivation for the switcher module is to optimize standard, day-to-day user interactions. In real-world edge deployments, the vast majority of user queries are simple conversations, factual lookups, or basic instructions that do not require multi-step reasoning. By aggressively thresholding the switcher to route these simple queries to the base instruct model, we can achieve massive aggregate savings in token generation, latency, and power consumption, reserving the LoRA reasoning adapters strictly for complex tasks.

\begin{figure}[!ht]
    \centering
    \includegraphics[width=\linewidth]{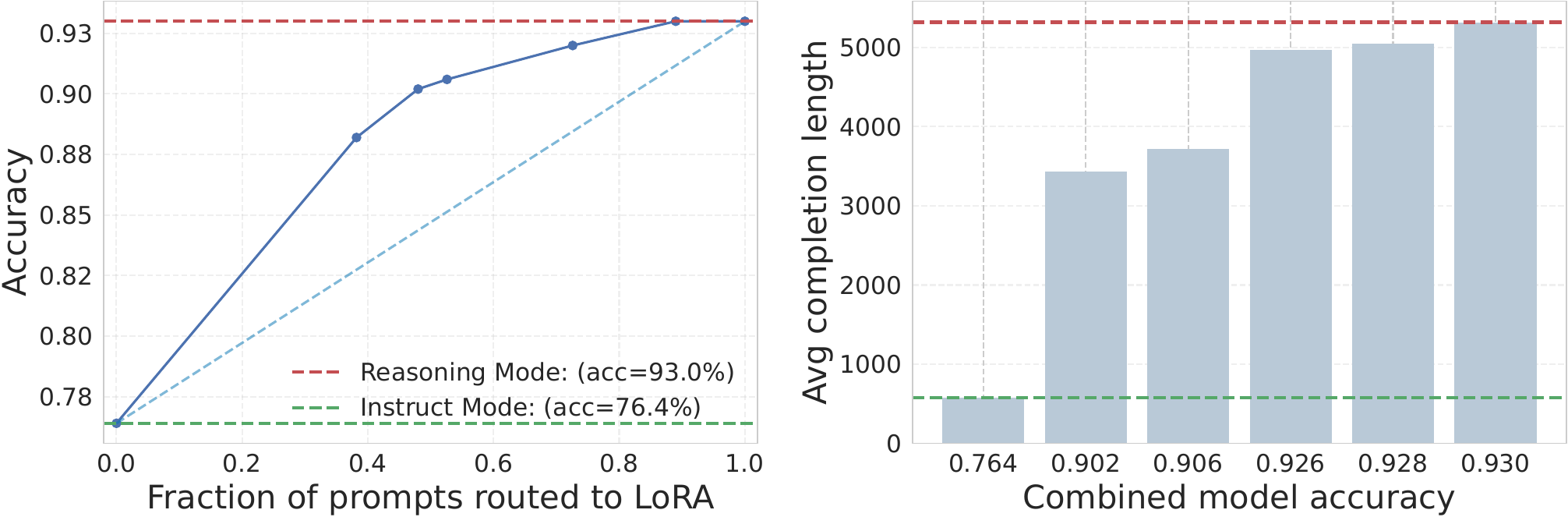}
    \caption{\textbf{Impact of the Switcher module on MATH500.} \textbf{Left:} Combined model accuracy as the fraction of queries routed to the reasoning adapters. \textbf{Right:} Average completion length versus overall accuracy across different switcher thresholds.}
    \label{fig:switcher_pareto}
\end{figure}

To rigorously assess the impact of this dynamic routing, we evaluated the switcher's performance on the challenging MATH500 benchmark as it contains questions with varying complexity. We swept across different switcher confidence thresholds to vary the fraction of prompts routed to the reasoning adapters versus the base model. Figure \ref{fig:switcher_pareto} illustrates how overall model performance changes as a larger fraction of queries is routed to the reasoning model. Here, we study the base Qwen2.5-7B-Instruct model and its counterpart trained on OT3 with LoRA adapters of rank 128 without any budget forcing. %

As more answers are generated with reasoning, accuracy rises smoothly from the base model baseline toward the reasoning-only upper bound. This demonstrates that the switcher effectively prioritizes the reasoning model on more complicated queries where it is most beneficial, allowing accuracy levels that cannot be achieved by the base model alone.

The right panel shows the corresponding computational cost, measured as average completion length. Choosing a higher-accuracy operating model requires a proportional increase in computational costs, while lower-cost regimes are possible when accuracy demands are modest. The switcher thus provides a flexible mechanism for navigating this tradeoff.

\section{Budget Forcing and Inference-Time Compute Optimization}\label{sec:budget_forcing}

CoT prompting \citep{wei2022chain} scales Inference-Time Compute (ITC) by decoding intermediate reasoning steps, substantially improving LLM performance on complex tasks, but often at the cost of high latency and large token footprints. Theoretical analysis \citep{zhang2025laws} argue that the optimal test-time compute should scale linearly with problem difficulty. Yet unconstrained models routinely violate this optimality, exhibiting degenerate verbosity and overthinking even on trivial tasks \citep{muennighoff2025s1}. 

To reconcile reasoning capability with computational efficiency, \textit{Budget Forcing} \citep{muennighoff2025s1} methods aim to align generation with explicit token/compute constraints. 
RL-based methods, among others \citep{xu2025cod,renze2024ccot,wang2025ton,huang2025hapo}, achieved impressive trade-off between performance and CoT length reduction. These methods typically require to augment the reward with a length-based penalty term \citep{aggarwal2025l1} or to enforces hard truncation constraints upon reaching a target budget \citep{liu2025dler}. Formally, a standard budget-forced reward objective can be expressed as an additive penalty:
\begin{equation}\label{eq:penalty}
    R(y,x) = R_{\text{accuracy}}(y,x) - \lambda \cdot R_{\text{budget}}(L)
\end{equation}
where $x$ is the prompt, $y$ is the generated response, $R_{\text{accuracy}}(y,x)$ is the accuracy reward, $L$ represents the total token length, and $R_{\text{budget}}(L)$ is a penalty function scaled by the hyperparameter $\lambda$.

\subsection{Soft-Barrier Reward Formulation}
\label{subsec:reward_design}

Building upon the foundations of budget-penalized RL, our reward is based on three core rationales: 
\begin{enumerate}
    \item \textbf{Avoidance of strict token matching:} We do not force the model to exactly match a predefined budget. Doing so assumes perfect a priori knowledge of the optimal compute required for a specific task, which contradicts the premise of generative exploration.
    \item \textbf{Trajectory exploration:} The model must retain sufficient degrees of freedom to explore diverse reasoning paths without premature truncation.
    \item \textbf{Prompt-adherent budget compliance:} The model must reliably satisfy the user-defined budget constraints provided in the prompt.
\end{enumerate}

To realize these principles, we prompt the model with discrete generation length budgets, specifically bucketing constraints into $1000$, $3000$, $4000$, and $6000$ tokens. Instead of an additive penalty, we introduce a multiplicative, piecewise-linear \textit{soft barrier}. This barrier decays the budget reward from $1.0$ to $0.0$ as the generation length exceeds the prompted bucket. The linear decay serves as a buffer, discouraging the model from exceeding the limit without inflicting catastrophic penalties for minor budget infractions. 

This decay operates within a symmetric window centered around the target budget $B$, where the half-size of the window, $m \in [0,1]$, is treated as a tunable hyperparameter. Formally, we define the budget reward modifier as:
\begin{equation}\label{eq:budget}
    R_{\text{budget}}(L) = 
    \begin{cases}
        1 & L \le L_{\text{low}} \\
        p & L > L_{\text{high}} \\
        1 - (1-p)\frac{L - L_{\text{low}}}{L_{\text{high}} - L_{\text{low}}} & L_{\text{low}} < L \le L_{\text{high}}
    \end{cases}
\end{equation}
where $L$ is the total length of the generated response, $p$ is the maximum budget penalty floor, $L_{\text{low}} = (1-m)B$, and $L_{\text{high}} = (1+m)B$. Through empirical observation, we noted that setting a negative penalty floor provided no optimization benefits; thus, we set $p = 0$.

The final holistic reward $R$ is then defined as the product of the task accuracy and the budget compliance modifier:
\begin{equation}\label{eqn:bf_reward}
    R(y,x) = R_{\text{accuracy}}(y,x) \times R_{\text{budget}}(L)
\end{equation}
where $R_{\text{accuracy}}(y,x) \in \{0, 1\}$ represents the binary accuracy reward. 

\paragraph{Challenges and reward hacking.} 
Because budget forcing imposes a strict constraint on the optimization manifold, it inherently induces a trade-off between reasoning performance and compute cost. We observed that naively applying a length penalty (for instance, penalizing \textit{only} the tokens within the CoT reasoning trace) is highly susceptible to reward hacking. During early iterations, the policy rapidly collapses into a degenerate, ``lazy'' strategy: the model learns to circumvent the penalty by prematurely closing the reasoning block with a \texttt{</think>} token, only to continue its verbose CoT in the final response output. By penalizing the total generation length $L$ rather than just the reasoning trace, our multiplicative formulation effectively neutralizes this exploit. Furthermore, while our final reward formulation strips away explicit format-following rewards, empirical evaluations confirm that the model consistently maintains the desired structural formatting throughout training.

\subsection{Experimental Setup}\label{subsec:experimental_setup}

\subsubsection{Training Details}
To demonstrate the efficacy of our soft-barrier reward formulation in compressing CoT reasoning, we conduct extensive experiments on state-of-the-art reasoning models. We utilize the DeepScaleR dataset \citep{luo2025deepscaler} as our primary training corpus. To maximize training stability and prevent degenerate optimization steps, we apply a rigorous filtering criterion to the dataset: any prompt exhibiting a group reward standard deviation of zero is removed, ensuring that the model always receives a meaningful comparative signal during policy updates.

We optimize our models using GRPO \citep{shao2024deepseekmath}. GRPO is particularly well-suited for reasoning tasks as it bypasses the need for a separate value model by leveraging group-scaled rewards. For a given prompt $x$, the objective minimizes the following loss:
\begin{equation}
\mathcal{L}_{\mathrm{GRPO}}(\theta \mid x) = -\frac{1}{G}\sum_{i=1}^{G} \min\!\Big( \rho_i\,A_i,\; \operatorname{clip}(\rho_i,\,1-\epsilon,\,1+\epsilon)\,A_i \Big) + \beta\,D_{\mathrm{KL}}\!\big(\pi_\theta(\cdot\mid x)\,\big\|\,\pi_{\mathrm{ref}}(\cdot\mid x)\big),
\end{equation}
where $G$ denotes the group size, and the probability ratio $\rho_i$ and the advantage $A_i$ are defined as:
\begin{equation}
\rho_i = \frac{\pi_\theta(y_i\mid x)}{\pi_{\mathrm{old}}(y_i\mid x)}, \qquad A_i = \frac{r_i - \mu_r}{\sigma_r + \varepsilon}
\end{equation}
Here, $\mu_r$ and $\sigma_r$ represent the mean and standard deviation of the rewards within the group, respectively:
\begin{equation}
\mu_r = \frac{1}{G}\sum_{j=1}^{G} r_j, \qquad \sigma_r = \sqrt{ \frac{1}{G}\sum_{j=1}^{G} (r_j - \mu_r)^2 }
\end{equation}

Our implementation is built on the \texttt{trl} library (version 0.26.2) \cite{vonwerra2020trl}. We execute training on a single compute node equipped with 8 NVIDIA H100 (80GB) GPUs. We sample 8 generations per prompt ($G=8$) during the GRPO rollouts. A comprehensive summary of the training hyperparameters is detailed in Table \ref{tab:hyperparams} in \autoref{app:budget_forcing}.

\subsubsection{Evaluation Details}
\textcolor{black}{To assess the impact of our budget forcing technique on mathematical reasoning, we use the large-scale Math500 \citep{lightman2024let} benchmark as our main testbed.
} For robust and reproducible evaluation, we employ the \texttt{lighteval} framework (version 0.8.1). All inference processes are accelerated using vLLM (version 0.10.2). To standardize the assessment of pass@1 accuracy, we apply a consistent sampling strategy across all benchmarks: generations are sampled with a temperature of 0.6, a $\text{top}\_\text{p}$ of 0.95, and a maximum completion length extended to 32K tokens to accommodate any lingering verbose trajectories from the baseline models.

\subsection{Results: Efficiency-Accuracy Trade-off}
\label{subsec:main_results}

As established, our primary objective is to compress the generated reasoning trajectories with minimal degradation in task performance. Because our soft-barrier reward formulation deliberately omits a strict regularizer weight for the budget penalty (see eq. \ref{eq:penalty}), we discovered that the Kullback-Leibler (KL) divergence penalty coefficient in GRPO ($\beta_{\text{KL}}$) serves as an effective control mechanism to enforce budget-friendly behavior. Empirically, setting $\beta_{\text{KL}} = 10^{-3}$ yields the optimal balance, significantly reducing generation length with negligible performance drops. Conversely, a relaxed penalty of $\beta_{\text{KL}} = 10^{-4}$ improves formatting adherence at very short completion lengths, albeit at the cost of a slightly higher performance regression when evaluated on larger, unbounded contexts.

\begin{figure}[!ht]
    \centering
    \includegraphics[width=\linewidth]{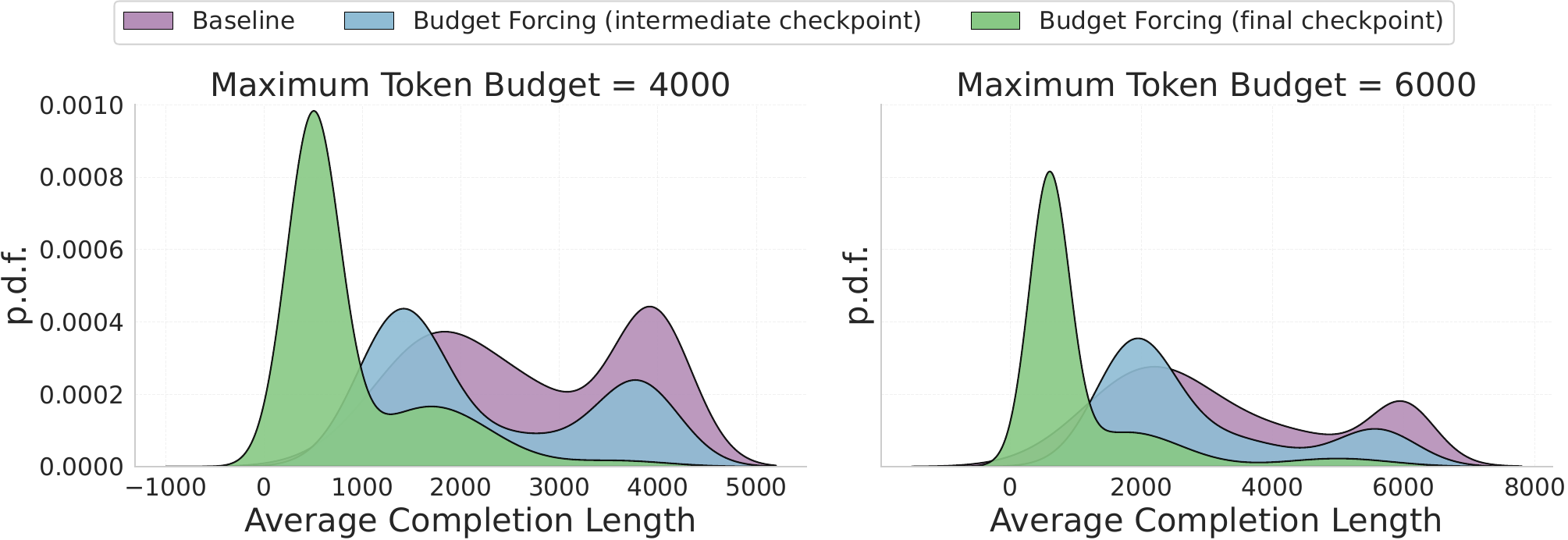}
    \caption{\textbf{Average Completion Length Distributions.} \textbf{Left:} Evaluation with a forced maximum completion length of 4K tokens. \textbf{Right:} Evaluation with a maximum of 6K tokens. Note that distribution tails extending below zero or above the maximum budget are standard artifacts of Kernel Density Estimation (KDE) curve smoothing. The progression from the baseline (\textcolor{purple}{purple}) through the intermediate (\textcolor{blue}{blue}) to the final RL fine-tuned checkpoint (\textcolor{green}{green}) demonstrates stable, progressive learning of concise generation ($\beta_{\text{KL}}=10^{-3}$).}
    \label{fig:cdf_avg_len}
\end{figure}

Figure \ref{fig:cdf_avg_len} illustrates the average completion length distributions for the unconstrained baseline (purple) and two intermediate checkpoints trained with $\beta_{\text{KL}} = 10^{-3}$. The left and right panels depict evaluations where the maximum completion length is strictly capped at 4K and 6K tokens, respectively. To enforce this hard budget during inference, we abruptly truncate the generation upon hitting the token limit and subsequently append a prompt forcing the model to immediately output its final answer.
As demonstrated in Figure \ref{fig:cdf_avg_len}, our RL fine-tuning effectively shifts the distribution density toward significantly shorter lengths. Crucially, the transition from the baseline (purple) through the intermediate checkpoint (blue) to the final policy (green) highlights a stable, progressive optimization trajectory. Rather than experiencing sudden, erratic policy collapse, the model smoothly and monotonically learns to generate more concise reasoning traces over the course of training to solve the tasks.

\begin{figure}[!ht]
    \centering
    \includegraphics[width=\linewidth]{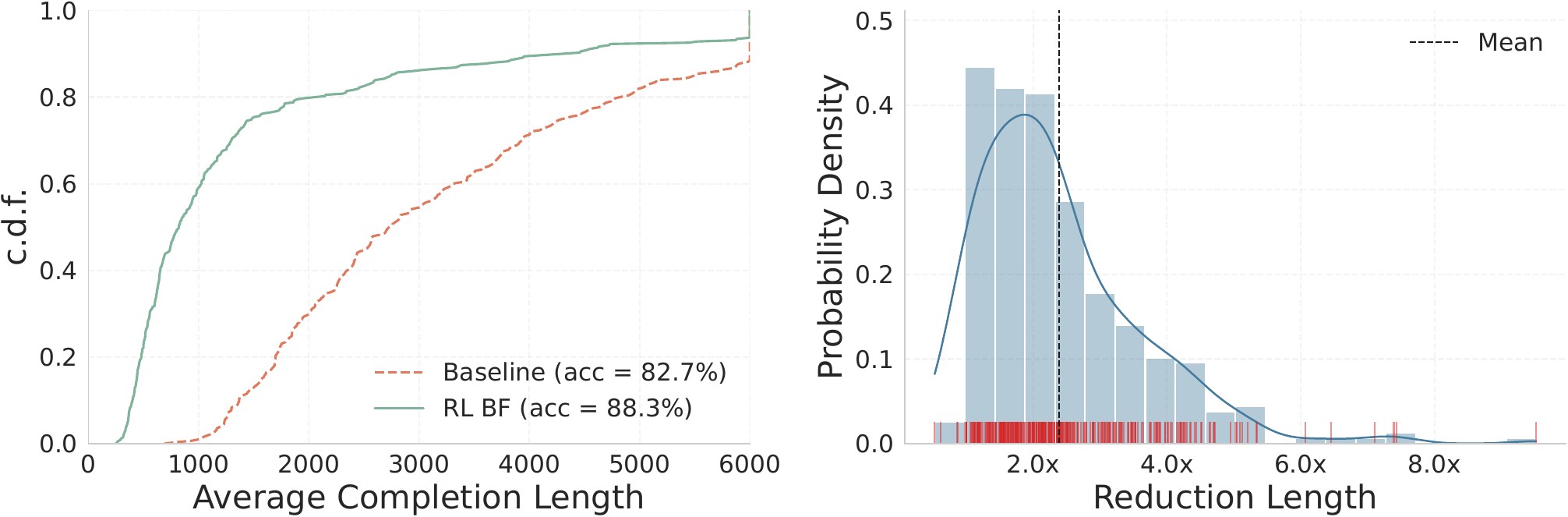}
    \caption{\textbf{Average Completion Length Comparison.} \textbf{Left:} C.D.F. of the average completion length for base model, \textcolor{orange}{orange} curve, and RL fine-tuned one, \textcolor{green}{green} curve, with $\beta_{KL}=1.e^{-3}$. We considered a maximum completion length of 6K. \textbf{Right:} Reduction in completion length from the RL fine-tuned model. We use the same models in the left plot. The RL fine tuned achieved average reduction length of $2.38 \pm 0.07$.}
    \label{fig:compl_dist_comparison}
\end{figure}

To quantify this compression, Figure \ref{fig:compl_dist_comparison} provides a granular breakdown of the actual CoT length reductions achieved via our RL fine-tuning. Specifically, the right panel of Figure \ref{fig:compl_dist_comparison} reveals that our approach yields an average completion length reduction factor of $\sim 2.4\times$, with maximum compression rates reaching up to $\sim8\times$ on certain queries. 
As noted previously, this aggressive reduction in verbosity is achieved while maintaining comparable performance to the base model, with only minimal-and in many instances, negligible-accuracy drops. %
Practically, this reduction directly translates to lower overall inference latency and a faster time-to-final-answer, making advanced reasoning models significantly more viable for deployment in resource-constrained environments.

\begin{table}[!ht]
\centering

\begin{threeparttable}

\caption{\textbf{Performance Results on MATH500.} Results for different budget values. For budgets $\in \{1K, 2K, 4K, 6K\}$, we manually force the model to generate the final answer once the limit is hit. In the last column, we do not force the model to generate the answer. We let it generate tokens untill the budget hit.}  

\label{tab:results}
\begin{tabularx}{\linewidth}{@{}l*{4}{Y}Zc@{}}
\toprule
  \textbf{Model} & \multicolumn{5}{c}{\textbf{Accuracy} $\uparrow (\%)$}  \\ \cline{2-6}
 
  & \textbf{Budget = 1K} & \textbf{Budget = 2K} & \textbf{Budget = 4K} & \textbf{Budget = 6K} & \textbf{{Average (Budget = 32K)}} \\
  
\midrule
SFT Baseline (r=128)  & 34 & 57 & 73 & 83 & \bf 95 \\  %
  \midrule
  BF RL${}^{\beta_{KL}=1.e-3}$ & \underline{62} & \underline{78} & \bf 85 & \bf 90 & \underline{92} \\  %
  BF RL${}^{\beta_{KL}=1.e-4}$ & \bf 72 & \bf 80 & \underline{84} & \underline{85} & 90 \\  %
  \bottomrule
\end{tabularx}

\end{threeparttable}

\end{table}

\begin{figure*}[!ht]
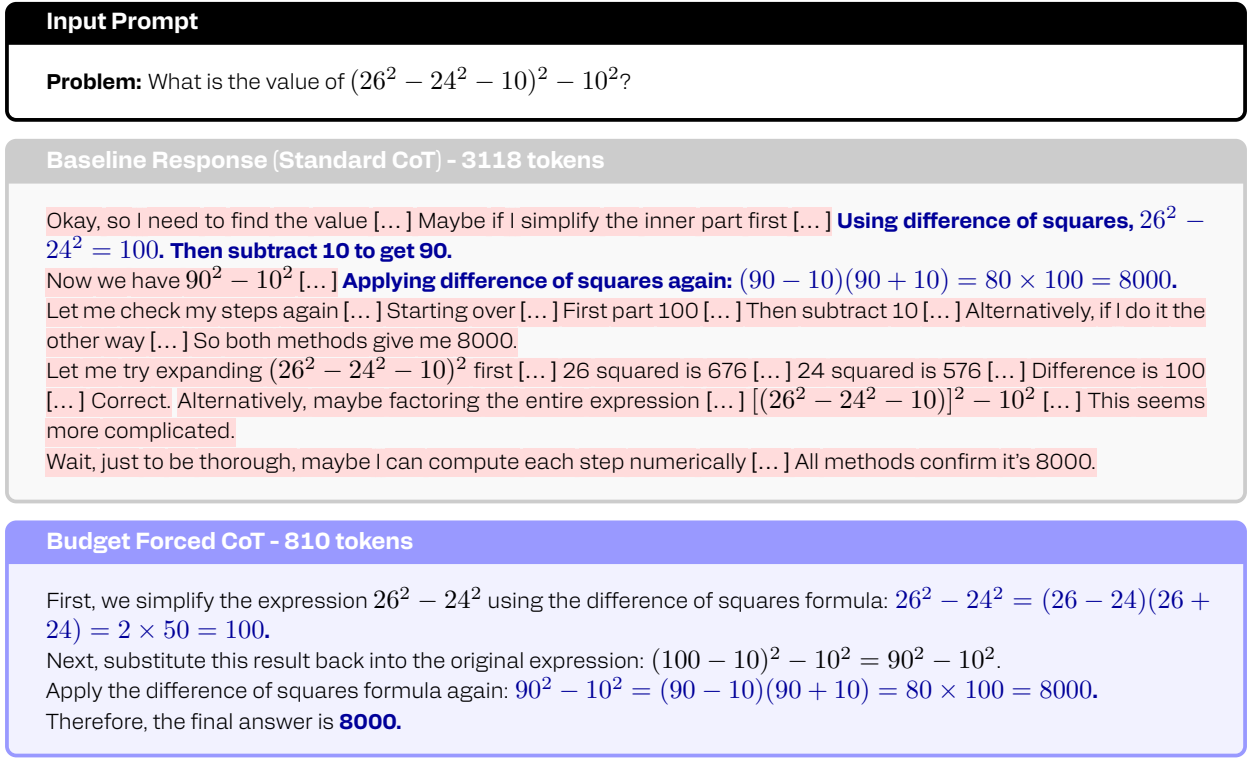

    \centering
    
    \begin{tcolorbox}[colback=white, colframe=black, title=\textbf{Input Prompt}]
    \small
    \textbf{Problem:} What is the value of $(26^2 - 24^2 - 10)^2 - 10^2$?
    \end{tcolorbox}
    
    \begin{tcolorbox}[colback=gray!5, colframe=gray!40, title=\textbf{Baseline Response (Standard CoT) - 3118 tokens}]
    \small 
    \bloat{Okay, so I need to find the value \textbf{[$\dots$]} Maybe if I simplify the inner part first \textbf{[$\dots$]}} \bridge{Using difference of squares, $26^2 - 24^2 = 100$. Then subtract 10 to get 90.}
    
    \bloat{Now we have $90^2 - 10^2$ \textbf{[$\dots$]}} \bridge{Applying difference of squares again: $(90-10)(90+10) = 80 \times 100 = 8000$.}
    
    \bloat{Let me check my steps again \textbf{[$\dots$]} Starting over \textbf{[$\dots$]} First part 100 \textbf{[$\dots$]} Then subtract 10 \textbf{[$\dots$]} Alternatively, if I do it the other way \textbf{[$\dots$]} So both methods give me 8000.}
    
    \bloat{Let me try expanding $(26^2 - 24^2 - 10)^2$ first \textbf{[$\dots$]} 26 squared is 676 \textbf{[$\dots$]} 24 squared is 576 \textbf{[$\dots$]} Difference is 100 \textbf{[$\dots$]} Correct.}
    \bloat{Alternatively, maybe factoring the entire expression \textbf{[$\dots$]} $[(26^2 - 24^2 - 10)]^2 - 10^2$ \textbf{[$\dots$]} This seems more complicated.}
    
    \bloat{Wait, just to be thorough, maybe I can compute each step numerically \textbf{[$\dots$]} All methods confirm it's 8000.}
    \end{tcolorbox}
    
    \begin{tcolorbox}[colback=blue!5, colframe=blue!40, title=\textbf{Budget Forced CoT - 810 tokens}]
    \small
    First, we simplify the expression $26^2 - 24^2$ using the difference of squares formula:
    \bridge{$26^2 - 24^2 = (26 - 24)(26 + 24) = 2 \times 50 = 100$.}
    
    Next, substitute this result back into the original expression:
    $(100 - 10)^2 - 10^2 = 90^2 - 10^2$.
    
    Apply the difference of squares formula again:
    \bridge{$90^2 - 10^2 = (90 - 10)(90 + 10) = 80 \times 100 = 8000$.}
    
    Therefore, the final answer is \bridge{8000.}
    \end{tcolorbox}
    \vskip -5pt
    \caption{\textbf{Qualitative comparison on algebraic simplification.} \textbf{Middle:} The Baseline trace correctly identifies the difference of squares strategy immediately but engages in excessive self-verification, re-calculating the result via expansion, direct computation, and alternative factorizations. \textbf{Bottom:} The Budget Forced trace recognizes the nested difference of squares structure and executes the solution linearly without redundant checking.}
    \label{fig:cot_example_2}
\end{figure*}

\subsection{Qualitative Analysis of Budget-Forced CoT}
\label{subsec:qualitative_examples}

\begin{figure*}[!t]
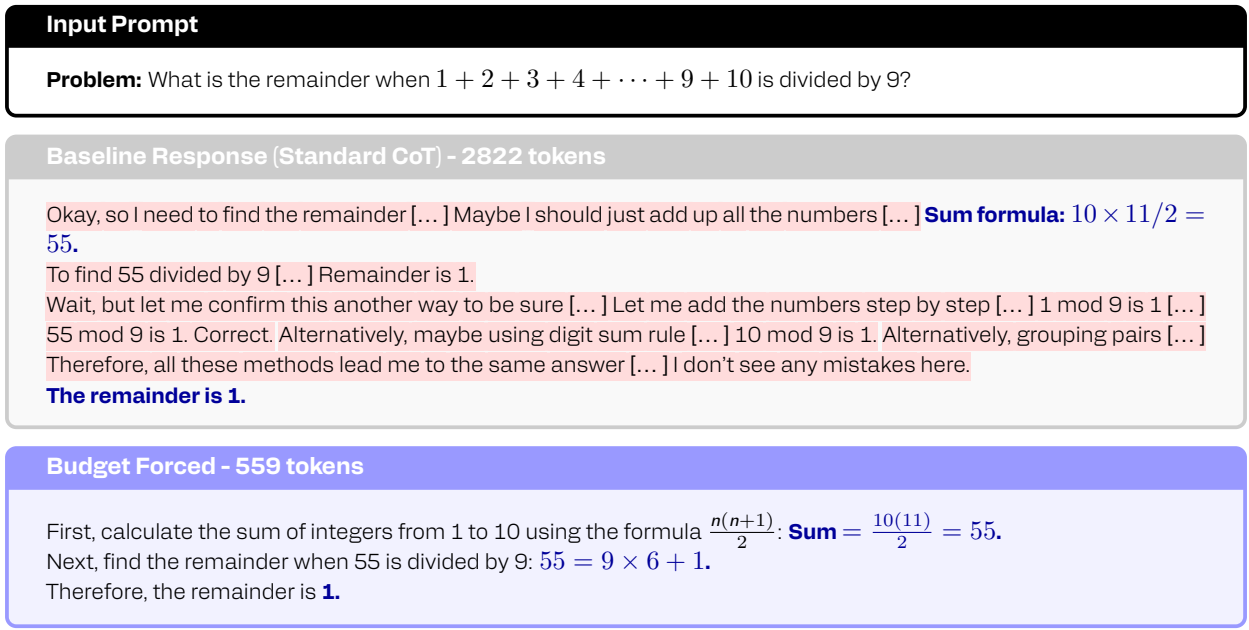

    \centering
    
    \begin{tcolorbox}[colback=white, colframe=black, title=\textbf{Input Prompt}]
    \small
    \textbf{Problem:} What is the remainder when $1 + 2 + 3 + 4 + \dots + 9 + 10$ is divided by 9?
    \end{tcolorbox}
    
    \begin{tcolorbox}[colback=gray!5, colframe=gray!40, title=\textbf{Baseline Response (Standard CoT) - 2822 tokens}]
    \small 
    \bloat{Okay, so I need to find the remainder \textbf{[$\dots$]} Maybe I should just add up all the numbers \textbf{[$\dots$]}} \bridge{Sum formula: $10 \times 11 / 2 = 55$.}
    
    \bloat{To find 55 divided by 9 \textbf{[$\dots$]} Remainder is 1.}
    
    \bloat{Wait, but let me confirm this another way to be sure \textbf{[$\dots$]} Let me add the numbers step by step \textbf{[$\dots$]} 1 mod 9 is 1 \textbf{[$\dots$]} 55 mod 9 is 1. Correct.}
    \bloat{Alternatively, maybe using digit sum rule \textbf{[$\dots$]} 10 mod 9 is 1.}
    \bloat{Alternatively, grouping pairs \textbf{[$\dots$]}}
    \bloat{Therefore, all these methods lead me to the same answer \textbf{[$\dots$]} I don't see any mistakes here.}
    
    \bridge{The remainder is 1.}
    \end{tcolorbox}
    
    \begin{tcolorbox}[colback=blue!5, colframe=blue!40, title=\textbf{Budget Forced - 559 tokens}]
    \small
    First, calculate the sum of integers from 1 to 10 using the formula $\frac{n(n+1)}{2}$:
    \bridge{Sum $= \frac{10(11)}{2} = 55$.}
    
    Next, find the remainder when 55 is divided by 9:
    \bridge{$55 = 9 \times 6 + 1$.}
    
    Therefore, the remainder is \bridge{1.}
    \end{tcolorbox}
    \vskip -5pt
    \caption{\textbf{Qualitative comparison on modular arithmetic.} \textbf{Middle:} The Baseline trace correctly computes the sum and remainder immediately but engages in extensive, redundant verification using four different methods (step-by-step addition, digit sum rule, pairing, and re-calculation). \textbf{Bottom:} The Budget Forced trace performs the direct calculation and returns the result without hesitation.}
    \label{fig:cot_example_4}
\end{figure*}

To examine the mechanics of our budget forcing objective at the trajectory level, we qualitatively compare the reasoning traces of the unconstrained baseline against our budget-forced model across four distinct mathematical domains: number theory, algebraic simplification, pattern recognition, and modular arithmetic (Figures \ref{fig:cot_example_2}--\ref{fig:cot_example_4} and Figures \ref{fig:cot_example_1}--\ref{fig:cot_example_3} in \autoref{app:budget_forcing}). 
A consistent pattern emerges from this analysis: the unconstrained baseline frequently suffers from severe epistemic hesitation. While it typically identifies the correct logical strategy early in the generation process, it expends thousands of tokens on redundant self-verification, testing alternative (and often less efficient) methods, and hypothesizing trivial errors. For instance, in Figure~\ref{fig:cot_example_2} and Figure~\ref{fig:cot_example_4}, the baseline arrives at the correct answer almost immediately but falls into extensive validation loops, re-calculating the result using three to four different approaches. 

In contrast, our budget-forced policy learns to confidently trust its initial, correct logical derivations. It successfully prunes these redundant validation loops and verbose syntactic parsing, drastically reducing the trace length while strictly preserving the essential reasoning backbone and human readability. Furthermore, although we omit special delimiter tokens (e.g., \texttt{<think>}) in the visualizations for brevity, we observe that the budget-forced model robustly generates them,  maintaining the required CoT and final-answer formatting constraints.

\section{Parallel Test-Time Scaling and Reasoning}\label{sec:parallel}

During autoregressive generation with LLMs, one of the dominant runtime bottlenecks is repeatedly loading model weights of the individual layers to produce the next token.
As a result, there is often potential to increase the total amount of computation performed per user query without incurring significant overhead in runtime.
A straightforward way to do this is to generate multiple independent samples in parallel: instead of producing one CoT trajectory and one final answer, this approach allocates additional compute to produce several trajectories concurrently.

Parallel generation is not only attractive from a systems perspective, but it also improves accuracy.
Across benchmarks and models, generating multiple candidate solutions and then aggregating them has repeatedly shown consistent performance gains.
A common and simple aggregation strategy is majority voting over individual answers, where the most frequently produced answer is selected as the final answer.
In practice, majority voting is widely used as a reliable tool for achieving an additional boost in final model performance from the same underlying base model. 

Initial works \citep{cobbe2021training,wang2022self} indicated that the same LLM can be made to generate multiple diverse and independent responses towards finding a better solution. Scaling compute by sampling multiple independent responses prior to aggregation has shown \cite{brown2024large, wu2025inference, snell2025scaling} to be an effective paradigm enabling smaller models to outperform larger models for the same inference compute budget. In these cases, the key is to score the solutions (e.g., by frequency, or using an external verifier) and predict the highest-scoring solution. Parallel reasoning design relies on selecting a sampling scheme, design and employ reward models, and finally aggregation scheme to generate the final answer.
Diversity among responses \citep{brown2024large} is crucial to increasing the probability of sampling the correct response.  Increasing diversity is typically done by auto-regressive generation at a higher sampling temperature \citep{renze2024effect}.  While responses are typically sampled independently, recent works explore inter-dependent sampling \citep{hsu2025group,pan2025learning,rodionov2025hogwild,zheng2025parallel} and guided-search \citep{yao2023tree,ning2023skeleton,li2025enhancing} towards generating a better response candidate pool. In this work, we focus on independent sampling schemes.
Reward models (also referred to as verifiers) estimate one or more scalar-valued scores for a given response. There is a long line of work on improving reward models, with  more accurate reward models \citep{liu2024skywork,wang2024helpsteer2}, more granular score assignment (e.g., per-step scores with process reward models) \citep{lightman2024let,wang2023math,zhang2025lessons}, scaling verification with more compute \citep{liu2025inference}, and  extending reward models to other reasoning domains \citep{chae2025web,chen2025scaling} beyond mathematical reasoning. In this work, we use the reward models for scoring the outcome. 
Finally, aggregation is essential to drafting a specific response from a pool of multiple response-score pairs.  Existing works have primarily looked at this from a rank-and-select lens: to rank the responses either by frequency of occurrence (`self-consistency' or `majority voting') \cite{wang2022self}, or using external reward models to score responses \citep{cobbe2021training,wang2023math}. Note that these strategies lead to a zero-sum situation: the top-ranking solution is selected and the rest discarded. Consequently, a more recent line of work \citep{khairi2025making, li2025drafts,qi2025learning,zhao2025majority} investigates synthesizing (rather than selecting) a final response based on the candidate responses.

Scaling compute at test-time generally implies additional latency and compute overhead and hence making it especially challenging to realize on resource-constrained edge devices.
Some benefits of parallel TTS on edge devices were discussed in \cite{hao2025scaling} with a separate verifier.
This separation prevents efficient reuse of intermediate computations, e.g., through the KV-cache.
Consequently, scaling parallel compute \textit{efficiently} has gained some attention in the research community. 
Towards efficiently reasoning, several works have investigated generating solutions by performing fewer FLOPs \citep{wu2025inference} and reducing the memory footprint \citep{sun2024fast,hooper2025ets} required during generation.
Closest to our verifier design presented in this paper is GenRM \citep{zhang2025generative}, where the authors investigate finetuning the base generative model to additionally perform prompt-based verification.
Such a joint generation-verification paradigm is appealing for edge devices, since generation and verification can be performed with minimal movement of parameters between DRAM and flash memory.

\subsection{Efficient Verifier Design for Edge Compute}
Parallel reasoning produces a set of $N$ independent CoT traces and corresponding final answers. 
The central question then becomes: given these $N$ candidates, how do we reliably choose the correct one?
Majority voting provides a strong baseline, but it is not always sufficient, especially when the answer space is large or ambiguous.
This motivates introducing an explicit selection mechanism that can score candidates and prefer those most likely to be correct, while still benefiting from the diversity created by independent sampling.

A natural approach is to add a verifier model that evaluates each candidate solution.
However, deploying a separate verifier on edge devices at the same scale as the generator can be prohibitive in storage and memory footprint, and often becomes especially costly in latency.
To avoid this, we aim to reuse the generator as effectively as possible.
Concretely, we keep the same base model and add a lightweight verifier head: a separate linear layer applied to the final token embedding, followed by a sigmoid activation to yield a scalar ``correctness'' score.
Figure~\ref{fig:overview}b illustrates this approach.
This design keeps additional parameters minimal and, crucially, allows KV-cache reuse for all generated responses, since the verifier head can operate on representations already computed during generation.

In addition to the linear head, we append a short verification prompt after each generated response, asking the model whether the proposed solution is correct.
Empirically, this extra query has been beneficial compared to relying on a linear head alone without an explicit verification prompt.
Operationally, this strategy only requires a small additional prefill step for the verification prompt for each individual generation, while preserving KV-cache reuse from the original generation.
The verifier therefore adds only modest overhead per candidate while improving the reliability of the selection process.

To obtain the best results, we combine majority voting with verifier scoring into a weighted majority vote.
Instead of counting each candidate answer equally, we weight each candidate’s vote by its verifier score.
Intuitively, candidates that the verifier deems more likely to be correct contribute more to the final decision, while still retaining the robustness benefits of aggregation across multiple samples.
This hybrid approach preserves the simplicity and stability of voting while incorporating a learned notion of solution quality, which is particularly helpful when the candidate set contains a mix of superficially plausible but incorrect solutions alongside correct ones.

\subsection{Training and Evaluation Details}\label{sec:training_and_quantization}
Verification is treated as a binary classification problem where a candidate response of the generator is labeled as correct or incorrect with respect to the ground-truth answer.
To generate training data for the verifier, we use 97.5\% of the 7.5k questions from the MATH training set~\cite{hendrycks2021measuring}.
The remaining 2.5\% of questions are reserved for validation of the verifier.
To construct a diverse training set and a validation set for fast evaluation, we generate 16 candidate responses for each training question, while we limit ourselves to 4 generated responses in the validation set.
The verifier head uses a sigmoid activation to produce a probability-like score, and we train it with binary cross-entropy loss.
This setup aligns the verifier objective directly with the downstream selection problem: distinguishing correct from incorrect candidate solutions produced by the generator under the same sampling procedure used at inference time.

\begin{table}[ht]
  \centering
  \caption{Accuracy of our lightweight verifier weighted majority voting compared to majority voting without verifier and greedy decoding on MATH500. The mean and standard deviation are computed from 20 random draws from 16 independent 4bit-weight-quantized Qwen-2.5-7B-Instruct responses.}
  \vskip 0.15in
  \label{tab:wmv_results}
  \begin{tabular}{lccccc}
    \toprule
    \textbf{Parallel Responses} & \textbf{1} & \textbf{2} & \textbf{4} & \textbf{6} & \textbf{8} \\
    \midrule
    Greedy (baseline) & 71.0 &  - & - &  - &  - \\
    Majority Vote & 69.9 $\pm$ 1.3 & 70.0 $\pm$ 1.3 & 75.1 $\pm$ 1.0 & 76.6 $\pm$ 1.0 & 77.5 $\pm$ 0.8 \\
    Weighted MV (ours) & 69.9 $\pm$ 1.3 & 72.7 $\pm$ 1.0 & 76.1 $\pm$ 0.9 & 77.5 $\pm$ 0.8 & 78.2 $\pm$ 0.7 \\
    \bottomrule
  \end{tabular}
\end{table}

\subsection{Results}
We evaluate the proposed lightweight verifier on MATH500 using a 4-bit-weight-quantized Qwen-2.5-7B-Instruct model, comparing greedy decoding, standard majority voting, and our weighted majority voting.
Table~\ref{tab:wmv_results} summarizes accuracy as a function of the number of parallel responses, sampled with temperature set to 0.7.

Even with very limited parallelism, parallel test-time scaling yields immediate benefits.
With just two parallel responses, weighted majority voting improves accuracy to 72.7\%, outperforming both the greedy baseline (71.0\%) and standard majority voting (70.0\%).
This highlights a key advantage of incorporating verification: when two sampled responses disagree, majority voting cannot break ties, whereas the verifier-weighted scheme can consistently select the more reliable candidate.

As the degree of parallelism increases, both majority voting and weighted majority voting exhibit steady gains, confirming prior observations that parallel sampling improves the probability of generating a correct solution.
However, weighted majority voting consistently outperforms unweighted majority voting across all parallelism levels, and at eight parallel responses, weighted majority voting improves upon the baselines by 10\%.
Importantly, the variance across random draws is also slightly reduced compared to majority voting, suggesting that verifier weighting provides a more stable aggregation mechanism.

Despite its simplicity, the verifier delivers strong benefits.
Architecturally, it amounts to generating only a minimal amount of overhead, effectively one extra token per stream.
Because the verifier reuses the generator’s KV-cache, it does not require reprocessing the original prompt or response, avoiding the dominant memory and latency costs typically associated with separate verifier models.
This makes the performance gains particularly compelling in edge settings, where memory bandwidth and storage are tightly constrained.

\section{Quantization}\label{sec:quantization}

In the following section, we discuss the details of our quantization methodology.
We begin by providing a brief overview of neural network quantization and a summary of recent methods for quantizing LLMs in Section~\ref{sec:quantization_background}.

We outline our strategy for quantizing base LLM in Section~\ref{sec:quantization_base} demonstrated on Qwen2.5-7B-Instruct, and later equip it with reasoning capabilities in Section~\ref{sec:qarm}.
Lastly, we provide the specifics of quantized model export and deployment on device in Section~\ref{sec:quantization_on_device}.

\subsection{Background and Related Work}\label{sec:quantization_background}
\paragraph{Quantization.}
Neural network quantization is one of the most powerful ways to reduce model footprint, data transfer and compute requirements~\citep{krishnamoorthi2018quantizing,nagel2021white}.
By quantizing a model, high bit-width floating point weights and activations can be represented using low-bit numbers. 
Next to reducing model size, the use of low-bit fixed-point representations, such as~\nf{INT8}, can also significantly reduce the latency and energy consumption~\citep{horowitz}.

\upd{We use the following definition of the quantization-dequantization function:}
\begin{equation}
    \widehat{\vx} := q\p{\vx;\,\vs,\vz,b} = \vs \cdot 
    \vphantom{\Bigg(} \Big(\,\smash{\underbrace{\clip\!\p{\round{\frac{\vx}{\vs}}+\vz;-2^{b-1},2^{b-1}-1}}_{\text{\normalsize $=: \vx_\Z$}}} - \vz \Big),
    \label{eq:dequant}
\end{equation}
\\[0.33em]
where $\vx$ denotes the quantizer input (i.e., network weight or activation tensor), 
$\vs$ the high precision (\nf{FP32} / \nf{FP16} / \nf{BF16}) quantization scale,
$\vz$ the integer zero offset, and $b$ the bitwidth.
$\round{\cdot}$ denotes the round-to-nearest-integer operator.
$\vx_\Z$ is a $b$-bit integer~\emph{quantized representation} of the input $\vx$.
Quantization parameters $\vs$, $\vz$ can be shared across the components of $\vx$ (typically per-channel or block-wise).
This quantization scheme is called~\emph{uniform affine} or~\emph{asymmetric} quantization~\citep{hubara2017quantized,krishnamoorthi2018quantizing,zhou2016dorefa} and is one of the most commonly used quantization schemes because it allows for efficient implementation of fixed-point arithmetic.
In the case of~\emph{symmetric} quantization, we restrict the quantization grid to be symmetric around $\vz=\vzero$.

\upd{
Quantization methods can generally be categorized into~\emph{post-training quantization} (PTQ) and~\emph{quantization-aware training} (QAT) families.
PTQ algorithms convert pretrained high-precision networks directly into fixed-point models without the need for the original training pipeline~\citep{banner2018post,cai2020zeroq,choukroun2019low,hubara2020improving,meller2019same,zhao2019improving,Nagel_2019_ICCV,nagel_up_2020,li2021brecq}. 
These approaches are fast, easy to use, and typically rely only on a small calibration dataset.
In contrast, QAT methods~\citep{gupta2015deep,jacob2018quantization,lsq,nagel_oscillations_2022} simulate quantization during training to find more optimal solutions, but generally require longer training, more memory, labeled data, and careful hyperparameter tuning.
}

\paragraph{LLM Quantization.}
The excessive training cost and memory usage of traditional QAT methods renders them less practical for quantizing modern LLMs, although some works such as LLM-QAT~\citep{liu2023llm} and BitDistiller~\citep{du2024bitdistiller} explore QAT with knowledge distillation.
Notably, \citep{liu_paretoq_2025,chen2025scaling_qat} are the only studies we are aware of that successfully scale QAT to billions of tokens.
Several papers explored the combination of QAT and parameter-efficient fine-tuning (PEFT), including~\citep{dettmers2024qlora,xu2023qa,li2023loftq,guo2023lq,kim2024memory,bondarenko2024low}.
Most of these approaches offer a substantial memory reduction compared to traditional QAT, but generally are not focused on inference efficiency.
For instance, QLoRA~\citep{dettmers2024qlora} quantizes the pretrained weights to 4 bit using (a non-uniform)~\nf{NF4} format but dequantizes them in the forward pass back to~\nf{BF16}.

Post-training quantization of LLMs is a challenging task due to presence of strong numerical outliers in weights and activations~\citep{bondarenko_understanding_2021,kovaleva_bert_2021,dettmers_gpt3_int8_2022,bondarenko_quantizable_2023,sun_massive_2024}.
The core challenge is that quantizing outliers onto a fixed-point grid forces a range-precision trade-off:
increasing the dynamic range captures outliers but sacrifices precision near zero, while retaining precision requires clipping them -- both of which strongly degrade model performance.

Existing LLM PTQ methods can be broadly categorized into~\emph{weights-only} quantization and~\emph{weight-activation} quantization.
Weights-only quantization focuses on converting weights to low-bit values.
GPTQ~\citep{frantar_gptq_2022} employs second-order information to iteratively round grouped weights and correct the quantization error in the remaining groups.
SpQR~\citep{dettmers2023spqr}, AWQ~\citep{lin2023awq} and OWQ~\citep{lee2024owq} emphasize the importance of so-called ``salient'' weights that correspond to high-magnitude activations.
Other recent W-only methods include~\citep{jeon2023frustratingly,lee2023flexround,luo2023long,chee2024quip}.
Weight-activation quantization compresses both weights and activations.
SmoothQuant~\citep{xiao2023smoothquant}, \texttt{LLM.int8()} / \texttt{GPT3.int8()}~\citep{dettmers_gpt3_int8_2022} and Outlier Suppression~\citep{wei2022outlier} achieve W8A8 quantization by managing activation outliers.
\texttt{LLM.int8()} uses mixed-precision decomposition, while the other two employ channel-wise scaling.
Some of the other recent W\&A PTQ methods are~\citep{lee2023enhancing,liu2023qllm,wei_outlier_2023,yuan2023rptq,tang2024easyquant,yao2022zeroquant,lin2024qserve}.

\paragraph{LLM Quantization using FPTs.}
\upd{
A promising direction in LLM quantization is the use of~\emph{rotations} and other~\emph{function-preserving transformations} (FPTs).
\citet{Nagel_2019_ICCV} first explored FPTs for CNN quantization, showing that ReLU and per-channel scaling commute, enabling cross-layer rescaling of weights. 
In the LLM setting, \citet{xiao2023smoothquant} propose migrating problematic outliers from the activations into the weights through online per-channel scaling applied before linear layers.
Follow-up work extends this idea by incorporating shifts into the scaling~\citep{wei_outlier_2023}, scaling vectors for queries and keys~\citep{shao_omniquant_2024}, channel-mixing transforms~\citep{chee2024quip}, randomized Hadamard transforms to reduce outliers~\citep{ashkboos_quarot_2024,liu_spinquant_2024}, other online rotations~\citep{lin_duquant_2024}, combinations of scaling and rotations~\citep{hu_ostquant_2025}, and Kronecker-structured matrix transforms~\citep{sun_flatquant_2024}.
}

Recently, FPTQuant~\citep{van2025fptquant} introduced three novel, lightweight, and expressive FPTs to facilitate quantization of transformers.
By leveraging the equivariances and independencies inherent to modern transformers, these FPTs are designed to maintain the model's function while shaping the intermediate activation distributions to be more quantization friendly.
As a result, FPTQuant enables static~\nf{INT4} quantization with virtually no overhead and no custom kernels, is very fast and performs on par or exceeds most prior work.

\subsection{Quantizing Base Language Model}\label{sec:quantization_base}

\paragraph{Quantization setup.}
To run the model efficiently on our hardware, we quantize the weights of all linear layers including the final LM head using~\nf{INT4} per-channel uniform affine quantization (\eqref{eq:dequant}).
To further improve efficiency and reduce latency, we use~\nf{INT8} KV-cache, \nf{INT8} input embeddings and~\nf{INT16} for all remaining activations (all per-tensor).
For brevity, we will refer to this configuration as `W4A16KV8'.
We use symmetric quantization for weights, KV-cache and embeddings, and asymmetric quantization for activations, which is a common setting.

\begin{figure*}[t]
    \centering
    \includegraphics[width=\textwidth]{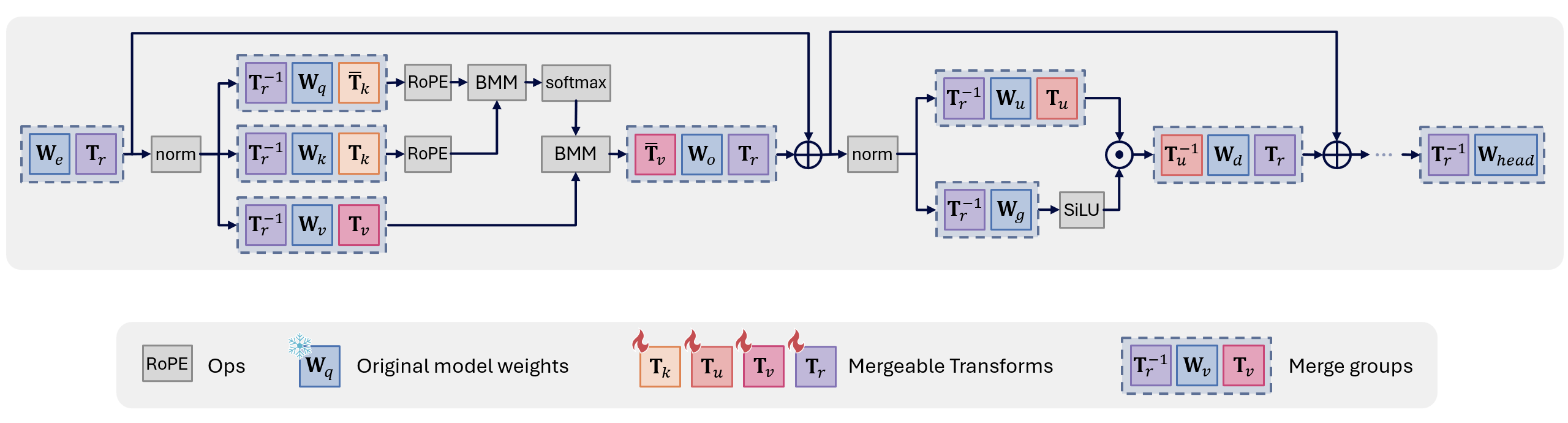}
    \caption{\small \textbf{Function-Preserving Transformations}.
We use 4 transform types from FPTQuant: scale-and-rotate transform $\TT_k$ merged into query and key, a per-channel scaler $\TT_u$ merged into up and down projection, and $\TT_v$ that consists of invertible matrices per head merged into value and output weights, and a rotation matrix $\TT_r$ for rotating residuals (shared across layers).
After training of the transforms is complete, the transformation parameters from each `merge group' are merged into the original model weights $\tW$.
    }
    \label{fig:quantization_transforms}
    \vspace{0mm}
\end{figure*}

\paragraph{Transformations.}
To maximize the accuracy of the quantized model, we apply the subset of fully-mergeable transformations from FPTQuant~\citep{van2025fptquant} (Figure~\ref{fig:quantization_transforms}):
\renewcommand\labelitemi{{\Large \boldmath$\cdot$}}
\begin{itemize}[leftmargin=*, itemsep=0.05em]
    \vspace{-0.5em}
    \item a pair of pre-RoPE transforms $(\TT_k$, $\bar{\TT}_k)$, where $\TT_k$ is applied to keys and $\bar{\TT}_k$ can be interpreted as an inverse of $\TT_k$, applied to the queries;
    \item $(\TT_u,\TT^{-1}_u)$ a per-channel scaler merged into up and down projection weights;
    \item multi-head value transforms $(\TT_v,\bar{\TT}_v)$, which consist of invertible matrices per head merged into value and output weights;
    \item and a rotation matrix $(\TT_r,\TT^{-1}_r)$ for rotating the residuals (applied at the beginning and the very end of each transformer block and is shared).
\end{itemize}
This set of transformations $\TT := \{\TT_k, \TT_u, \TT_v, \TT_r\}$ will help shape the intermediate activation distributions to be more quantization friendly, while keeping the unquantized model outputs intact.
All of the above FPTs are fully mergeable, so that we can run the model without any extra inference overhead on our hardware.

\paragraph{Training and evaluation details.}
Following literature~\citep{van2025fptquant,lee2025unifying}, we train the model on DCLM-Edu~\citep{allal2025smollm2}, a cleaner filtered version of DCLM~\citep{li2024datacomp} obtained by applying an educational quality classifier~\citep{penedo2024fineweb}.
We initialize quantization parameters by minimizing the $L^p$ ($p=2$) norm between quantized and unquantized tensors, and then train transformation $\TT$ and quantization parameters $\{\vs, \vz\}$ end-to-end, closely following the pipeline of FPTQuant.
We simulate quantization using FastForward~\citep{fastforward}.
For brevity, we denote to the aforementioned quantization pipeline as \boldourquant.

To assess the predictive performance of the quantized base language model, we follow the previous work~\citep{frantar_gptq_2022,xiao2023smoothquant,shao_omniquant_2024,van2025fptquant,sun_flatquant_2024}, and report WikiText-2 test perplexity (assuming a sequence length of $4096$).
We also report an average zero-shot accuracy of a set of common sense reasoning (CSR) tasks, that includes PIQA~\citep{bisk_piqa_2020}, WinoGrande~\citep{sakaguchi_winogrande_2021}, HellaSwag~\citep{zellers_hellaswag_2019}, ARC-e and ARC-c~\citep{clark_think_2018}, and LAMBADA~\citep{paperno_lambada_2016}.
Finally, we report 5-shot accuracy on MMLU~\citep{hendrycks2020measuring}.
For CSR and MMLU evaluation, we use the LM Harness framework~\citep{sutawika2025eleutherai}.

\paragraph{Results.}
We summarize our results for quantized base model in Table~\ref{tbl:quantization_results_base_model}.
As we can see, the simplest PTQ pipeline with min-max range estimation experiences an unacceptable accuracy/perplexity drop.
Both strong numerical outliers in the activations (even with 16-bits!) but mainly the catastrophic loss of precision in the quantized 4-bit weights lead to such poor performance. 

Employing the set of function-preserving mergeable transformations $\TT$ already significantly improves the distribution of weights and activations, leading to much better accuracy, even with min-max range setting.
Further, using a better range initialization together with end-to-end learning both progressively recover a greater portion of the full-precision model performance.
In the end, we match full-precision accuracy on CSR, and have about 0.4 perplexity drop on WikiText-2 and just under 1.5\% accuracy drop on MMLU, where the latter is known to be quite a challenging benchmark.
Overall, \ourquant-quantized base model demonstrates strong performance, given that the entire process took less than 24 hours on a single Nvidia H100 80GB GPU.

\begin{table}[bt]
    \setlength{\tabcolsep}{6pt}
    \centering
    \caption{\upd{\small \textbf{Quantized W4A16KV8 base model (Qwen2.5-7B-Instruct) results}. 
We report Wikitext perplexity, average 0-shot CSR, and 5-shot MMLU accuracies.
`$L^p$' denotes the use of $L^p$ range initialization, $\TT$ = using the set of mergeable transforms, `train' = end-to-end training of transformation and quantization parameters.}
    }
    \label{tbl:quantization_results_base_model}
    \vspace{0.25cm}
    \renewcommand{\arraystretch}{1.1}
    \begin{tabular}{lccccccc }
        \toprule
        \textbf{Method} & \textbf{Bitwidth} & $L^p$ & $\TT$ & \textbf{train} & \textbf{WikiText-2} ($\downarrow$) & \textbf{CSR} ($\uparrow$) & \textbf{MMLU} ($\uparrow$) \\
        \hline
        Full-precision & \nf{BF16} & {\na} & {\na} & {\na} & 6.85 & 72.90 & 74.28 \\
        \hline
        {Min-max quantization} & {W4A16KV8} & {\no} & {\no} & {\no} & 102.4 & 51.71 & 62.35 \\
        & {W4A16KV8} & {\yes} & {\no}  & {\no} & 9.18 & 65.83 & 67.59 \\
        & {W4A16KV8} & {\no}  & {\yes} & {\no} & 8.48 & 67.85 & 69.06 \\
        & {W4A16KV8} & {\yes} & {\yes} & {\no} & 7.53 & 70.68 & 72.26 \\
        \hline \\[-1.44em]
        \boldourquant{}~(ours) & {{W4A16KV8}} & {\yes} & {\yes} & {\yes} & \textbf{7.26} & \textbf{72.94} & \textbf{72.81} \\
        \bottomrule
    \end{tabular}
    \renewcommand{\arraystretch}{1}
        \vspace{-.15cm}
\end{table}

\subsection{Quantization-Aware Modular Reasoning}\label{sec:qarm}
Quantizing the base model will inevitably affect the underlying activation distributions.
To achieve the best performance, it is crucial to account for these changes when applying subsequent fine-tuning, including our Reasoning LoRA protocol described in Section~\ref{sec:lora}. 

Our approach follows the general paradigm of QLoRA~\citep{dettmers2024qlora} and related techniques~\citep{xu2023qa,li2023loftq}, in which LoRA adapters are trained on top of a frozen, quantized base model.
To further improve memory and runtime efficiency, we quantize the trained LoRA adapter weights to~\nf{INT8} and use~\nf{INT16} activations during inference.
As with the base model, we use symmetric per-channel quantization for weights and asymmetric per-tensor quantization for activations.
We denote the aforementioned technique as~\emph{Quantization-Aware Modular Reasoning} (QAMR).

\paragraph{Training and evaluation details.}

We follow the training and evaluation protocols described in Section~\ref{sec:lora}.
For training, we use the OT3~\cite{guha2025openthoughts} dataset. 
To assess the reasoning capabilities of our quantized reasoning model, we use a comprehensive set of benchmarks, including {AIME 24/25}~\cite{aime}, {MATH500}~\cite{hendrycks2021measuring}, {GPQA Diamond}~\cite{rein2024gpqa}, and {AMC23}~\cite{amc23}.
We conduct an ablation study to assess the contributions of both the improved base model quantization pipeline and the proposed QAMR approach using a random subset of 50k training examples, while reserving the full training dataset for the final set of results.

\paragraph{Results.}

We observe in Table~\ref{tbl:quantization_results_reasoning_model} that a naïvely quantized base model combined with a full-precision reasoning module (i.e., without QARM) is essentially non-functional.
Qualitatively such model outputs seemingly random tokens, without any structure or relevance to the tasks at hand.

In contrast, applying QARM -- even with relatively short training -- recovers a substantial portion of the
performance on tasks such as MATH500 and GPQA.
Notably, quantization-aware modular reasoning is~\emph{crucial for learning anything at all}.

Further, using {\ourquant} offers a stronger starting point for training the reasoning module and consistently improves performance across all benchmarks except AMC23, for which longer training is required.
By shaping the underlying activation distributions to be more quantization-friendly, using {\ourquant}-quantized base model leads to significantly fewer training instabilities and enables faster learning compared to a base model quantized with a standard min-max range setting.

Finally, when combined with extended training, our approach achieves performance within roughly 2\% of an equivalently trained full-precision reasoning model on average, while being significantly more compact, and inference-efficient.

\begin{table}[bt]
    \setlength{\tabcolsep}{6pt}
    \centering
    \caption{\upd{\small \textbf{Quantized W4A16KV8 reasoning model results (Qwen2.5-7B-Instruct base)}. We report AIME24/25, MATH500, GPQA, and AMC23 accuracies (higher is better). N = number of OT data examples used for training reasoning module.}
    }
    \label{tbl:quantization_results_reasoning_model}
    \vspace{0.25cm}
    \renewcommand{\arraystretch}{1.1}
    \begin{tabular}{cccccccccc }
        \toprule
        \textbf{Bitwidth} & {\boldourquant} & \textbf{QAMR} & \textbf{N} & \textbf{AIME24}  & \textbf{AIME25} & \textbf{MATH500} & \textbf{GPQA} & \textbf{AMC23} & \textbf{Avg} \\
        \hline
        \nf{BF16} & {\na} & {\na} & {50k} & 21.8 & 20.3 & 82.6 & 38.6 & 65.2 & {45.70} \\
        W4A16KV8 & {\no} & {\no} & {50k} & 0.0 & 0.0 & 0.0 & 0.0 & 0.0 & {0.00}  \\ 
        W4A16KV8 & {\no} & {\yes} & {50k} & 17.3 & 6.3 & 75.6 & 33.0 & \textbf{64.0} & {39.25} \\ 
        W4A16KV8 & {\yes} & {\yes} & {50k} & \textbf{23.3} & \textbf{15.0} & \textbf{79.6} & \textbf{33.7} & 57.0 & \textbf{41.72} \\
        \hline\\[-1.44em]
        \nf{BF16} & {\na} & {\na} & {1.2M} & 53.3 & 33.0 & 94.0 & 39.9 & 82.5 & {60.54} \\
        {W4A16KV8} & {\yes} & {\yes} & {1.2M} & \textbf{46.6} & \textbf{36.6} & \textbf{89.6} & \textbf{37.8} & \textbf{80.0} & \textbf{58.12}  \\ 
        \bottomrule
    \end{tabular}
    \renewcommand{\arraystretch}{1}
\end{table}

\subsection{Verifier Quantization and On-Device Deployment}\label{sec:quantization_on_device}

As the final stage of our quantization pipeline, we address the quantization of the verifier, which is essential for deploying the proposed verifier on resource-constrained hardware.
To minimize degradation from reduced numerical precision, we train the verifier directly on embeddings produced by the 4-bit weight-quantized Qwen-2.5-7B-Instruct model obtained in section~\ref{sec:quantization_base}.
This choice reduces distribution shift between training and inference, ensuring that the verifier learns to operate on similar representations it will encounter at deployment time.
After training the verifier head under this setting, we further quantize both activations and verifier head weights to 8 bit representations using FastForward~\citep{fastforward}.

Once all components including the base model, reasoning adapters, and verifier are quantized, we prepare the models for on-device deployment. The first step after quantization is the model transformation to assure compatibility at pytorch representation level with the format supported by GENIE SDK \cite{GENIE}. These pertains to aspects of autoregressive parallel and sequential generation for prefill and decoding, as well as handling the attention operations and masking, and position embeddings. 

Next, we establish compatibility with GENIE at the ONNX representation level. We use Qualcomm FastForward \cite{fastforward} for this stage and implement transformations at linear layers and multi-head attention, as well as model partitioning. We use Pytorch with FastForward to get the ONNX graph and the associated quantization encodings. We export to Deep Learning Container format, and Quantize any remaining non-quantized nodes missed by FastForward (e.g. biases). We compile for the deployment target (e.g. aarch64-android) and upload to the device using \textit{adb} (Android Device Bridge). 
\section{Discussions and Challenges}

Deploying capable reasoning models on resource-constrained edge devices requires navigating a complex trade-off between task performance, latency, memory footprint, and power consumption. In this work, we proposed a practical end-to-end framework to overcome these limitations. By decoupling reasoning from the base weights using modular LoRA adapters, we demonstrate that parameter-efficient fine-tuning can achieve competitive reasoning accuracy relative to computationally expensive full-parameter distillation methods like DeepSeek-R1-Distill. To optimize day-to-day user interactions, our lightweight dynamic Switcher routes standard conversations to the highly efficient base model, reserving the reasoning adapters strictly for complex queries where multi-step logic is required. To further combat latency, our budget-forced RL alignment explicitly penalizes generation verbosity, yielding a 2.4$\times$ reduction in average reasoning tokens without sacrificing task accuracy. At inference time, we exploit the memory-bound nature of autoregressive decoding by introducing parallel test-time scaling, coupled with a lightweight verifier that provides up to a 10\% accuracy boost on complex reasoning benchmarks. Finally, we show that 4-bit weight quantization via FPTQuant and Quantization-Aware Modular Reasoning preserves this robust performance, achieving within 2\% of the full-precision reasoning model's accuracy while delivering massive memory savings. Below, we summarize the key insights derived from each component of our pipeline and outline the remaining challenges that pave the way for future research.

\paragraph{LoRA for modular reasoning.}
Our experiments revealed that parameter-efficient fine-tuning via LoRA is highly effective at eliciting reasoning capabilities in small (3B and 7B) base models, often rivaling the performance of computationally expensive full-parameter distillation. A key insight is that the success of LoRA is heavily dependent on adapter capacity and base model scale. While a LoRA rank of 128 allowed the 7B model to nearly match dense fine-tuning baselines on challenging benchmarks, the 3B model exhibited a wider performance gap, indicating that smaller backbones are more sensitive to adapter capacity limits. Furthermore, while reasoning specialization improves performance on complex tasks (e.g., LiveCodeBench), it introduces a trade-off, occasionally degrading zero-shot performance on simpler coding tasks that require direct answers. Managing this specialization-forgetting trade-off remains an ongoing challenge.

\paragraph{Dynamic LoRA routing via the switcher.}
We demonstrated that the computational overhead of reasoning can be drastically reduced by recognizing that not all queries require complex multi-step logic. The lightweight Switcher module successfully acts as an on-demand router, preserving the base model's speed for standard queries while activating reasoning LoRA adapters only when necessary. A major deployment insight was the necessity of \textit{masked LoRA training} during the prefill stage. This strategy ensures that the KV-cache generated by the base model can be seamlessly reused by the reasoning adapters, completely eliminating the severe latency penalty of re-encoding prompt tokens when switching modes. 

While the current switcher relies on a supervised classifier head to evaluate query complexity, a promising direction for future work is to learn this routing policy via reinforcement learning. This would act synergistically with our budget-forcing objectives: while budget-forced RL explicitly shortens the reasoning traces when LoRA is active, an RL-driven router optimized for both accuracy and length would learn to bypass the adapters entirely whenever possible. Because the frozen base model natively produces direct answers without verbose chain-of-thought, successfully routing a query to the non-LoRA mode automatically yields a drastically shorter response, organically reserving the reasoning adapters strictly for complex queries where the base model would fail.

Future work could extend the switcher beyond binary routing and turn it into a general mechanism for dynamic LoRA selection. Instead of choosing only between the base model and a single reasoning adapter, the system could route each query to a bank of task-specific adapters \cite{huang2024mixture, feng2024mixture}, for example specialized for mathematics, coding, or other domains, allowing the same backbone to support richer capabilities while preserving modular deployment. An especially promising direction is to include adapters tailored for latent reasoning, since recent work suggests that latent-reasoning LoRA adapters can retain strong reasoning performance while substantially reducing the token overhead of explicit CoT generation \cite{hao2024trainingllms, shen2025codi, wu2025parallel, kuzina2026kava, wang2025system}. In this setting, the switcher would not only decide whether reasoning is needed, but also which form of reasoning is most efficient for a given query, making dynamic adapter routing a natural path toward more capable and more compute-efficient on-device systems.

\paragraph{Budget forcing.}
We presented our RL recipe to finetune LLMs to generate shorter completions. By leveraging a multiplicative penalty (eq. \ref{eqn:bf_reward}), we successfully aligned LLMs to reduce the number of generated tokens to answers given questions. Our empirical evaluations demonstrate an average completion length reduction of $2.4\times$-and up to $8\times$ maximum compression-with minimal degradation in task accuracy. 
Crucially, our formulation structurally mitigates the reward hacking we observed in early experiments, ensuring that efficiency gains stem from genuine rationale compression rather than formatting exploits. 
While our ``soft-barrier'' approach offers a robust mechanism for compute-aware inference, it also opens several exciting avenues for future research. We outline these forward-looking directions, which naturally address the current boundaries of our methodology.

Our experiments provide a snapshot of budget forcing on state-of-the-art reasoning models. However, the relationship between base model scale (e.g., parameter count) and the capacity for rationale compression remains an open question. Investigating whether larger, more capable models exhibit greater ``epistemic hesitation'', and thus offer a larger margin for ITC reduction, will be critical for formulating generalized scaling laws for budget forcing.

A fundamental limitation of current budget forcing approaches, including ours, is the assumption of uniform token cost where every generated token contributes equally to the budget consumption regardless of its utility. However, a token representing a crucial logical leap carries significantly higher semantic value than a token used for syntactic glue or hedging (e.g., ``Let me think about this...''). A promising future direction lies in developing \textit{semantic-aware} budget priors that weight penalties dynamically based on information density or local entropy as explored in \cite{Massoli2026-eu}. By shifting the optimization objective from pure length minimization to \textit{reasoning density maximization}, we can encourage models to prioritize high-utility tokens while disproportionately penalizing low-entropy filler, effectively decoupling computational cost from reasoning depth.

\paragraph{Parallel test-time scaling and reasoning.} We have demonstrated that even a light verifier design can substantially improve the effectiveness of parallel test-time scaling for reasoning on edge devices. By combining the robustness of aggregation with a learned notion of correctness at negligible additional cost, weighted majority voting offers a practical and efficient approach for deploying parallel reasoning on resource-constrained devices. The current design can be extended to score the steps with a process reward model. Another interesting extension is the parallel reasoning schemes with interdependent generation as in \cite{hsu2025group,rodionov2025hogwild,cesa2026lanerope}.

\paragraph{Quantization}

Our experiments on Qwen2.5-7B-Instruct highlight the importance of starting from a strong quantized base model.
Leveraging function-preserving transformations, improved range initialization, and joint fine-tuning of transformation and quantization parameters offers a straightforward and lightweight path to quantizing modern LLMs while preserving robust predictive performance.
When extending such models with reasoning capabilities, we further show that it is essential to account for distribution shifts induced by quantization.
Inspired by prior PEFT literature, we address this challenge by proposing the Quantization-Aware Modular Reasoning (QAMR) approach, %
which mitigates the quantization noise and distribution shifts by training reasoning adapters directly on the quantized base model.
Finally, our results demonstrate that a relatively compact 4-bit weight-quantized quantized 7B model can achieve reasoning performance comparable to that of substantially larger models.

Reasoning remains a token-generation-intensive task, making it fundamentally memory-bound rather than compute-bound.
As a result, further improvements in efficiency and performance will likely depend on reducing the memory footprint of the model weights.
A promising direction for future work is to push quantization below 4 bits by leveraging state-of-the-art compression techniques such as Quip\#~\citep{tseng2024quip}, or exploring 2-3-bit QAT methods, such as ParetoQ~\citep{liu_paretoq_2025}.

\section{Conclusion}
In this work, we presented an end-to-end framework that makes state-of-the-art LLM reasoning practical on resource-constrained edge devices. We demonstrated that parameter-efficient LoRA adaptation, governed by a dynamic routing switcher, unlocks powerful reasoning capabilities without compromising the speed of everyday interactions. By introducing budget-forced reinforcement learning, we successfully curbed model verbosity to fit strict on-device token limits. To further maximize hardware utilization, we leveraged parallel test-time scaling and a lightweight latent verifier to boost accuracy during the memory-bound generation phase. Throughout this pipeline, hardware-awareness remains the unifying principle: from maximizing KV-cache reuse to intertwining quantization directly into the training of the adapters, switcher, and verifier. Ultimately, this co-designed approach bridges the gap between cloud-based reasoning and the strict memory, latency, and power budgets of mobile hardware, providing a practical blueprint for on-device AI.

\clearpage
\newpage
\bibliographystyle{unsrtnat}
\bibliography{bibliography}

\begin{thebibliography}{164}
\providecommand{\natexlab}[1]{#1}
\providecommand{\url}[1]{\texttt{#1}}
\expandafter\ifx\csname urlstyle\endcsname\relax
  \providecommand{\doi}[1]{doi: #1}\else
  \providecommand{\doi}{doi: \begingroup \urlstyle{rm}\Url}\fi

\bibitem[Anthropic(2024)]{anthropic2024claude}
Anthropic.
\newblock The claude 3 model family: Opus, sonnet, haiku.
\newblock \emph{Claude-3 Model Card}, 1\penalty0 (1):\penalty0 4, 2024.

\bibitem[Jaech et~al.(2024)Jaech, Kalai, Lerer, Richardson, El-Kishky, Low, Helyar, Madry, Beutel, Carney, et~al.]{jaech2024openai}
Aaron Jaech, Adam Kalai, Adam Lerer, Adam Richardson, Ahmed El-Kishky, Aiden Low, Alec Helyar, Aleksander Madry, Alex Beutel, Alex Carney, et~al.
\newblock Openai o1 system card.
\newblock \emph{arXiv preprint arXiv:2412.16720}, 2024.

\bibitem[Abouzaid et~al.(2026)Abouzaid, Blumberg, Hairer, Kileel, Kolda, Nelson, Spielman, Srivastava, Ward, Weinberger, and Williams]{Abouzaid2026-first-proof}
Mohammed Abouzaid, Andrew~J Blumberg, Martin Hairer, Joe Kileel, Tamara~G Kolda, Paul~D Nelson, Daniel Spielman, Nikhil Srivastava, Rachel Ward, Shmuel Weinberger, and Lauren Williams.
\newblock First proof.
\newblock \emph{arXiv preprint arXiv:2602.05192}, 5~February 2026.

\bibitem[Cao et~al.(2025)Cao, Chen, Guo, Song, Yue, Zhang, and Zhao]{Cao2025-ai4math}
Yang Cao, Yubin Chen, Xuyang Guo, Zhao Song, Song Yue, Jiahao Zhang, and Jiale Zhao.
\newblock Evaluating frontier {LLMs} on {PhD}-level mathematical reasoning: A benchmark on a textbook in theoretical computer science about randomized algorithms.
\newblock \emph{arXiv [cs.AI]}, 16~December 2025.

\bibitem[Feng et~al.(2026)Feng, Trinh, Bingham, Hwang, Chervonyi, Jung, Lee, Pagano, Kim, Pasqualotto, Gukov, Lee, Kim, Hou, Ghiasi, Tay, Li, Kuang, Liu, Lin, Liu, Nayakanti, Yang, Cheng, Hassabis, Kavukcuoglu, Le, and Luong]{Feng2026-ai4math}
Tony Feng, Trieu~H Trinh, Garrett Bingham, Dawsen Hwang, Yuri Chervonyi, Junehyuk Jung, Joonkyung Lee, Carlo Pagano, Sang-Hyun Kim, Federico Pasqualotto, Sergei Gukov, Jonathan~N Lee, Junsu Kim, Kaiying Hou, Golnaz Ghiasi, Yi~Tay, Yaguang Li, Chenkai Kuang, Yuan Liu, Hanzhao Lin, Evan~Zheran Liu, Nigamaa Nayakanti, Xiaomeng Yang, Heng-Tze Cheng, Demis Hassabis, Koray Kavukcuoglu, Quoc~V Le, and Thang Luong.
\newblock Towards autonomous mathematics research.
\newblock \emph{arXiv preprint arXiv:2602.10177}, 12~February 2026.

\bibitem[Woodruff et~al.(2026)Woodruff, Cohen-Addad, Jain, Mao, Zuo, Bateni, Branzei, Brenner, Chen, Feng, Fortnow, Fu, Guan, Hadizadeh, Hajiaghayi, JafariRaviz, Javanmard, C.~S., Kawarabayashi, Kumar, Lattanzi, Lee, Li, Panageas, Paparas, Przybocki, Subercaseaux, Svensson, Taherijam, Wu, Yogev, Zadimoghaddam, Zhou, Matias, Manyika, and Mirrokni]{Woodruff2026-gemini}
David~P Woodruff, Vincent Cohen-Addad, Lalit Jain, Jieming Mao, Song Zuo, Mohammadhossein Bateni, Simina Branzei, Michael~P Brenner, Lin Chen, Ying Feng, Lance Fortnow, Gang Fu, Ziyi Guan, Zahra Hadizadeh, Mohammad~T Hajiaghayi, Mahdi JafariRaviz, Adel Javanmard, Karthik C.~S., Ken-Ichi Kawarabayashi, Ravi Kumar, Silvio Lattanzi, Euiwoong Lee, Yi~Li, Ioannis Panageas, Dimitris Paparas, Benjamin Przybocki, Bernardo Subercaseaux, Ola Svensson, Shayan Taherijam, Xuan Wu, Eylon Yogev, Morteza Zadimoghaddam, Samson Zhou, Yossi Matias, James Manyika, and Vahab Mirrokni.
\newblock Accelerating scientific research with gemini: Case studies and common techniques.
\newblock \emph{arXiv preprint arXiv:2602.03837}, 16~February 2026.

\bibitem[Yao et~al.(2022{\natexlab{a}})Yao, Zhao, Yu, Du, Shafran, Narasimhan, and Cao]{Yao2022-react}
Shunyu Yao, Jeffrey Zhao, Dian Yu, Nan Du, Izhak Shafran, Karthik~R Narasimhan, and Yuan Cao.
\newblock {ReAct}: Synergizing reasoning and acting in language models.
\newblock In \emph{The Eleventh International Conference on Learning Representations}, 29~September 2022{\natexlab{a}}.

\bibitem[Zhou et~al.(2025)Zhou, Zhang, Tong, Zhang, Chen, Kong, Cai, Liu, Wang, Zhou, and Hoi]{Zhou2025-maiui}
Hanzhang Zhou, Xu~Zhang, Panrong Tong, Jianan Zhang, Liangyu Chen, Quyu Kong, Chenglin Cai, Chen Liu, Yue Wang, Jingren Zhou, and Steven Hoi.
\newblock {MAI}-{UI} technical report: Real-world centric foundation {GUI} agents.
\newblock \emph{arXiv [cs.CV]}, 26~December 2025.

\bibitem[Wang et~al.(2025{\natexlab{a}})Wang, Zou, Song, Feng, Fang, Lu, Liu, Luo, Liang, Huang, Zhong, Ye, Qin, Xiong, Song, Wu, Li, Li, Dun, Liu, Zan, Leng, Wang, Yu, Chen, Guo, Su, Huang, Shen, Shi, Yan, Zhao, Liu, Ye, Zheng, Xin, Zhao, Heng, Huang, Wang, Qin, Lin, Wu, Chen, Wang, Zhong, Zhang, Li, Li, Zhao, Jiang, Wu, Zhou, Pang, Han, Liu, Ma, Liu, Cai, Fu, Liu, Wang, Zhang, Zhou, Li, Shi, Yang, Tang, Li, Han, Lu, Lin, Tong, Li, Zhang, Miao, Jiang, Li, Zhao, Li, Ma, Lin, Zhang, Yang, Guo, Zhu, Liu, Du, Cai, Li, Yuan, Han, Wang, Guo, Cheng, Ma, Xiao, Huang, Chen, Du, Chen, Wang, Li, Yang, Zeng, Jin, Li, Chen, Chen, Chen, Zhao, and Shi]{Wang2025-uitars2}
Haoming Wang, Haoyang Zou, Huatong Song, Jiazhan Feng, Junjie Fang, Junting Lu, Longxiang Liu, Qinyu Luo, Shihao Liang, Shijue Huang, Wanjun Zhong, Yining Ye, Yujia Qin, Yuwen Xiong, Yuxin Song, Zhiyong Wu, Aoyan Li, Bo~Li, Chen Dun, Chong Liu, Daoguang Zan, Fuxing Leng, Hanbin Wang, Hao Yu, Haobin Chen, Hongyi Guo, Jing Su, Jingjia Huang, Kai Shen, Kaiyu Shi, Lin Yan, Peiyao Zhao, Pengfei Liu, Qinghao Ye, Renjie Zheng, Shulin Xin, Wayne~Xin Zhao, Wen Heng, Wenhao Huang, Wenqian Wang, Xiaobo Qin, Yi~Lin, Youbin Wu, Zehui Chen, Zihao Wang, Baoquan Zhong, Xinchun Zhang, Xujing Li, Yuanfan Li, Zhongkai Zhao, Chengquan Jiang, Faming Wu, Haotian Zhou, Jinlin Pang, Li~Han, Qi~Liu, Qianli Ma, Siyao Liu, Songhua Cai, Wenqi Fu, Xin Liu, Yaohui Wang, Zhi Zhang, Bo~Zhou, Guoliang Li, Jiajun Shi, Jiale Yang, Jie Tang, Li~Li, Qihua Han, Taoran Lu, Woyu Lin, Xiaokang Tong, Xinyao Li, Yichi Zhang, Yu~Miao, Zhengxuan Jiang, Zili Li, Ziyuan Zhao, Chenxin Li, Dehua Ma, Feng Lin, Ge~Zhang, Haihua Yang, Hangyu Guo, Hongda Zhu,
  Jiaheng Liu, Junda Du, Kai Cai, Kuanye Li, Lichen Yuan, Meilan Han, Minchao Wang, Shuyue Guo, Tianhao Cheng, Xiaobo Ma, Xiaojun Xiao, Xiaolong Huang, Xinjie Chen, Yidi Du, Yilin Chen, Yiwen Wang, Zhaojian Li, Zhenzhu Yang, Zhiyuan Zeng, Chaolin Jin, Chen Li, Hao Chen, Haoli Chen, Jian Chen, Qinghao Zhao, and Guang Shi.
\newblock {UI}-{TARS}-2 technical report: Advancing {GUI} agent with multi-turn reinforcement learning.
\newblock \emph{arXiv [cs.AI]}, 5~September 2025{\natexlab{a}}.

\bibitem[{Venus Team} et~al.(2026){Venus Team}, Gao, Gu, Liu, Qiu, Shen, Wen, Xia, Xu, Zeng, et~al.]{Venus-Team2026-uivenus1.5}
{Venus Team}, Changlong Gao, Zhangxuan Gu, Yulin Liu, Xinyu Qiu, Shuheng Shen, Yue Wen, Tianyu Xia, Zhenyu Xu, Zhengwen Zeng, et~al.
\newblock {UI}-venus-1.5 technical report.
\newblock \emph{arXiv [cs.CV]}, 9~February 2026.

\bibitem[thi(2025)]{thinking-efficiency2025}
Measuring thinking efficiency in reasoning models: The missing benchmark.
\newblock \url{https://nousresearch.com/measuring-thinking-efficiency-in-reasoning-models-the-missing-benchmark/}, 14~August 2025.
\newblock Accessed: 2026-2-19.

\bibitem[Alizadeh et~al.(2024)Alizadeh, Mirzadeh, Belenko, Khatamifard, Cho, Mundo, Rastegari, and Farajtabar]{Alizadeh2024-llm-flash}
Keivan Alizadeh, Iman Mirzadeh, Dmitry Belenko, S~Karen Khatamifard, Minsik Cho, Carlo C~Del Mundo, Mohammad Rastegari, and Mehrdad Farajtabar.
\newblock {LLM} in a flash: Efficient large language model inference with limited memory.
\newblock In \emph{ACL}, 2024.

\bibitem[Xiao et~al.(2026)Xiao, Huang, Chen, and Tian]{Xiao2026-llm-mobile}
Jie Xiao, Qianyi Huang, Xu~Chen, and Chen Tian.
\newblock Understanding large language models in your pockets: Performance study on {COTS} mobile devices.
\newblock \emph{arXiv [cs.LG]}, 9~February 2026.

\bibitem[Hu et~al.(2022)Hu, Shen, Wallis, Allen-Zhu, Li, Wang, Wang, and Chen]{hulora}
Edward~J Hu, Yelong Shen, Phillip Wallis, Zeyuan Allen-Zhu, Yuanzhi Li, Shean Wang, Lu~Wang, and Weizhu Chen.
\newblock Lora: Low-rank adaptation of large language models.
\newblock In \emph{International Conference on Learning Representations}, 2022.

\bibitem[DeepSeek-AI et~al.(2025)DeepSeek-AI, Guo, Yang, Zhang, Song, Wang, Zhu, Xu, Zhang, Ma, and et~al.]{deepseekai2025deepseekr1}
DeepSeek-AI, Daya Guo, Dejian Yang, Haowei Zhang, Junxiao Song, Peiyi Wang, Qihao Zhu, Runxin Xu, Ruoyu Zhang, Shirong Ma, and et~al.
\newblock Deepseek-r1: Incentivizing reasoning capability in llms via reinforcement learning, 2025.
\newblock URL \url{https://arxiv.org/abs/2501.12948}.

\bibitem[Muennighoff et~al.(2025)Muennighoff, Yang, Shi, Li, Fei-Fei, Hajishirzi, Zettlemoyer, Liang, Cand\`es, and Hashimoto]{muennighoff2025s1}
Niklas Muennighoff, Zitong Yang, Weijia Shi, Xiang~Lisa Li, Li~Fei-Fei, Hannaneh Hajishirzi, Luke Zettlemoyer, Percy Liang, Emmanuel Cand\`es, and Tatsunori Hashimoto.
\newblock s1: Simple test-time scaling.
\newblock \emph{arXiv preprint arXiv:2501.19393}, 2025.
\newblock \doi{10.48550/arXiv.2501.19393}.
\newblock URL \url{https://arxiv.org/abs/2501.19393}.

\bibitem[Li et~al.(2025{\natexlab{a}})Li, Zhao, Zhang, and Gan]{li2025steering}
Junyan Li, Wenshuo Zhao, Yang Zhang, and Chuang Gan.
\newblock Steering llm thinking with budget guidance.
\newblock \emph{arXiv preprint arXiv:2506.13752}, 2025{\natexlab{a}}.

\bibitem[fas()]{fastforward}
Fastforward: Neural network quantization for research and prototyping.
\newblock \url{https://github.com/Qualcomm-AI-research/fastforward}.

\bibitem[GEN()]{GENIE}
Qualcomm gen ai inference extensions (genie).
\newblock \url{https://www.qualcomm.com/developer/software/gen-ai-inference-extensions}.

\bibitem[Kojima et~al.(2022)Kojima, Gu, Reid, Matsuo, and Iwasawa]{kojima2022large}
Takeshi Kojima, Shixiang~Shane Gu, Machel Reid, Yutaka Matsuo, and Yusuke Iwasawa.
\newblock Large language models are zero-shot reasoners.
\newblock \emph{Advances in neural information processing systems}, 35:\penalty0 22199--22213, 2022.

\bibitem[Nye et~al.(2022)Nye, Andreassen, Gur-Ari, Michalewski, Austin, Bieber, Dohan, Lewkowycz, Bosma, Luan, Sutton, and Odena]{nye2022show}
Maxwell Nye, Anders~Johan Andreassen, Guy Gur-Ari, Henryk Michalewski, Jacob Austin, David Bieber, David Dohan, Aitor Lewkowycz, Maarten Bosma, David Luan, Charles Sutton, and Augustus Odena.
\newblock Show your work: Scratchpads for intermediate computation with language models, 2022.
\newblock URL \url{https://openreview.net/forum?id=iedYJm92o0a}.

\bibitem[Wei et~al.(2022{\natexlab{a}})Wei, Wang, Schuurmans, Bosma, Ichter, Xia, Chi, Le, and Zhou]{wei2022chain}
Jason Wei, Xuezhi Wang, Dale Schuurmans, Maarten Bosma, Brian Ichter, Fei Xia, Ed~Chi, Quoc~V Le, and Denny Zhou.
\newblock Chain-of-thought prompting elicits reasoning in large language models.
\newblock In S.~Koyejo, S.~Mohamed, A.~Agarwal, D.~Belgrave, K.~Cho, and A.~Oh, editors, \emph{Advances in Neural Information Processing Systems}, volume~35, pages 24824--24837. Curran Associates, Inc., 2022{\natexlab{a}}.
\newblock URL \url{https://proceedings.neurips.cc/paper_files/paper/2022/file/9d5609613524ecf4f15af0f7b31abca4-Paper-Conference.pdf}.

\bibitem[Wang et~al.(2025{\natexlab{b}})Wang, Asilis, Akg{\"u}l, Bilgin, Liu, and Neiswanger]{wang2025tina}
Shangshang Wang, Julian Asilis, {\"O}mer~Faruk Akg{\"u}l, Enes~Burak Bilgin, Ollie Liu, and Willie Neiswanger.
\newblock Tina: Tiny reasoning models via lora.
\newblock \emph{arXiv preprint arXiv:2504.15777}, 2025{\natexlab{b}}.

\bibitem[Xu et~al.(2025{\natexlab{a}})Xu, Peng, Awadalla, Chen, Chen, Gao, Kim, Li, Ren, Shen, et~al.]{xu2025phi}
Haoran Xu, Baolin Peng, Hany Awadalla, Dongdong Chen, Yen-Chun Chen, Mei Gao, Young~Jin Kim, Yunsheng Li, Liliang Ren, Yelong Shen, et~al.
\newblock Phi-4-mini-reasoning: Exploring the limits of small reasoning language models in math.
\newblock \emph{arXiv preprint arXiv:2504.21233}, 2025{\natexlab{a}}.

\bibitem[Jiang et~al.(2025)Jiang, Wu, Huang, Dong, Chi, Dong, Zhang, Lv, Cui, and Wei]{jiang2025think}
Lingjie Jiang, Xun Wu, Shaohan Huang, Qingxiu Dong, Zewen Chi, Li~Dong, Xingxing Zhang, Tengchao Lv, Lei Cui, and Furu Wei.
\newblock Think only when you need with large hybrid-reasoning models.
\newblock \emph{arXiv preprint arXiv:2505.14631}, 2025.

\bibitem[Schulman and Lab(2025)]{schulman2025lora}
John Schulman and Thinking~Machines Lab.
\newblock Lora without regret, 2025.
\newblock https://thinkingmachines.ai/blog/lora/.

\bibitem[Chen et~al.(2025{\natexlab{a}})Chen, Yang, Liu, Lee, Xu, Shoeybi, Catanzaro, and Ping]{chen2025acereason}
Yang Chen, Zhuolin Yang, Zihan Liu, Chankyu Lee, Peng Xu, Mohammad Shoeybi, Bryan Catanzaro, and Wei Ping.
\newblock Acereason-nemotron: Advancing math and code reasoning through reinforcement learning.
\newblock \emph{arXiv preprint arXiv:2505.16400}, 2025{\natexlab{a}}.

\bibitem[Dang and Ngo(2025)]{dang2025reinforcement}
Quy-Anh Dang and Chris Ngo.
\newblock Reinforcement learning for reasoning in small llms: What works and what doesn't.
\newblock \emph{arXiv preprint arXiv:2503.16219}, 2025.

\bibitem[Alomrani et~al.(2025)Alomrani, Zhang, Li, Sun, Pal, Zhang, Hu, Ajwani, Valkanas, Karimi, et~al.]{alomrani2025reasoning}
Mohammad~Ali Alomrani, Yingxue Zhang, Derek Li, Qianyi Sun, Soumyasundar Pal, Zhanguang Zhang, Yaochen Hu, Rohan~Deepak Ajwani, Antonios Valkanas, Raika Karimi, et~al.
\newblock Reasoning on a budget: A survey of adaptive and controllable test-time compute in llms.
\newblock \emph{arXiv preprint arXiv:2507.02076}, 2025.

\bibitem[Shao et~al.(2024{\natexlab{a}})Shao, Wang, Zhu, Xu, Song, Bi, Zhang, Zhang, Li, Wu, et~al.]{shao2024deepseekmath}
Zhihong Shao, Peiyi Wang, Qihao Zhu, Runxin Xu, Junxiao Song, Xiao Bi, Haowei Zhang, Mingchuan Zhang, YK~Li, Yang Wu, et~al.
\newblock Deepseekmath: Pushing the limits of mathematical reasoning in open language models.
\newblock \emph{arXiv preprint arXiv:2402.03300}, 2024{\natexlab{a}}.

\bibitem[Team(2025)]{QwQ-32B}
Qwen Team.
\newblock Qwq-32b: Embracing the power of reinforcement learning.
\newblock \url{https://qwen.ai/blog?id=qwq-32b}, 2025.

\bibitem[Guha et~al.(2025)Guha, Marten, Keh, Raoof, Smyrnis, Bansal, Nezhurina, Mercat, Vu, Sprague, et~al.]{guha2025openthoughts}
Etash Guha, Ryan Marten, Sedrick Keh, Negin Raoof, Georgios Smyrnis, Hritik Bansal, Marianna Nezhurina, Jean Mercat, Trung Vu, Zayne Sprague, et~al.
\newblock Openthoughts: Data recipes for reasoning models.
\newblock \emph{arXiv preprint arXiv:2506.04178}, 2025.

\bibitem[Qwen et~al.(2025)Qwen, Yang, Yang, Zhang, Hui, Zheng, Yu, Li, Liu, Huang, Wei, Lin, Yang, Tu, Zhang, Yang, Yang, Zhou, Lin, Dang, Lu, Bao, Yang, Yu, Li, Xue, Zhang, Zhu, Men, Lin, Li, Tang, Xia, Ren, Ren, Fan, Su, Zhang, Wan, Liu, Cui, Zhang, and Qiu]{qwen2025qwen25technicalreport}
Qwen, An~Yang, Baosong Yang, Beichen Zhang, Binyuan Hui, Bo~Zheng, Bowen Yu, Chengyuan Li, Dayiheng Liu, Fei Huang, Haoran Wei, Huan Lin, Jian Yang, Jianhong Tu, Jianwei Zhang, Jianxin Yang, Jiaxi Yang, Jingren Zhou, Junyang Lin, Kai Dang, Keming Lu, Keqin Bao, Kexin Yang, Le~Yu, Mei Li, Mingfeng Xue, Pei Zhang, Qin Zhu, Rui Men, Runji Lin, Tianhao Li, Tianyi Tang, Tingyu Xia, Xingzhang Ren, Xuancheng Ren, Yang Fan, Yang Su, Yichang Zhang, Yu~Wan, Yuqiong Liu, Zeyu Cui, Zhenru Zhang, and Zihan Qiu.
\newblock Qwen2.5 technical report, 2025.
\newblock URL \url{https://arxiv.org/abs/2412.15115}.

\bibitem[Face(2025)]{openr1}
Hugging Face.
\newblock Open r1: A fully open reproduction of deepseek-r1, January 2025.
\newblock URL \url{https://github.com/huggingface/open-r1}.

\bibitem[aim(2024)]{aime}
Art of problem solving. aime problems and solutions.
\newblock 2024.
\newblock URL \url{https://artofproblemsolving.com/wiki/index.php/AIME_Problems_and_Solutions}.

\bibitem[amc(2023)]{amc23}
Art of problem solving. amc problems and solutions.
\newblock 2023.
\newblock URL \url{https://artofproblemsolving.com/wiki/index.php/AMC_12_Problems_and_Solutions}.

\bibitem[Hendrycks et~al.(2021)Hendrycks, Burns, Kadavath, Arora, Basart, Tang, Song, and Steinhardt]{hendrycks2021measuring}
Dan Hendrycks, Collin Burns, Saurav Kadavath, Akul Arora, Steven Basart, Eric Tang, Dawn Song, and Jacob Steinhardt.
\newblock Measuring mathematical problem solving with the math dataset.
\newblock \emph{arXiv preprint arXiv:2103.03874}, 2021.

\bibitem[Rein et~al.(2024)Rein, Hou, Stickland, Petty, Pang, Dirani, Michael, and Bowman]{rein2024gpqa}
David Rein, Betty~Li Hou, Asa~Cooper Stickland, Jackson Petty, Richard~Yuanzhe Pang, Julien Dirani, Julian Michael, and Samuel~R Bowman.
\newblock Gpqa: A graduate-level google-proof q\&a benchmark.
\newblock In \emph{First Conference on Language Modeling}, 2024.

\bibitem[Jain et~al.(2024)Jain, Han, Gu, Li, Yan, Zhang, Wang, Solar-Lezama, Sen, and Stoica]{livecodebench}
Naman Jain, King Han, Alex Gu, Wen-Ding Li, Fanjia Yan, Tianjun Zhang, Sida Wang, Armando Solar-Lezama, Koushik Sen, and Ion Stoica.
\newblock {LiveCodeBench}: Holistic and contamination free evaluation of large language models for code.
\newblock \emph{arXiv [cs.SE]}, March 2024.

\bibitem[Chen et~al.(2021)Chen, Tworek, Jun, Yuan, de~Oliveira~Pinto, Kaplan, Edwards, Burda, Joseph, Brockman, Ray, Puri, Krueger, Petrov, Khlaaf, Sastry, Mishkin, Chan, Gray, Ryder, Pavlov, Power, Kaiser, Bavarian, Winter, Tillet, Such, Cummings, Plappert, Chantzis, Barnes, Herbert-Voss, Guss, Nichol, Paino, Tezak, Tang, Babuschkin, Balaji, Jain, Saunders, Hesse, Carr, Leike, Achiam, Misra, Morikawa, Radford, Knight, Brundage, Murati, Mayer, Welinder, McGrew, Amodei, McCandlish, Sutskever, and Zaremba]{humaneval}
Mark Chen, Jerry Tworek, Heewoo Jun, Qiming Yuan, Henrique~Ponde de~Oliveira~Pinto, Jared Kaplan, Harri Edwards, Yuri Burda, Nicholas Joseph, Greg Brockman, Alex Ray, Raul Puri, Gretchen Krueger, Michael Petrov, Heidy Khlaaf, Girish Sastry, Pamela Mishkin, Brooke Chan, Scott Gray, Nick Ryder, Mikhail Pavlov, Alethea Power, Lukasz Kaiser, Mohammad Bavarian, Clemens Winter, Philippe Tillet, Felipe~Petroski Such, Dave Cummings, Matthias Plappert, Fotios Chantzis, Elizabeth Barnes, Ariel Herbert-Voss, William~Hebgen Guss, Alex Nichol, Alex Paino, Nikolas Tezak, Jie Tang, Igor Babuschkin, Suchir Balaji, Shantanu Jain, William Saunders, Christopher Hesse, Andrew~N Carr, Jan Leike, Josh Achiam, Vedant Misra, Evan Morikawa, Alec Radford, Matthew Knight, Miles Brundage, Mira Murati, Katie Mayer, Peter Welinder, Bob McGrew, Dario Amodei, Sam McCandlish, Ilya Sutskever, and Wojciech Zaremba.
\newblock Evaluating large language models trained on code.
\newblock \emph{arXiv [cs.LG]}, July 2021.

\bibitem[Austin et~al.(2021)Austin, Odena, Nye, Bosma, Michalewski, Dohan, Jiang, Cai, Terry, Le, and Sutton]{mbpp}
Jacob Austin, Augustus Odena, Maxwell Nye, Maarten Bosma, Henryk Michalewski, David Dohan, Ellen Jiang, Carrie Cai, Michael Terry, Quoc Le, and Charles Sutton.
\newblock Program synthesis with large language models.
\newblock \emph{arXiv [cs.PL]}, August 2021.

\bibitem[Liu et~al.(2023{\natexlab{a}})Liu, Xia, Wang, and Zhang]{evalplus}
Jiawei Liu, Chunqiu~Steven Xia, Yuyao Wang, and Lingming Zhang.
\newblock Is your code generated by {ChatGPT} really correct? rigorous evaluation of large language models for code generation.
\newblock \emph{arXiv [cs.SE]}, May 2023{\natexlab{a}}.

\bibitem[Habib et~al.(2023)Habib, Fourrier, Kydlíček, Wolf, and Tunstall]{lighteval}
Nathan Habib, Clémentine Fourrier, Hynek Kydlíček, Thomas Wolf, and Lewis Tunstall.
\newblock Lighteval: A lightweight framework for llm evaluation, 2023.
\newblock URL \url{https://github.com/huggingface/lighteval}.

\bibitem[Kwon et~al.(2023)Kwon, Li, Zhuang, Sheng, Zheng, Yu, Gonzalez, Zhang, and Stoica]{kwon2023efficient}
Woosuk Kwon, Zhuohan Li, Siyuan Zhuang, Ying Sheng, Lianmin Zheng, Cody~Hao Yu, Joseph~E. Gonzalez, Hao Zhang, and Ion Stoica.
\newblock Efficient memory management for large language model serving with pagedattention.
\newblock In \emph{Proceedings of the ACM SIGOPS 29th Symposium on Operating Systems Principles}, 2023.

\bibitem[Huan et~al.(2025)Huan, Li, Zheng, Xu, Kim, Du, Poovendran, Neubig, and Yue]{huan2025does}
Maggie Huan, Yuetai Li, Tuney Zheng, Xiaoyu Xu, Seungone Kim, Minxin Du, Radha Poovendran, Graham Neubig, and Xiang Yue.
\newblock Does math reasoning improve general llm capabilities? understanding transferability of llm reasoning.
\newblock \emph{arXiv preprint arXiv:2507.00432}, 2025.

\bibitem[Lai et~al.(2025)Lai, Zhao, Feng, Ma, Liu, Zhao, Lin, Yi, Zhang, Liu, et~al.]{lai2025reinforcement}
Song Lai, Haohan Zhao, Rong Feng, Changyi Ma, Wenzhuo Liu, Hongbo Zhao, Xi~Lin, Dong Yi, Qingfu Zhang, Hongbin Liu, et~al.
\newblock Reinforcement fine-tuning naturally mitigates forgetting in continual post-training.
\newblock \emph{arXiv preprint arXiv:2507.05386}, 2025.

\bibitem[Rajpurkar et~al.(2018)Rajpurkar, Jia, and Liang]{rajpurkar2018knowdontknowunanswerable}
Pranav Rajpurkar, Robin Jia, and Percy Liang.
\newblock Know what you don't know: Unanswerable questions for squad, 2018.
\newblock URL \url{https://arxiv.org/abs/1806.03822}.

\bibitem[Geva et~al.(2021)Geva, Khashabi, Segal, Khot, Roth, and Berant]{geva2021didaristotleuselaptop}
Mor Geva, Daniel Khashabi, Elad Segal, Tushar Khot, Dan Roth, and Jonathan Berant.
\newblock Did aristotle use a laptop? a question answering benchmark with implicit reasoning strategies, 2021.
\newblock URL \url{https://arxiv.org/abs/2101.02235}.

\bibitem[Zhang et~al.(2025{\natexlab{a}})Zhang, Sun, Leng, Shen, Ziyin, Liang, and Zhang]{zhang2025laws}
Junyu Zhang, Yifan Sun, Tianang Leng, Jingyan Shen, Liu Ziyin, Paul~Pu Liang, and Huan Zhang.
\newblock When reasoning meets its laws.
\newblock In \emph{NeurIPS 2025 Workshop on Efficient Reasoning}, 2025{\natexlab{a}}.
\newblock URL \url{https://openreview.net/forum?id=lWjcbodr4M}.

\bibitem[Xu et~al.(2025{\natexlab{b}})Xu, Xie, Zhao, and He]{xu2025cod}
Silei Xu, Wenhao Xie, Lingxiao Zhao, and Pengcheng He.
\newblock Chain of draft: Thinking faster by writing less.
\newblock \emph{arXiv preprint arXiv:2502.18600}, 2025{\natexlab{b}}.
\newblock URL \url{https://arxiv.org/pdf/2502.18600}.

\bibitem[Renze and Guven(2024)]{renze2024ccot}
Matthew Renze and Erhan Guven.
\newblock The benefits of a concise chain of thought on problem-solving in large language models.
\newblock \emph{arXiv preprint arXiv:2401.05618}, 2024.
\newblock URL \url{https://arxiv.org/pdf/2401.05618v1.pdf}.

\bibitem[Wang et~al.(2025{\natexlab{c}})Wang, Lin, Cheng, and Shou]{wang2025ton}
Jiaqi Wang, Kevin~Qinghong Lin, James Cheng, and Mike~Zheng Shou.
\newblock Think or not? selective reasoning via reinforcement learning for vision-language models.
\newblock \emph{arXiv preprint arXiv:2505.16854}, 2025{\natexlab{c}}.
\newblock URL \url{https://arxiv.org/pdf/2505.16854}.

\bibitem[Huang et~al.(2025)Huang, Zhang, and Cardie]{huang2025hapo}
Chengyu Huang, Zhengxin Zhang, and Claire Cardie.
\newblock Hapo: Training language models to reason concisely via history-aware policy optimization.
\newblock \emph{arXiv preprint arXiv:2505.11225}, 2025.

\bibitem[Aggarwal and Welleck(2025)]{aggarwal2025l1}
Pranjal Aggarwal and Sean Welleck.
\newblock L1: Controlling how long a reasoning model thinks with reinforcement learning.
\newblock \emph{arXiv preprint arXiv:2503.04697}, 2025.
\newblock URL \url{https://arxiv.org/pdf/2503.04697}.

\bibitem[Liu et~al.(2025{\natexlab{a}})Liu, Dong, Lu, Diao, Liu, Chen, Yin, Wang, Cheng, Choi, et~al.]{liu2025dler}
Shih-Yang Liu, Xin Dong, Ximing Lu, Shizhe Diao, Mingjie Liu, Min-Hung Chen, Hongxu Yin, Yu-Chiang~Frank Wang, Kwang-Ting Cheng, Yejin Choi, et~al.
\newblock Dler: Doing length penalty right-incentivizing more intelligence per token via reinforcement learning.
\newblock \emph{arXiv preprint arXiv:2510.15110}, 2025{\natexlab{a}}.

\bibitem[Luo et~al.(2025)Luo, Tan, Wong, Shi, Tang, Roongta, Cai, Luo, Zhang, Li, et~al.]{luo2025deepscaler}
Michael Luo, Sijun Tan, Justin Wong, Xiaoxiang Shi, William~Y Tang, Manan Roongta, Colin Cai, Jeffrey Luo, Tianjun Zhang, Li~Erran Li, et~al.
\newblock Deepscaler: Surpassing o1-preview with a 1.5 b model by scaling rl.
\newblock \emph{Notion Blog}, 2025.

\bibitem[von Werra et~al.(2020)von Werra, Belkada, Tunstall, Beeching, Thrush, Lambert, Huang, Rasul, and Gallouédec]{vonwerra2020trl}
Leandro von Werra, Younes Belkada, Lewis Tunstall, Edward Beeching, Tristan Thrush, Nathan Lambert, Shengyi Huang, Kashif Rasul, and Quentin Gallouédec.
\newblock {TRL: Transformers Reinforcement Learning}, 2020.
\newblock URL \url{https://github.com/huggingface/trl}.

\bibitem[Lightman et~al.(2024)Lightman, Kosaraju, Burda, Edwards, Baker, Lee, Leike, Schulman, Sutskever, and Cobbe]{lightman2024let}
Hunter Lightman, Vineet Kosaraju, Yuri Burda, Harrison Edwards, Bowen Baker, Teddy Lee, Jan Leike, John Schulman, Ilya Sutskever, and Karl Cobbe.
\newblock Let's verify step by step.
\newblock In \emph{International Conference on Learning Representations}, volume 2024, pages 39578--39601, 2024.

\bibitem[Cobbe et~al.(2021)Cobbe, Kosaraju, Bavarian, Chen, Jun, Kaiser, Plappert, Tworek, Hilton, Nakano, et~al.]{cobbe2021training}
Karl Cobbe, Vineet Kosaraju, Mohammad Bavarian, Mark Chen, Heewoo Jun, Lukasz Kaiser, Matthias Plappert, Jerry Tworek, Jacob Hilton, Reiichiro Nakano, et~al.
\newblock Training verifiers to solve math word problems.
\newblock \emph{arXiv preprint arXiv:2110.14168}, 2021.

\bibitem[Wang et~al.(2022)Wang, Wei, Schuurmans, Le, Chi, Narang, Chowdhery, and Zhou]{wang2022self}
Xuezhi Wang, Jason Wei, Dale Schuurmans, Quoc Le, Ed~Chi, Sharan Narang, Aakanksha Chowdhery, and Denny Zhou.
\newblock Self-consistency improves chain of thought reasoning in language models.
\newblock \emph{arXiv preprint arXiv:2203.11171}, 2022.

\bibitem[Brown et~al.(2024)Brown, Juravsky, Ehrlich, Clark, Le, R{\'e}, and Mirhoseini]{brown2024large}
Bradley Brown, Jordan Juravsky, Ryan Ehrlich, Ronald Clark, Quoc~V Le, Christopher R{\'e}, and Azalia Mirhoseini.
\newblock Large language monkeys: Scaling inference compute with repeated sampling.
\newblock \emph{arXiv preprint arXiv:2407.21787}, 2024.

\bibitem[Wu et~al.(2025{\natexlab{a}})Wu, Sun, Li, Welleck, and Yang]{wu2025inference}
Yangzhen Wu, Zhiqing Sun, Shanda Li, Sean Welleck, and Yiming Yang.
\newblock Inference scaling laws: An empirical analysis of compute-optimal inference for llm problem-solving.
\newblock In \emph{The Thirteenth International Conference on Learning Representations}, 2025{\natexlab{a}}.

\bibitem[Snell et~al.(2025)Snell, Lee, Xu, and Kumar]{snell2025scaling}
Charlie Snell, Jaehoon Lee, Kelvin Xu, and Aviral Kumar.
\newblock Scaling llm test-time compute optimally can be more effective than scaling model parameters.
\newblock In \emph{The Thirteenth International Conference on Learning Representations}, 2025.

\bibitem[Renze(2024)]{renze2024effect}
Matthew Renze.
\newblock The effect of sampling temperature on problem solving in large language models.
\newblock In \emph{Findings of the association for computational linguistics: EMNLP 2024}, pages 7346--7356, 2024.

\bibitem[Hsu et~al.(2025)Hsu, Buffelli, McGowan, Liao, Chen, Vakili, and Shiu]{hsu2025group}
Chan-Jan Hsu, Davide Buffelli, Jamie McGowan, Feng-Ting Liao, Yi-Chang Chen, Sattar Vakili, and Da-shan Shiu.
\newblock Group think: Multiple concurrent reasoning agents collaborating at token level granularity.
\newblock \emph{arXiv preprint arXiv:2505.11107}, 2025.

\bibitem[Pan et~al.(2025)Pan, Li, Lian, Snell, Zhou, Yala, Darrell, Keutzer, and Suhr]{pan2025learning}
Jiayi Pan, Xiuyu Li, Long Lian, Charlie Snell, Yifei Zhou, Adam Yala, Trevor Darrell, Kurt Keutzer, and Alane Suhr.
\newblock Learning adaptive parallel reasoning with language models.
\newblock \emph{arXiv preprint arXiv:2504.15466}, 2025.

\bibitem[Rodionov et~al.(2025)Rodionov, Garipov, Shutova, Yakushev, Schultheis, Egiazarian, Sinitsin, Kuznedelev, and Alistarh]{rodionov2025hogwild}
Gleb Rodionov, Roman Garipov, Alina Shutova, George Yakushev, Erik Schultheis, Vage Egiazarian, Anton Sinitsin, Denis Kuznedelev, and Dan Alistarh.
\newblock Hogwild! inference: Parallel llm generation via concurrent attention.
\newblock \emph{arXiv preprint arXiv:2504.06261}, 2025.

\bibitem[Zheng et~al.(2025)Zheng, Zhang, Yu, Wang, Yang, Dai, Liu, Bao, Huang, Huang, et~al.]{zheng2025parallel}
Tong Zheng, Hongming Zhang, Wenhao Yu, Xiaoyang Wang, Xinyu Yang, Runpeng Dai, Rui Liu, Huiwen Bao, Chengsong Huang, Heng Huang, et~al.
\newblock Parallel-r1: Towards parallel thinking via reinforcement learning.
\newblock \emph{arXiv preprint arXiv:2509.07980}, 2025.

\bibitem[Yao et~al.(2023)Yao, Yu, Zhao, Shafran, Griffiths, Cao, and Narasimhan]{yao2023tree}
Shunyu Yao, Dian Yu, Jeffrey Zhao, Izhak Shafran, Tom Griffiths, Yuan Cao, and Karthik Narasimhan.
\newblock Tree of thoughts: Deliberate problem solving with large language models.
\newblock \emph{Advances in neural information processing systems}, 36:\penalty0 11809--11822, 2023.

\bibitem[Ning et~al.(2023)Ning, Lin, Zhou, Wang, Yang, and Wang]{ning2023skeleton}
Xuefei Ning, Zinan Lin, Zixuan Zhou, Zifu Wang, Huazhong Yang, and Yu~Wang.
\newblock Skeleton-of-thought: Prompting llms for efficient parallel generation.
\newblock \emph{arXiv preprint arXiv:2307.15337}, 2023.

\bibitem[Li et~al.(2025{\natexlab{b}})Li, Dong, Luan, Di, and Ding]{li2025enhancing}
Shuangtao Li, Shuaihao Dong, Kexin Luan, Xinhan Di, and Chaofan Ding.
\newblock Enhancing reasoning through process supervision with monte carlo tree search.
\newblock \emph{arXiv preprint arXiv:2501.01478}, 2025{\natexlab{b}}.

\bibitem[Liu et~al.(2024{\natexlab{a}})Liu, Zeng, Liu, Yan, He, Wang, Yan, Liu, and Zhou]{liu2024skywork}
Chris~Yuhao Liu, Liang Zeng, Jiacai Liu, Rui Yan, Jujie He, Chaojie Wang, Shuicheng Yan, Yang Liu, and Yahui Zhou.
\newblock Skywork-reward: Bag of tricks for reward modeling in llms.
\newblock \emph{arXiv preprint arXiv:2410.18451}, 2024{\natexlab{a}}.

\bibitem[Wang et~al.(2024)Wang, Dong, Delalleau, Zeng, Shen, Egert, Zhang, Sreedhar, and Kuchaiev]{wang2024helpsteer2}
Zhilin Wang, Yi~Dong, Olivier Delalleau, Jiaqi Zeng, Gerald Shen, Daniel Egert, Jimmy~J Zhang, Makesh~N Sreedhar, and Oleksii Kuchaiev.
\newblock Helpsteer 2: Open-source dataset for training top-performing reward models, 2024.

\bibitem[Wang et~al.(2023)Wang, Li, Shao, Xu, Dai, Li, Chen, Wu, and Sui]{wang2023math}
Peiyi Wang, Lei Li, Zhihong Shao, Runxin Xu, Damai Dai, Yifei Li, Deli Chen, Yu~Wu, and Zhifang Sui.
\newblock Math-shepherd: Verify and reinforce llms step-by-step without human annotations.
\newblock \emph{arXiv preprint arXiv:2312.08935}, 2023.

\bibitem[Zhang et~al.(2025{\natexlab{b}})Zhang, Zheng, Wu, Zhang, Lin, Yu, Liu, Zhou, and Lin]{zhang2025lessons}
Zhenru Zhang, Chujie Zheng, Yangzhen Wu, Beichen Zhang, Runji Lin, Bowen Yu, Dayiheng Liu, Jingren Zhou, and Junyang Lin.
\newblock The lessons of developing process reward models in mathematical reasoning.
\newblock \emph{arXiv preprint arXiv:2501.07301}, 2025{\natexlab{b}}.

\bibitem[Liu et~al.(2025{\natexlab{b}})Liu, Wang, Xu, Ma, Ruan, Li, Liu, and Wu]{liu2025inference}
Zijun Liu, Peiyi Wang, Runxin Xu, Shirong Ma, Chong Ruan, Peng Li, Yang Liu, and Yu~Wu.
\newblock Inference-time scaling for generalist reward modeling.
\newblock \emph{arXiv preprint arXiv:2504.02495}, 2025{\natexlab{b}}.

\bibitem[Chae et~al.(2025)Chae, Kim, Cho, Kim, Moon, Hwangbo, Lim, Kim, Hwang, Gwak, et~al.]{chae2025web}
Hyungjoo Chae, Sunghwan Kim, Junhee Cho, Seungone Kim, Seungjun Moon, Gyeom Hwangbo, Dongha Lim, Minjin Kim, Yeonjun Hwang, Minju Gwak, et~al.
\newblock Web-shepherd: Advancing prms for reinforcing web agents.
\newblock \emph{arXiv preprint arXiv:2505.15277}, 2025.

\bibitem[Chen et~al.(2025{\natexlab{b}})Chen, Chen, Sun, Liu, and Gan]{chen2025scaling}
Zhenfang Chen, Delin Chen, Rui Sun, Wenjun Liu, and Chuang Gan.
\newblock Scaling autonomous agents via automatic reward modeling and planning.
\newblock \emph{arXiv preprint arXiv:2502.12130}, 2025{\natexlab{b}}.

\bibitem[Khairi et~al.(2025)Khairi, D'souza, Fadaee, and Kreutzer]{khairi2025making}
Ammar Khairi, Daniel D'souza, Marzieh Fadaee, and Julia Kreutzer.
\newblock Making, not taking, the best of n.
\newblock \emph{arXiv preprint arXiv:2510.00931}, 2025.

\bibitem[Li et~al.(2025{\natexlab{c}})Li, Wang, Fu, Cui, Yang, and Cheng]{li2025drafts}
Yafu Li, Zhilin Wang, Tingchen Fu, Ganqu Cui, Sen Yang, and Yu~Cheng.
\newblock From drafts to answers: Unlocking llm potential via aggregation fine-tuning.
\newblock \emph{arXiv preprint arXiv:2501.11877}, 2025{\natexlab{c}}.

\bibitem[Qi et~al.(2025)Qi, Ye, Tang, Zhu, and Choi]{qi2025learning}
Jianing Qi, Xi~Ye, Hao Tang, Zhigang Zhu, and Eunsol Choi.
\newblock Learning to reason across parallel samples for llm reasoning.
\newblock \emph{arXiv preprint arXiv:2506.09014}, 2025.

\bibitem[Zhao et~al.(2025)Zhao, Aggarwal, Saha, Celikyilmaz, Weston, and Kulikov]{zhao2025majority}
Wenting Zhao, Pranjal Aggarwal, Swarnadeep Saha, Asli Celikyilmaz, Jason Weston, and Ilia Kulikov.
\newblock The majority is not always right: Rl training for solution aggregation.
\newblock \emph{arXiv preprint arXiv:2509.06870}, 2025.

\bibitem[Hao et~al.(2026)Hao, Wei, Wang, Huang, Jiang, Jiang, Cao, and Ren]{hao2025scaling}
Zixu Hao, Jianyu Wei, Tuowei Wang, Minxing Huang, Huiqiang Jiang, Shiqi Jiang, Ting Cao, and Ju~Ren.
\newblock Scaling llm test-time compute with mobile npu on smartphones.
\newblock In \emph{EuroSys 2026}. ACM, November 2026.
\newblock URL \url{https://www.microsoft.com/en-us/research/publication/scaling-llm-test-time-compute-with-mobile-npu-on-smartphones/}.

\bibitem[Sun et~al.(2024{\natexlab{a}})Sun, Haider, Zhang, Yang, Qiu, Yin, Wang, Bartlett, and Zanette]{sun2024fast}
Hanshi Sun, Momin Haider, Ruiqi Zhang, Huitao Yang, Jiahao Qiu, Ming Yin, Mengdi Wang, Peter Bartlett, and Andrea Zanette.
\newblock Fast best-of-n decoding via speculative rejection.
\newblock \emph{Advances in Neural Information Processing Systems}, 37:\penalty0 32630--32652, 2024{\natexlab{a}}.

\bibitem[Hooper et~al.(2025)Hooper, Kim, Moon, Dilmen, Maheswaran, Lee, Mahoney, Shao, Keutzer, and Gholami]{hooper2025ets}
Coleman Hooper, Sehoon Kim, Suhong Moon, Kerem Dilmen, Monishwaran Maheswaran, Nicholas Lee, Michael~W Mahoney, Sophia Shao, Kurt Keutzer, and Amir Gholami.
\newblock Ets: Efficient tree search for inference-time scaling.
\newblock \emph{arXiv preprint arXiv:2502.13575}, 2025.

\bibitem[Zhang et~al.(2025{\natexlab{c}})Zhang, Hosseini, Bansal, Kazemi, Kumar, and Agarwal]{zhang2025generative}
Lunjun Zhang, Arian Hosseini, Hritik Bansal, Mehran Kazemi, Aviral Kumar, and Rishabh Agarwal.
\newblock Generative verifiers: Reward modeling as next-token prediction.
\newblock In \emph{The Thirteenth International Conference on Learning Representations}, 2025{\natexlab{c}}.

\bibitem[Krishnamoorthi(2018)]{krishnamoorthi2018quantizing}
Raghuraman Krishnamoorthi.
\newblock Quantizing deep convolutional networks for efficient inference: A whitepaper.
\newblock \emph{arXiv preprint arXiv:1806.08342}, 2018.

\bibitem[Nagel et~al.(2021)Nagel, Fournarakis, Amjad, Bondarenko, Van~Baalen, and Blankevoort]{nagel2021white}
Markus Nagel, Marios Fournarakis, Rana~Ali Amjad, Yelysei Bondarenko, Mart Van~Baalen, and Tijmen Blankevoort.
\newblock A white paper on neural network quantization.
\newblock \emph{arXiv preprint arXiv:2106.08295}, 2021.

\bibitem[{Horowitz}(2014)]{horowitz}
M.~{Horowitz}.
\newblock 1.1 computing's energy problem (and what we can do about it).
\newblock In \emph{2014 IEEE International Solid-State Circuits Conference Digest of Technical Papers (ISSCC)}, pages 10--14, 2014.
\newblock \doi{10.1109/ISSCC.2014.6757323}.

\bibitem[Hubara et~al.(2017)Hubara, Courbariaux, Soudry, El-Yaniv, and Bengio]{hubara2017quantized}
Itay Hubara, Matthieu Courbariaux, Daniel Soudry, Ran El-Yaniv, and Yoshua Bengio.
\newblock Quantized neural networks: Training neural networks with low precision weights and activations.
\newblock \emph{The Journal of Machine Learning Research}, 18\penalty0 (1):\penalty0 6869--6898, 2017.

\bibitem[Zhou et~al.(2016)Zhou, Wu, Ni, Zhou, Wen, and Zou]{zhou2016dorefa}
Shuchang Zhou, Yuxin Wu, Zekun Ni, Xinyu Zhou, He~Wen, and Yuheng Zou.
\newblock Dorefa-net: Training low bitwidth convolutional neural networks with low bitwidth gradients.
\newblock \emph{arXiv preprint arXiv:1606.06160}, 2016.

\bibitem[Banner et~al.(2018)Banner, Nahshan, Hoffer, and Soudry]{banner2018post}
Ron Banner, Yury Nahshan, Elad Hoffer, and Daniel Soudry.
\newblock Post-training 4-bit quantization of convolution networks for rapid-deployment.
\newblock \emph{arXiv preprint arXiv:1810.05723}, 2018.

\bibitem[Cai et~al.(2020)Cai, Yao, Dong, Gholami, Mahoney, and Keutzer]{cai2020zeroq}
Yaohui Cai, Zhewei Yao, Zhen Dong, Amir Gholami, Michael~W Mahoney, and Kurt Keutzer.
\newblock Zeroq: A novel zero shot quantization framework.
\newblock In \emph{Proceedings of the IEEE/CVF Conference on Computer Vision and Pattern Recognition}, pages 13169--13178, 2020.

\bibitem[Choukroun et~al.(2019)Choukroun, Kravchik, Yang, and Kisilev]{choukroun2019low}
Yoni Choukroun, Eli Kravchik, Fan Yang, and Pavel Kisilev.
\newblock Low-bit quantization of neural networks for efficient inference.
\newblock In \emph{ICCV Workshops}, pages 3009--3018, 2019.

\bibitem[Hubara et~al.(2020)Hubara, Nahshan, Hanani, Banner, and Soudry]{hubara2020improving}
Itay Hubara, Yury Nahshan, Yair Hanani, Ron Banner, and Daniel Soudry.
\newblock Improving post training neural quantization: Layer-wise calibration and integer programming.
\newblock \emph{arXiv preprint arXiv:2006.10518}, 2020.

\bibitem[Meller et~al.(2019)Meller, Finkelstein, Almog, and Grobman]{meller2019same}
Eldad Meller, Alexander Finkelstein, Uri Almog, and Mark Grobman.
\newblock Same, same but different: Recovering neural network quantization error through weight factorization.
\newblock In \emph{International Conference on Machine Learning}, pages 4486--4495. PMLR, 2019.

\bibitem[Zhao et~al.(2019)Zhao, Hu, Dotzel, De~Sa, and Zhang]{zhao2019improving}
Ritchie Zhao, Yuwei Hu, Jordan Dotzel, Chris De~Sa, and Zhiru Zhang.
\newblock Improving neural network quantization without retraining using outlier channel splitting.
\newblock In \emph{International conference on machine learning}, pages 7543--7552. PMLR, 2019.

\bibitem[Nagel et~al.(2019)Nagel, Baalen, Blankevoort, and Welling]{Nagel_2019_ICCV}
Markus Nagel, Mart~van Baalen, Tijmen Blankevoort, and Max Welling.
\newblock Data-free quantization through weight equalization and bias correction.
\newblock In \emph{Proceedings of the IEEE/CVF International Conference on Computer Vision (ICCV)}, October 2019.

\bibitem[Nagel et~al.(2020)Nagel, Amjad, van Baalen, Louizos, and Blankevoort]{nagel_up_2020}
Markus Nagel, Rana~Ali Amjad, Mart van Baalen, Christos Louizos, and Tijmen Blankevoort.
\newblock Up or {Down}? {Adaptive} {Rounding} for {Post}-{Training} {Quantization}, April 2020.
\newblock URL \url{https://arxiv.org/abs/2004.10568v2}.

\bibitem[Li et~al.(2021)Li, Gong, Tan, Yang, Hu, Zhang, Yu, Wang, and Gu]{li2021brecq}
Yuhang Li, Ruihao Gong, Xu~Tan, Yang Yang, Peng Hu, Qi~Zhang, Fengwei Yu, Wei Wang, and Shi Gu.
\newblock Brecq: Pushing the limit of post-training quantization by block reconstruction.
\newblock \emph{arXiv preprint arXiv:2102.05426}, 2021.

\bibitem[Gupta et~al.(2015)Gupta, Agrawal, Gopalakrishnan, and Narayanan]{gupta2015deep}
Suyog Gupta, Ankur Agrawal, Kailash Gopalakrishnan, and Pritish Narayanan.
\newblock Deep learning with limited numerical precision.
\newblock In \emph{International conference on machine learning}, pages 1737--1746. PMLR, 2015.

\bibitem[Jacob et~al.(2018)Jacob, Kligys, Chen, Zhu, Tang, Howard, Adam, and Kalenichenko]{jacob2018quantization}
Benoit Jacob, Skirmantas Kligys, Bo~Chen, Menglong Zhu, Matthew Tang, Andrew Howard, Hartwig Adam, and Dmitry Kalenichenko.
\newblock Quantization and training of neural networks for efficient integer-arithmetic-only inference.
\newblock In \emph{Proceedings of the IEEE Conference on Computer Vision and Pattern Recognition}, pages 2704--2713, 2018.

\bibitem[Esser et~al.(2020)Esser, McKinstry, Bablani, Appuswamy, and Modha]{lsq}
Steven~K. Esser, Jeffrey~L. McKinstry, Deepika Bablani, Rathinakumar Appuswamy, and Dharmendra~S. Modha.
\newblock Learned step size quantization.
\newblock In \emph{International Conference on Learning Representations (ICLR)}, 2020.

\bibitem[Nagel et~al.(2022)Nagel, Fournarakis, Bondarenko, and Blankevoort]{nagel_oscillations_2022}
Markus Nagel, Marios Fournarakis, Yelysei Bondarenko, and Tijmen Blankevoort.
\newblock Overcoming oscillations in quantization-aware training.
\newblock In \emph{International Conference on Machine Learning}, pages 16318--16330. PMLR, 2022.

\bibitem[Liu et~al.(2023{\natexlab{b}})Liu, Oguz, Zhao, Chang, Stock, Mehdad, Shi, Krishnamoorthi, and Chandra]{liu2023llm}
Zechun Liu, Barlas Oguz, Changsheng Zhao, Ernie Chang, Pierre Stock, Yashar Mehdad, Yangyang Shi, Raghuraman Krishnamoorthi, and Vikas Chandra.
\newblock Llm-qat: Data-free quantization aware training for large language models.
\newblock \emph{arXiv preprint arXiv:2305.17888}, 2023{\natexlab{b}}.

\bibitem[Du et~al.(2024)Du, Zhang, Cao, Guo, Cao, Chu, and Xu]{du2024bitdistiller}
Dayou Du, Yijia Zhang, Shijie Cao, Jiaqi Guo, Ting Cao, Xiaowen Chu, and Ningyi Xu.
\newblock Bitdistiller: Unleashing the potential of sub-4-bit llms via self-distillation.
\newblock \emph{arXiv preprint arXiv:2402.10631}, 2024.

\bibitem[Liu et~al.(2025{\natexlab{c}})Liu, Zhao, Huang, Chen, Zhang, Zhao, Roy, Jin, Xiong, Shi, Xiao, Tian, Soran, Krishnamoorthi, Blankevoort, and Chandra]{liu_paretoq_2025}
Zechun Liu, Changsheng Zhao, Hanxian Huang, Sijia Chen, Jing Zhang, Jiawei Zhao, Scott Roy, Lisa Jin, Yunyang Xiong, Yangyang Shi, Lin Xiao, Yuandong Tian, Bilge Soran, Raghuraman Krishnamoorthi, Tijmen Blankevoort, and Vikas Chandra.
\newblock Paretoq: Improving scaling laws in extremely low-bit llm quantization, 2025{\natexlab{c}}.
\newblock URL \url{https://arxiv.org/abs/2502.02631}.

\bibitem[Chen et~al.(2025{\natexlab{c}})Chen, Zhang, Liu, Zeng, Xue, Liu, Li, Ma, Huang, Zhou, et~al.]{chen2025scaling_qat}
Mengzhao Chen, Chaoyi Zhang, Jing Liu, Yutao Zeng, Zeyue Xue, Zhiheng Liu, Yunshui Li, Jin Ma, Jie Huang, Xun Zhou, et~al.
\newblock Scaling law for quantization-aware training.
\newblock \emph{arXiv preprint arXiv:2505.14302}, 2025{\natexlab{c}}.

\bibitem[Dettmers et~al.(2024)Dettmers, Pagnoni, Holtzman, and Zettlemoyer]{dettmers2024qlora}
Tim Dettmers, Artidoro Pagnoni, Ari Holtzman, and Luke Zettlemoyer.
\newblock Qlora: Efficient finetuning of quantized llms.
\newblock \emph{Advances in Neural Information Processing Systems}, 36, 2024.

\bibitem[Xu et~al.(2023)Xu, Xie, Gu, Chen, Chang, Zhang, Chen, Zhang, and Tian]{xu2023qa}
Yuhui Xu, Lingxi Xie, Xiaotao Gu, Xin Chen, Heng Chang, Hengheng Zhang, Zhengsu Chen, Xiaopeng Zhang, and Qi~Tian.
\newblock Qa-lora: Quantization-aware low-rank adaptation of large language models.
\newblock \emph{arXiv preprint arXiv:2309.14717}, 2023.

\bibitem[Li et~al.(2023)Li, Yu, Liang, He, Karampatziakis, Chen, and Zhao]{li2023loftq}
Yixiao Li, Yifan Yu, Chen Liang, Pengcheng He, Nikos Karampatziakis, Weizhu Chen, and Tuo Zhao.
\newblock Loftq: Lora-fine-tuning-aware quantization for large language models.
\newblock \emph{arXiv preprint arXiv:2310.08659}, 2023.

\bibitem[Guo et~al.(2023)Guo, Greengard, Xing, and Kim]{guo2023lq}
Han Guo, Philip Greengard, Eric~P Xing, and Yoon Kim.
\newblock Lq-lora: Low-rank plus quantized matrix decomposition for efficient language model finetuning.
\newblock \emph{arXiv preprint arXiv:2311.12023}, 2023.

\bibitem[Kim et~al.(2024)Kim, Lee, Kim, Park, Yoo, Kwon, and Lee]{kim2024memory}
Jeonghoon Kim, Jung~Hyun Lee, Sungdong Kim, Joonsuk Park, Kang~Min Yoo, Se~Jung Kwon, and Dongsoo Lee.
\newblock Memory-efficient fine-tuning of compressed large language models via sub-4-bit integer quantization.
\newblock \emph{Advances in Neural Information Processing Systems}, 36, 2024.

\bibitem[Bondarenko et~al.(2024)Bondarenko, Del~Chiaro, and Nagel]{bondarenko2024low}
Yelysei Bondarenko, Riccardo Del~Chiaro, and Markus Nagel.
\newblock Low-rank quantization-aware training for llms.
\newblock \emph{arXiv preprint arXiv:2406.06385}, 2024.

\bibitem[Bondarenko et~al.(2021)Bondarenko, Nagel, and Blankevoort]{bondarenko_understanding_2021}
Yelysei Bondarenko, Markus Nagel, and Tijmen Blankevoort.
\newblock Understanding and overcoming the challenges of efficient transformer quantization.
\newblock In \emph{Proceedings of the 2021 Conference on Empirical Methods in Natural Language Processing}, pages 7947--7969, Online and Punta Cana, Dominican Republic, November 2021. Association for Computational Linguistics.
\newblock \doi{10.18653/v1/2021.emnlp-main.627}.
\newblock URL \url{https://aclanthology.org/2021.emnlp-main.627}.

\bibitem[Kovaleva et~al.(2021)Kovaleva, Kulshreshtha, Rogers, and Rumshisky]{kovaleva_bert_2021}
Olga Kovaleva, Saurabh Kulshreshtha, Anna Rogers, and Anna Rumshisky.
\newblock Bert busters: Outlier dimensions that disrupt transformers.
\newblock In \emph{Findings of the Association for Computational Linguistics: ACL-IJCNLP 2021}, pages 3392--3405, 2021.

\bibitem[Dettmers et~al.(2022)Dettmers, Lewis, Belkada, and Zettlemoyer]{dettmers_gpt3_int8_2022}
Tim Dettmers, Mike Lewis, Younes Belkada, and Luke Zettlemoyer.
\newblock Gpt3. int8 (): 8-bit matrix multiplication for transformers at scale.
\newblock In \emph{Advances in Neural Information Processing Systems}, 2022.

\bibitem[Bondarenko et~al.(2023)Bondarenko, Nagel, and Blankevoort]{bondarenko_quantizable_2023}
Yelysei Bondarenko, Markus Nagel, and Tijmen Blankevoort.
\newblock Quantizable {Transformers}: {Removing} {Outliers} by {Helping} {Attention} {Heads} {Do} {Nothing}.
\newblock \emph{Advances in Neural Information Processing Systems}, 2023.
\newblock URL \url{https://arxiv.org/abs/2306.12929v2}.

\bibitem[Sun et~al.(2024{\natexlab{b}})Sun, Chen, Kolter, and Liu]{sun_massive_2024}
Mingjie Sun, Xinlei Chen, J~Zico Kolter, and Zhuang Liu.
\newblock Massive activations in large language models.
\newblock \emph{arXiv preprint arXiv:2402.17762}, 2024{\natexlab{b}}.

\bibitem[Frantar et~al.(2022)Frantar, Ashkboos, Hoefler, and Alistarh]{frantar_gptq_2022}
Elias Frantar, Saleh Ashkboos, Torsten Hoefler, and Dan Alistarh.
\newblock Gptq: Accurate post-training quantization for generative pre-trained transformers.
\newblock \emph{arXiv preprint arXiv:2210.17323}, 2022.

\bibitem[Dettmers et~al.(2023)Dettmers, Svirschevski, Egiazarian, Kuznedelev, Frantar, Ashkboos, Borzunov, Hoefler, and Alistarh]{dettmers2023spqr}
Tim Dettmers, Ruslan Svirschevski, Vage Egiazarian, Denis Kuznedelev, Elias Frantar, Saleh Ashkboos, Alexander Borzunov, Torsten Hoefler, and Dan Alistarh.
\newblock Spqr: A sparse-quantized representation for near-lossless llm weight compression.
\newblock \emph{arXiv preprint arXiv:2306.03078}, 2023.

\bibitem[Lin et~al.(2023)Lin, Tang, Tang, Yang, Dang, Gan, and Han]{lin2023awq}
Ji~Lin, Jiaming Tang, Haotian Tang, Shang Yang, Xingyu Dang, Chuang Gan, and Song Han.
\newblock Awq: Activation-aware weight quantization for llm compression and acceleration.
\newblock \emph{arXiv preprint arXiv:2306.00978}, 2023.

\bibitem[Lee et~al.(2024)Lee, Jin, Kim, Kim, and Park]{lee2024owq}
Changhun Lee, Jungyu Jin, Taesu Kim, Hyungjun Kim, and Eunhyeok Park.
\newblock Owq: Outlier-aware weight quantization for efficient fine-tuning and inference of large language models.
\newblock In \emph{Proceedings of the AAAI Conference on Artificial Intelligence}, volume~38, pages 13355--13364, 2024.

\bibitem[Jeon et~al.(2023)Jeon, Lee, Park, and Kim]{jeon2023frustratingly}
Yongkweon Jeon, Chungman Lee, Kyungphil Park, and Ho-young Kim.
\newblock A frustratingly easy post-training quantization scheme for llms.
\newblock In \emph{Proceedings of the 2023 Conference on Empirical Methods in Natural Language Processing}, pages 14446--14461, 2023.

\bibitem[Lee et~al.(2023{\natexlab{a}})Lee, Kim, Kwon, and Lee]{lee2023flexround}
Jung~Hyun Lee, Jeonghoon Kim, Se~Jung Kwon, and Dongsoo Lee.
\newblock Flexround: Learnable rounding based on element-wise division for post-training quantization.
\newblock In \emph{International Conference on Machine Learning}, pages 18913--18939. PMLR, 2023{\natexlab{a}}.

\bibitem[Luo et~al.(2023)Luo, Gao, Zhang, Fan, Zhang, and Xu]{luo2023long}
Yan Luo, Yangcheng Gao, Zhao Zhang, Jicong Fan, Haijun Zhang, and Mingliang Xu.
\newblock Long-range zero-shot generative deep network quantization.
\newblock \emph{Neural Networks}, 166:\penalty0 683--691, 2023.

\bibitem[Chee et~al.(2024)Chee, Cai, Kuleshov, and De~Sa]{chee2024quip}
Jerry Chee, Yaohui Cai, Volodymyr Kuleshov, and Christopher~M De~Sa.
\newblock Quip: 2-bit quantization of large language models with guarantees.
\newblock \emph{Advances in Neural Information Processing Systems}, 36, 2024.

\bibitem[Xiao et~al.(2023)Xiao, Lin, Seznec, Wu, Demouth, and Han]{xiao2023smoothquant}
Guangxuan Xiao, Ji~Lin, Mickael Seznec, Hao Wu, Julien Demouth, and Song Han.
\newblock Smoothquant: Accurate and efficient post-training quantization for large language models.
\newblock In \emph{International Conference on Machine Learning}, pages 38087--38099. PMLR, 2023.

\bibitem[Wei et~al.(2022{\natexlab{b}})Wei, Zhang, Zhang, Gong, Zhang, Zhang, Yu, and Liu]{wei2022outlier}
Xiuying Wei, Yunchen Zhang, Xiangguo Zhang, Ruihao Gong, Shanghang Zhang, Qi~Zhang, Fengwei Yu, and Xianglong Liu.
\newblock Outlier suppression: Pushing the limit of low-bit transformer language models.
\newblock \emph{arXiv preprint arXiv:2209.13325}, 2022{\natexlab{b}}.

\bibitem[Lee et~al.(2023{\natexlab{b}})Lee, Kim, Baek, Hwang, Sung, and Choi]{lee2023enhancing}
Janghwan Lee, Minsoo Kim, Seungcheol Baek, Seok~Joong Hwang, Wonyong Sung, and Jungwook Choi.
\newblock Enhancing computation efficiency in large language models through weight and activation quantization.
\newblock \emph{arXiv preprint arXiv:2311.05161}, 2023{\natexlab{b}}.

\bibitem[Liu et~al.(2023{\natexlab{c}})Liu, Gong, Wei, Dong, Cai, and Zhuang]{liu2023qllm}
Jing Liu, Ruihao Gong, Xiuying Wei, Zhiwei Dong, Jianfei Cai, and Bohan Zhuang.
\newblock Qllm: Accurate and efficient low-bitwidth quantization for large language models.
\newblock \emph{arXiv preprint arXiv:2310.08041}, 2023{\natexlab{c}}.

\bibitem[Wei et~al.(2023)Wei, Zhang, Li, Zhang, Gong, Guo, and Liu]{wei_outlier_2023}
Xiuying Wei, Yunchen Zhang, Yuhang Li, Xiangguo Zhang, Ruihao Gong, Jinyang Guo, and Xianglong Liu.
\newblock Outlier {Suppression}+: {Accurate} quantization of large language models by equivalent and optimal shifting and scaling, October 2023.
\newblock URL \url{http://arxiv.org/abs/2304.09145}.
\newblock arXiv:2304.09145 [cs].

\bibitem[Yuan et~al.(2023)Yuan, Niu, Liu, Liu, Wang, Shang, Sun, Wu, Wu, and Wu]{yuan2023rptq}
Zhihang Yuan, Lin Niu, Jiawei Liu, Wenyu Liu, Xinggang Wang, Yuzhang Shang, Guangyu Sun, Qiang Wu, Jiaxiang Wu, and Bingzhe Wu.
\newblock Rptq: Reorder-based post-training quantization for large language models.
\newblock \emph{arXiv preprint arXiv:2304.01089}, 2023.

\bibitem[Tang et~al.(2024)Tang, Sun, Wu, Liu, Zhu, and Kang]{tang2024easyquant}
Hanlin Tang, Yifu Sun, Decheng Wu, Kai Liu, Jianchen Zhu, and Zhanhui Kang.
\newblock Easyquant: An efficient data-free quantization algorithm for llms.
\newblock \emph{arXiv preprint arXiv:2403.02775}, 2024.

\bibitem[Yao et~al.(2022{\natexlab{b}})Yao, Yazdani~Aminabadi, Zhang, Wu, Li, and He]{yao2022zeroquant}
Zhewei Yao, Reza Yazdani~Aminabadi, Minjia Zhang, Xiaoxia Wu, Conglong Li, and Yuxiong He.
\newblock Zeroquant: Efficient and affordable post-training quantization for large-scale transformers.
\newblock In S.~Koyejo, S.~Mohamed, A.~Agarwal, D.~Belgrave, K.~Cho, and A.~Oh, editors, \emph{Advances in Neural Information Processing Systems}, volume~35, pages 27168--27183. Curran Associates, Inc., 2022{\natexlab{b}}.
\newblock URL \url{https://proceedings.neurips.cc/paper_files/paper/2022/file/adf7fa39d65e2983d724ff7da57f00ac-Paper-Conference.pdf}.

\bibitem[Lin et~al.(2024{\natexlab{a}})Lin, Tang, Yang, Zhang, Xiao, Gan, and Han]{lin2024qserve}
Yujun Lin, Haotian Tang, Shang Yang, Zhekai Zhang, Guangxuan Xiao, Chuang Gan, and Song Han.
\newblock Qserve: W4a8kv4 quantization and system co-design for efficient llm serving.
\newblock \emph{arXiv preprint arXiv:2405.04532}, 2024{\natexlab{a}}.

\bibitem[Shao et~al.(2024{\natexlab{b}})Shao, Chen, Zhang, Xu, Zhao, Li, Zhang, Gao, Qiao, and Luo]{shao_omniquant_2024}
Wenqi Shao, Mengzhao Chen, Zhaoyang Zhang, Peng Xu, Lirui Zhao, Zhiqian Li, Kaipeng Zhang, Peng Gao, Yu~Qiao, and Ping Luo.
\newblock {OmniQuant}: {Omnidirectionally} {Calibrated} {Quantization} for {Large} {Language} {Models}, March 2024{\natexlab{b}}.
\newblock URL \url{http://arxiv.org/abs/2308.13137}.
\newblock arXiv:2308.13137 [cs].

\bibitem[Ashkboos et~al.(2024)Ashkboos, Mohtashami, Croci, Li, Jaggi, Alistarh, Hoefler, and Hensman]{ashkboos_quarot_2024}
Saleh Ashkboos, Amirkeivan Mohtashami, Maximilian~L. Croci, Bo~Li, Martin Jaggi, Dan Alistarh, Torsten Hoefler, and James Hensman.
\newblock {QuaRot}: {Outlier}-{Free} 4-{Bit} {Inference} in {Rotated} {LLMs}, March 2024.
\newblock URL \url{https://arxiv.org/abs/2404.00456v1}.

\bibitem[Liu et~al.(2024{\natexlab{b}})Liu, Zhao, Fedorov, Soran, Choudhary, Krishnamoorthi, Chandra, Tian, and Blankevoort]{liu_spinquant_2024}
Zechun Liu, Changsheng Zhao, Igor Fedorov, Bilge Soran, Dhruv Choudhary, Raghuraman Krishnamoorthi, Vikas Chandra, Yuandong Tian, and Tijmen Blankevoort.
\newblock {SpinQuant}: {LLM} quantization with learned rotations, May 2024{\natexlab{b}}.
\newblock URL \url{https://arxiv.org/abs/2405.16406v2}.

\bibitem[Lin et~al.(2024{\natexlab{b}})Lin, Xu, Wu, Cui, Zhang, Mou, Song, Sun, and Wei]{lin_duquant_2024}
Haokun Lin, Haobo Xu, Yichen Wu, Jingzhi Cui, Yingtao Zhang, Linzhan Mou, Linqi Song, Zhenan Sun, and Ying Wei.
\newblock {DuQuant}: {Distributing} {Outliers} via {Dual} {Transformation} {Makes} {Stronger} {Quantized} {LLMs}.
\newblock In \emph{Advances in {Neural} {Information} {Processing} {Systems}}. arXiv, November 2024{\natexlab{b}}.
\newblock \doi{10.48550/arXiv.2406.01721}.
\newblock URL \url{http://arxiv.org/abs/2406.01721}.
\newblock arXiv:2406.01721.

\bibitem[Hu et~al.(2025)Hu, Cheng, Yang, Xu, Yuan, Yu, Xu, Jiang, and Zhou]{hu_ostquant_2025}
Xing Hu, Yuan Cheng, Dawei Yang, Zukang Xu, Zhihang Yuan, Jiangyong Yu, Chen Xu, Zhe Jiang, and Sifan Zhou.
\newblock {OSTQuant}: {Refining} {Large} {Language} {Model} {Quantization} with {Orthogonal} and {Scaling} {Transformations} for {Better} {Distribution} {Fitting}.
\newblock In \emph{The {Thirteenth} {International} {Conference} on {Learning} {Representations}}, 2025.
\newblock URL \url{https://openreview.net/forum?id=rAcgDBdKnP}.

\bibitem[Sun et~al.(2025)Sun, Liu, Bai, Bao, Zhao, Li, Hu, Yu, Hou, Yuan, Jiang, Liu, and Yao]{sun_flatquant_2024}
Yuxuan Sun, Ruikang Liu, Haoli Bai, Han Bao, Kang Zhao, Yuening Li, Jiaxin Hu, Xianzhi Yu, Lu~Hou, Chun Yuan, Xin Jiang, Wulong Liu, and Jun Yao.
\newblock Flatquant: Flatness matters for {LLM} quantization.
\newblock In \emph{Forty-second International Conference on Machine Learning}, 2025.
\newblock URL \url{https://openreview.net/forum?id=uTz2Utym5n}.

\bibitem[van Breugel et~al.(2025)van Breugel, Bondarenko, Whatmough, and Nagel]{van2025fptquant}
Boris van Breugel, Yelysei Bondarenko, Paul Whatmough, and Markus Nagel.
\newblock Fptquant: Function-preserving transforms for llm quantization.
\newblock \emph{arXiv preprint arXiv:2506.04985}, 2025.

\bibitem[Lee et~al.(2025)Lee, Shin, Kim, You, and Chen]{lee2025unifying}
Jung~Hyun Lee, Seungjae Shin, Vinnam Kim, Jaeseong You, and An~Chen.
\newblock Unifying block-wise ptq and distillation-based qat for progressive quantization toward 2-bit instruction-tuned llms.
\newblock \emph{arXiv preprint arXiv:2506.09104}, 2025.

\bibitem[Allal et~al.(2025)Allal, Lozhkov, Bakouch, Bl{\'a}zquez, Penedo, Tunstall, Marafioti, Kydl{\'\i}{\v{c}}ek, Lajar{\'\i}n, Srivastav, et~al.]{allal2025smollm2}
Loubna~Ben Allal, Anton Lozhkov, Elie Bakouch, Gabriel~Mart{\'\i}n Bl{\'a}zquez, Guilherme Penedo, Lewis Tunstall, Andr{\'e}s Marafioti, Hynek Kydl{\'\i}{\v{c}}ek, Agust{\'\i}n~Piqueres Lajar{\'\i}n, Vaibhav Srivastav, et~al.
\newblock Smollm2: When smol goes big--data-centric training of a small language model.
\newblock \emph{arXiv preprint arXiv:2502.02737}, 2025.

\bibitem[Li et~al.(2024)Li, Fang, Smyrnis, Ivgi, Jordan, Gadre, Bansal, Guha, Keh, Arora, et~al.]{li2024datacomp}
Jeffrey Li, Alex Fang, Georgios Smyrnis, Maor Ivgi, Matt Jordan, Samir~Yitzhak Gadre, Hritik Bansal, Etash Guha, Sedrick~Scott Keh, Kushal Arora, et~al.
\newblock Datacomp-lm: In search of the next generation of training sets for language models.
\newblock \emph{Advances in Neural Information Processing Systems}, 37:\penalty0 14200--14282, 2024.

\bibitem[Penedo et~al.(2024)Penedo, Kydl{\'\i}{\v{c}}ek, Lozhkov, Mitchell, Raffel, Von~Werra, Wolf, et~al.]{penedo2024fineweb}
Guilherme Penedo, Hynek Kydl{\'\i}{\v{c}}ek, Anton Lozhkov, Margaret Mitchell, Colin~A Raffel, Leandro Von~Werra, Thomas Wolf, et~al.
\newblock The fineweb datasets: Decanting the web for the finest text data at scale.
\newblock \emph{Advances in Neural Information Processing Systems}, 37:\penalty0 30811--30849, 2024.

\bibitem[Bisk et~al.(2020)Bisk, Zellers, Bras, Gao, and Choi]{bisk_piqa_2020}
Yonatan Bisk, Rowan Zellers, Ronan~Le Bras, Jianfeng Gao, and Yejin Choi.
\newblock {PIQA}: {Reasoning} about {Physical} {Commonsense} in {Natural} {Language}.
\newblock \emph{Proceedings of the AAAI Conference on Artificial Intelligence}, 34\penalty0 (05):\penalty0 7432--7439, April 2020.
\newblock ISSN 2374-3468.
\newblock \doi{10.1609/aaai.v34i05.6239}.
\newblock URL \url{https://ojs.aaai.org/index.php/AAAI/article/view/6239}.
\newblock Number: 05.

\bibitem[Sakaguchi et~al.(2021)Sakaguchi, Bras, Bhagavatula, and Choi]{sakaguchi_winogrande_2021}
Keisuke Sakaguchi, Ronan~Le Bras, Chandra Bhagavatula, and Yejin Choi.
\newblock {WinoGrande}: an adversarial winograd schema challenge at scale.
\newblock \emph{Commun. ACM}, 64\penalty0 (9):\penalty0 99--106, August 2021.
\newblock ISSN 0001-0782.
\newblock \doi{10.1145/3474381}.
\newblock URL \url{https://dl.acm.org/doi/10.1145/3474381}.

\bibitem[Zellers et~al.(2019)Zellers, Holtzman, Bisk, Farhadi, and Choi]{zellers_hellaswag_2019}
Rowan Zellers, Ari Holtzman, Yonatan Bisk, Ali Farhadi, and Yejin Choi.
\newblock {HellaSwag}: {Can} a {Machine} {Really} {Finish} {Your} {Sentence}?, May 2019.
\newblock URL \url{http://arxiv.org/abs/1905.07830}.
\newblock arXiv:1905.07830 [cs].

\bibitem[Clark et~al.(2018)Clark, Cowhey, Etzioni, Khot, Sabharwal, Schoenick, and Tafjord]{clark_think_2018}
Peter Clark, Isaac Cowhey, Oren Etzioni, Tushar Khot, Ashish Sabharwal, Carissa Schoenick, and Oyvind Tafjord.
\newblock Think you have {Solved} {Question} {Answering}? {Try} {ARC}, the {AI2} {Reasoning} {Challenge}, March 2018.
\newblock URL \url{http://arxiv.org/abs/1803.05457}.
\newblock arXiv:1803.05457 [cs].

\bibitem[Paperno et~al.(2016)Paperno, Kruszewski, Lazaridou, Pham, Bernardi, Pezzelle, Baroni, Boleda, and Fern{\'a}ndez]{paperno_lambada_2016}
Denis Paperno, Germ{\'a}n Kruszewski, Angeliki Lazaridou, Ngoc-Quan Pham, Raffaella Bernardi, Sandro Pezzelle, Marco Baroni, Gemma Boleda, and Raquel Fern{\'a}ndez.
\newblock The lambada dataset: Word prediction requiring a broad discourse context.
\newblock In \emph{Proceedings of the 54th Annual Meeting of the Association for Computational Linguistics (Volume 1: Long Papers)}, pages 1525--1534, 2016.

\bibitem[Hendrycks et~al.(2020)Hendrycks, Burns, Basart, Zou, Mazeika, Song, and Steinhardt]{hendrycks2020measuring}
Dan Hendrycks, Collin Burns, Steven Basart, Andy Zou, Mantas Mazeika, Dawn Song, and Jacob Steinhardt.
\newblock Measuring massive multitask language understanding.
\newblock \emph{arXiv preprint arXiv:2009.03300}, 2020.

\bibitem[Sutawika et~al.(2025)Sutawika, Schoelkopf, Gao, Abbasi, Biderman, Tow, Lovering, Phang, Thite, Wang, et~al.]{sutawika2025eleutherai}
Lintang Sutawika, Hailey Schoelkopf, Leo Gao, Baber Abbasi, Stella Biderman, Jonathan Tow, Charles Lovering, Jason Phang, Anish Thite, Thomas Wang, et~al.
\newblock Eleutherai/lm-evaluation-harness: v0. 4.9.
\newblock \emph{Zenodo}, 2025.

\bibitem[Huang and Wei(2024)]{huang2024mixture}
Shaohan Huang and Furu Wei.
\newblock Mixture of lora experts.
\newblock In \emph{ICLR 2024}, April 2024.

\bibitem[Feng et~al.(2024)Feng, Hao, Zhang, Han, and Wang]{feng2024mixture}
Wenfeng Feng, Chuzhan Hao, Yuewei Zhang, Yu~Han, and Hao Wang.
\newblock Mixture-of-loras: An efficient multitask tuning method for large language models.
\newblock In \emph{Proceedings of the 2024 Joint International Conference on Computational Linguistics, Language Resources and Evaluation (LREC-COLING 2024)}, pages 11371--11380, 2024.

\bibitem[Hao et~al.(2024)Hao, Sukhbaatar, Su, Li, Hu, Weston, and Tian]{hao2024trainingllms}
Shibo Hao, Sainbayar Sukhbaatar, DiJia Su, Xian Li, Zhiting Hu, Jason Weston, and Yuandong Tian.
\newblock Training large language models to reason in a continuous latent space, 2024.
\newblock URL \url{https://arxiv.org/abs/2412.06769}.

\bibitem[Shen et~al.(2025)Shen, Yan, Zhang, Hu, Du, and He]{shen2025codi}
Zhenyi Shen, Hanqi Yan, Linhai Zhang, Zhanghao Hu, Yali Du, and Yulan He.
\newblock Codi: Compressing chain-of-thought into continuous space via self-distillation.
\newblock In \emph{Proceedings of the 2025 Conference on Empirical Methods in Natural Language Processing}, pages 677--693, 2025.

\bibitem[Wu et~al.(2025{\natexlab{b}})Wu, Teng, and Tu]{wu2025parallel}
Haoyi Wu, Zhihao Teng, and Kewei Tu.
\newblock Parallel continuous chain-of-thought with jacobi iteration.
\newblock In \emph{Proceedings of the 2025 Conference on Empirical Methods in Natural Language Processing}, pages 914--926, 2025{\natexlab{b}}.

\bibitem[Kuzina et~al.(2026)Kuzina, Pi{\'o}ro, Whatmough, and Bejnordi]{kuzina2026kava}
Anna Kuzina, Maciej Pi{\'o}ro, Paul~N. Whatmough, and Babak~Ehteshami Bejnordi.
\newblock Kava: Latent reasoning via compressed {KV}-cache distillation.
\newblock In \emph{The Fourteenth International Conference on Learning Representations}, 2026.
\newblock URL \url{https://openreview.net/forum?id=ePrhcLbtGv}.

\bibitem[Wang et~al.(2025{\natexlab{d}})Wang, Wang, Zhu, and Liu]{wang2025system}
Xiaoqiang Wang, Suyuchen Wang, Yun Zhu, and Bang Liu.
\newblock System-1.5 reasoning: Traversal in language and latent spaces with dynamic shortcuts.
\newblock \emph{arXiv preprint arXiv:2505.18962}, 2025{\natexlab{d}}.

\bibitem[Massoli et~al.(2026)Massoli, Kuzmin, and Behboodi]{Massoli2026-eu}
Fabio~Valerio Massoli, Andrey Kuzmin, and Arash Behboodi.
\newblock Reasoning as compression: Unifying budget forcing via the conditional information bottleneck.
\newblock \emph{arXiv preprint arXiv:2603.08462}, 9~March 2026.

\bibitem[Cesa et~al.(2026)Cesa, Hehn, Torres-Camps, Casellas, Ros-Giralt, Behboodi, and Orekondy]{cesa2026lanerope}
Gabriele Cesa, Thomas Hehn, Aleix Torres-Camps, {\`A}lex~Batlle Casellas, Jordi Ros-Giralt, Arash Behboodi, and Tribhuvanesh Orekondy.
\newblock Lanero{PE}: Positional encoding for collaborative parallel reasoning and generation.
\newblock In \emph{Workshop on Multi-Agent Learning and Its Opportunities in the Era of Generative AI}, 2026.
\newblock URL \url{https://openreview.net/forum?id=6WAuvwZjmw}.

\bibitem[Tseng et~al.(2024)Tseng, Chee, Sun, Kuleshov, and De~Sa]{tseng2024quip}
Albert Tseng, Jerry Chee, Qingyao Sun, Volodymyr Kuleshov, and Christopher De~Sa.
\newblock Quip\#: Even better llm quantization with hadamard incoherence and lattice codebooks.
\newblock \emph{Proceedings of machine learning research}, 235:\penalty0 48630, 2024.

\end{thebibliography}

\newpage
\appendix

\section{Benchmark Description}\label{App:benchmark}

To comprehensively assess the reasoning capabilities of our fine-tuned models, we leverage a diverse suite of benchmarks spanning the mathematics, science, and coding domains.
\begin{itemize}
        \item \textbf{AIME 24/25} \cite{aime} consists of 30 highly challenging mathematics competition problems from the 2024 and 2025 American Invitational Mathematics Examination. The questions cover a range of topics including algebra, geometry, and number theory, and are aimed for high-school students. The problems require multi-step reasoning, and the answers are integers between 0 and 999.

        \item \textbf{MATH500} \cite{hendrycks2021measuring} is a  benchmark consisting of 500 mathematical questions spanning different topics including algebra, geometry, number theory, precalculus, and probability. The problems require multi-step solutions, and answers may include LaTeX-formatted expressions.

        \item \textbf{GPQA Diamond} \cite{rein2024gpqa} consists of 198 science PhD-level questions from physics, chemistry, and biology. All questions are presented in a multiple-choice format.

        \item \textbf{AMC23} \cite{amc23} contains 40 problems from the 2023 American Mathematics Competition with integer answers.

        \item \textbf{LiveCodeBench} \cite{livecodebench} is a continuously updated (hence "live") coding benchmark. In every release, a certain number of coding problems is sourced from competitive programming platforms (e.g. CodeForces, LeetCode), and each problem is used to build four coding "scenarios": Code Generation, Code Repair, Test Output Prediction, and Code Execution. In this work we use the v2 release, comprising 511 problems, and confine ourselves to the Code Generation scenario.

        \item \textbf{HumanEval and HumanEval+} \cite{humaneval,evalplus} are the original and improved versions of the HumanEval benchmark, comprising 164 problems in which a model, based on a Python function's signature and docstring, must generate its body. The resulting function is then verified with several unit tests. \textbf{HumanEval+}~\cite{evalplus} improves the original benchmark by increasing the number unit tests for verification by 80 times.

        \item \textbf{MBPP and MBPP+} \cite{mbpp,evalplus} are the original and improved versions of the Most Basic Python Programs benchmark, consisting of 1000 basic Python programming tasks sourced from human coders. Each task involves writing a simple Python function based on natural language requirements and three unit tests it must pass. The \textbf{MBPP+} enhancement selects a subset of 378 task and increases the number of unit tests by 35 times.

\end{itemize}

\section{LoRA ablation study}\label{sec:lora_ablation}

This appendix provides the complete, detailed results of the parameter-efficient fine-tuning (PEFT) ablation study introduced in Section \Cref{sec:ablations}, in Tables~\ref{tab:ablations_qwen3b} and~\ref{tab:qwen7_ablations}. The study explores the impact of varying learning rates, batch sizes, and LoRA adapter ranks on the reasoning capabilities of both the 3B and 7B model backbones. The models were trained on a 50,000-entry subset of the OpenThoughts3 (OT3) dataset for one epoch and evaluated across core mathematical and scientific benchmarks (AIME24, AIME25, MATH500, GPQA, and AMC23).

\begin{table}[h!]
\centering
\caption{Ablation results for Qwen2.5-3B-Instruct. LR denotes learning rate and BS stands for batch size. In each learning rate subgroup, the best performance is marked in bold.}
\vskip 0.15in

\label{tab:ablations_qwen3b}
\begin{tabular}{ccc|ccccc|c}
\hline
\textbf{LR} & \textbf{BS} & \textbf{Rank} & \textbf{AIME24} & \textbf{AIME25} & \textbf{MATH500} & \textbf{GPQA} & \textbf{AMC23} & \textbf{Avg} \\
\hline
0.0001 & 32 & 32 & 0.05 & 0.01 & 0.55 & 0.25 & 0.24 & 0.220 \\
0.0001 & 32 & 64 & 0.03 & 0.02 & 0.58 & 0.29 & 0.28 & 0.240 \\
0.0001 & 32 & 128 & \textbf{0.08} & 0.03 & 0.56 & 0.28 & 0.30 & 0.250 \\
0.0001 & 32 & 256 & 0.03 & 0.04 & 0.57 & 0.28 & 0.28 & 0.240 \\
0.0001 & 64 & 32 & 0.05 & 0.01 & 0.54 & 0.27 & 0.28 & 0.230 \\
0.0001 & 64 & 64 & 0.03 & 0.03 & 0.57 & 0.29 & 0.29 & 0.242 \\
0.0001 & 64 & 128 & 0.02 & 0.03 & 0.58 & 0.27 & 0.30 & 0.240 \\
0.0001 & 64 & 256 & 0.04 & \textbf{0.06} & 0.59 & \textbf{0.30} & \textbf{0.34} & \textbf{0.266} \\
0.0001 & 128 & 32 & 0.04 & 0.01 & \textbf{0.61} & 0.26 & 0.29 & 0.242 \\
0.0001 & 128 & 64 & 0.05 & 0.00 & 0.59 & 0.29 & 0.27 & 0.240 \\
0.0001 & 128 & 128 & 0.03 & 0.01 & 0.56 & 0.27 & 0.25 & 0.224 \\
0.0001 & 128 & 256 & 0.03 & 0.05 & 0.58 & 0.27 & 0.25 & 0.236 \\
\hline
0.0002 & 32 & 32 & \textbf{0.08} & 0.01 & 0.58 & 0.26 & 0.33 & 0.252 \\
0.0002 & 32 & 64 & 0.05 & 0.03 & 0.59 & \textbf{0.30} & 0.32 & 0.258 \\
0.0002 & 32 & 128 & 0.06 & 0.04 & 0.57 & 0.25 & 0.29 & 0.242 \\
0.0002 & 32 & 256 & 0.07 & 0.04 & \textbf{0.60} & 0.28 & \textbf{0.34} & 0.266 \\
0.0002 & 64 & 32 & 0.03 & 0.02 & 0.56 & 0.27 & 0.30 & 0.236 \\
0.0002 & 64 & 64 & 0.07 & 0.03 & 0.58 & 0.26 & 0.28 & 0.244 \\
0.0002 & 64 & 128 & 0.07 & \textbf{0.06} & 0.55 & 0.29 & 0.25 & 0.244 \\
0.0002 & 64 & 256 & 0.06 & 0.04 & 0.58 & 0.27 & 0.30 & 0.250 \\
0.0002 & 128 & 32 & 0.00 & 0.03 & 0.59 & 0.27 & 0.28 & 0.234 \\
0.0002 & 128 & 64 & 0.00 & 0.02 & 0.55 & 0.28 & 0.28 & 0.226 \\
0.0002 & 128 & 128 & 0.05 & 0.04 & 0.55 & 0.27 & 0.30 & 0.242 \\
0.0002 & 128 & 256 & 0.07 & 0.04 & 0.59 & 0.29 & 0.32 & \textbf{0.262} \\
\hline
0.0005 & 32 & 32 & 0.05 & 0.02 & 0.55 & \textbf{0.31} & 0.29 & 0.244 \\
0.0005 & 32 & 64 & 0.05 & 0.03 & 0.53 & 0.28 & 0.27 & 0.232 \\
0.0005 & 32 & 128 & \textbf{0.06} & 0.04 & 0.54 & 0.26 & 0.27 & 0.234 \\
0.0005 & 32 & 256 & 0.04 & 0.05 & 0.53 & 0.28 & 0.24 & 0.228 \\
0.0005 & 64 & 32 & \textbf{0.06} & 0.04 & 0.56 & 0.26 & 0.29 & 0.242 \\
0.0005 & 64 & 64 & 0.05 & 0.05 & 0.57 & 0.28 & 0.30 & 0.250 \\
0.0005 & 64 & 128 & \textbf{0.06} & 0.04 & 0.57 & 0.29 & 0.27 & 0.246 \\
0.0005 & 64 & 256 & 0.03 & 0.05 & 0.54 & 0.28 & 0.24 & 0.228 \\
0.0005 & 128 & 32 & \textbf{0.06} & 0.03 & 0.57 & 0.28 & 0.26 & 0.240 \\
0.0005 & 128 & 64 & 0.03 & 0.00 & 0.57 & 0.28 & 0.27 & 0.230 \\
0.0005 & 128 & 128 & \textbf{0.06} & 0.05 & 0.56 & 0.27 & \textbf{0.35} & \textbf{0.258} \\
0.0005 & 128 & 256 & \textbf{0.06} & \textbf{0.06} & \textbf{0.58} & 0.29 & 0.25 & 0.248 \\
\hline
\end{tabular}
\end{table}

\newpage
\begin{table}[h!]
\centering
\caption{Ablation results for Qwen2.5-7B-Instruct. LR denotes learning rate and BS stands for batch size. In each learning rate subgroup, the best performance is marked in bold.}
\vskip 0.15in

\label{tab:qwen7_ablations}
\begin{tabular}{lcc|ccccc|c}
\hline
\textbf{LR} & \textbf{BS} & \textbf{Rank} & \textbf{AIME24} & \textbf{AIME25} & \textbf{MATH500} & \textbf{GPQA} & \textbf{AMC23} & \textbf{Avg} \\
\hline
0.0001 & 32 & 32 & 0.18 & 0.12 & \textbf{0.79} & 0.36 & 0.50 & 0.390 \\
0.0001 & 32 & 64 & 0.13 & 0.16 & 0.78 & 0.36 & 0.51 & 0.388 \\
0.0001 & 32 & 128 & 0.17 & 0.11 & 0.78 & 0.37 & 0.55 & 0.396 \\
0.0001 & 32 & 256 & 0.14 & 0.14 & \textbf{0.79} & 0.35 & 0.54 & 0.392 \\
0.0001 & 64 & 32 & 0.14 & 0.12 & 0.77 & 0.35 & 0.54 & 0.384 \\
0.0001 & 64 & 64 & 0.16 & \textbf{0.17} & 0.76 & 0.38 & \textbf{0.56} & \textbf{0.406} \\
0.0001 & 64 & 128 & 0.15 & 0.15 & 0.78 & \textbf{0.39} & 0.55 & 0.404 \\
0.0001 & 64 & 256 & 0.15 & 0.15 & 0.76 & 0.38 & 0.53 & 0.394 \\
0.0001 & 128 & 32 & 0.12 & 0.12 & 0.77 & 0.37 & 0.55 & 0.386 \\
0.0001 & 128 & 64 & 0.18 & 0.13 & 0.78 & 0.35 & 0.51 & 0.390 \\
0.0001 & 128 & 128 & \textbf{0.19} & 0.11 & 0.78 & 0.36 & 0.52 & 0.392 \\
0.0001 & 128 & 256 & 0.18 & 0.11 & 0.76 & 0.37 & 0.53 & 0.390 \\
\hline
0.0002 & 32 & 32 & 0.12 & 0.14 & 0.77 & 0.36 & 0.54 & 0.386 \\
0.0002 & 32 & 64 & 0.14 & 0.16 & 0.79 & 0.34 & 0.51 & 0.388 \\
0.0002 & 32 & 128 & 0.18 & \textbf{0.18} & 0.78 & 0.35 & 0.54 & 0.406 \\
0.0002 & 32 & 256 & 0.18 & 0.17 & 0.77 & 0.35 & 0.56 & 0.406 \\
0.0002 & 64 & 32 & \textbf{0.21} & 0.11 & 0.79 & 0.36 & 0.52 & 0.398 \\
0.0002 & 64 & 64 & 0.16 & 0.16 & 0.77 & 0.37 & \textbf{0.58} & 0.408 \\
0.0002 & 64 & 128 & 0.20 & 0.14 & \textbf{0.80} & \textbf{0.38} & 0.52 & 0.408 \\
0.0002 & 64 & 256 & 0.14 & 0.17 & 0.76 & 0.33 & 0.52 & 0.384 \\
0.0002 & 128 & 32 & 0.15 & 0.11 & 0.77 & 0.35 & 0.57 & 0.390 \\
0.0002 & 128 & 64 & 0.18 & 0.12 & 0.79 & 0.33 & 0.55 & 0.394 \\
0.0002 & 128 & 128 & 0.18 & 0.15 & 0.77 & 0.35 & 0.57 & 0.404 \\
0.0002 & 128 & 256 & 0.20 & 0.17 & 0.79 & \textbf{0.38} & 0.54 & \textbf{0.416} \\
\hline
0.0005 & 32 & 32 & 0.17 & 0.15 & 0.76 & 0.38 & 0.53 & 0.398 \\
0.0005 & 32 & 64 & 0.11 & 0.09 & 0.67 & \textbf{0.40} & 0.50 & 0.354 \\
0.0005 & 32 & 128 & 0.16 & 0.11 & 0.76 & 0.37 & 0.47 & 0.374 \\
0.0005 & 32 & 256 & 0.00 & 0.00 & 0.00 & 0.24 & 0.00 & 0.048 \\
0.0005 & 64 & 32 & 0.17 & 0.14 & 0.76 & 0.38 & 0.54 & 0.398 \\
0.0005 & 64 & 64 & 0.15 & 0.14 & 0.77 & 0.34 & 0.50 & 0.380 \\
0.0005 & 64 & 128 & 0.22 & 0.12 & \textbf{0.79} & 0.37 & \textbf{0.59} &\textbf{0.418} \\
0.0005 & 64 & 256 & 0.00 & 0.00 & 0.01 & 0.27 & 0.02 & 0.060 \\
0.0005 & 128 & 32 & 0.19 & 0.13 & \textbf{0.79} & 0.34 & 0.55 & 0.400 \\
0.0005 & 128 & 64 & 0.17 & 0.12 & 0.77 & 0.38 & 0.55 & 0.398 \\
0.0005 & 128 & 128 & \textbf{0.23} & \textbf{0.17} & 0.78 & 0.37 & 0.53 & 0.416 \\
0.0005 & 128 & 256 & 0.21  & \textbf{0.17}  & 0.76  &  0.36 & 0.49  & 0.398 \\
\hline
\end{tabular}
\end{table}

\clearpage
\section{Budget Forcing Details} \label{app:budget_forcing}

\autoref{tab:hyperparams} reports the list of hyperparameters we used in the budget forcing RL training.

\begin{table}[!ht]
\centering
\caption{\textbf{GRPO Training Hyperparameters.} All experiments share these settings unless otherwise noted.}\label{tab:hyperparams}
\begin{tabular}{l|c}
\toprule
\textbf{Hyperparameter} & \textbf{Value} \\ 
\midrule
Optimizer & AdamW \\
Learning Rate & $2 \times 10^{-5}$ \\
LR Scheduler & Cosine \\
Warmup Ratio & 0.05 \\
Batch Size (Global) & 256 \\
Generations per Prompt ($G$) & 8 \\
Temperature & 0.8 \\
Max Completion Length & 6144  \\
Max Gradient Norm & 1.0 \\
KL Penalty Coefficient ($\beta_{\text{KL}}$) & $\{10^{-3}, 10^{-4}\}$ \\
Training Steps & 200 \\ 
\bottomrule
\end{tabular}
\end{table}

\begin{figure}[!ht]
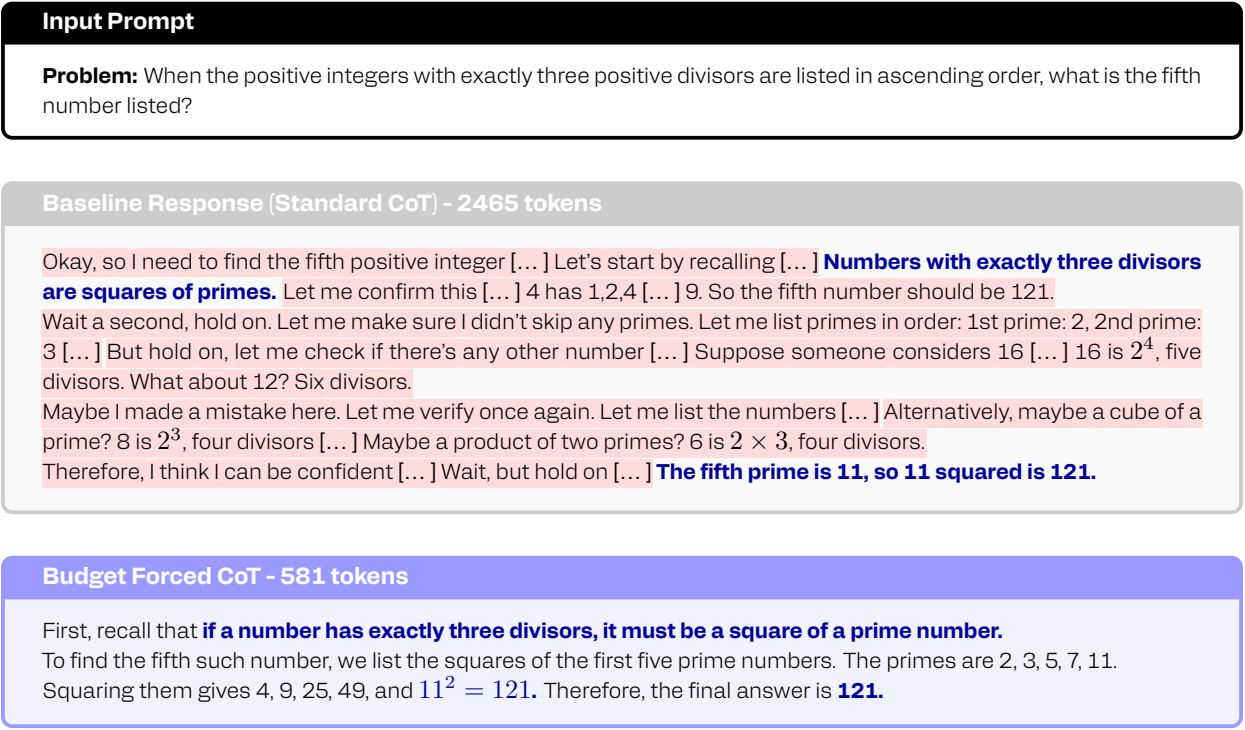

    \centering
    
    \begin{tcolorbox}[colback=white, colframe=black, title=\textbf{Input Prompt}]
    \small
    \textbf{Problem:} When the positive integers with exactly three positive divisors are listed in ascending order, what is the fifth number listed?
    \end{tcolorbox}
    \vspace{0.1cm}
    
    \begin{tcolorbox}[colback=gray!5, colframe=gray!40, title=\textbf{Baseline Response (Standard CoT) - 2465 tokens}]
    \small 
    \bloat{Okay, so I need to find the fifth positive integer \textbf{[$\dots$]} Let's start by recalling \textbf{[$\dots$]}} \bridge{Numbers with exactly three divisors are squares of primes.} 
    \bloat{Let me confirm this \textbf{[$\dots$]} 4 has 1,2,4 \textbf{[$\dots$]} 9. So the fifth number should be 121.}
    
    \bloat{Wait a second, hold on. Let me make sure I didn't skip any primes. Let me list primes in order: 1st prime: 2, 2nd prime: 3 \textbf{[$\dots$]}} \bloat{But hold on, let me check if there's any other number \textbf{[$\dots$]} Suppose someone considers 16 \textbf{[$\dots$]} 16 is $2^4$, five divisors. What about 12? Six divisors.}
    
    \bloat{Maybe I made a mistake here. Let me verify once again. Let me list the numbers \textbf{[$\dots$]}} \bloat{Alternatively, maybe a cube of a prime? 8 is $2^3$, four divisors \textbf{[$\dots$]} Maybe a product of two primes? 6 is $2\times3$, four divisors.}
    
    \bloat{Therefore, I think I can be confident \textbf{[$\dots$]} Wait, but hold on \textbf{[$\dots$]}}
    \bridge{The fifth prime is 11, so 11 squared is 121.}
    \end{tcolorbox}
    \vspace{0.1cm} 
    
    \begin{tcolorbox}[colback=blue!5, colframe=blue!40, title=\textbf{Budget Forced CoT - 581 tokens}]
    \small
    First, recall that \bridge{if a number has exactly three divisors, it must be a square of a prime number.}
    
    To find the fifth such number, we list the squares of the first five prime numbers.
    The primes are 2, 3, 5, 7, 11.
    
    Squaring them gives 4, 9, 25, 49, and \bridge{$11^2 = 121$.}
    Therefore, the final answer is \bridge{121.}
    \end{tcolorbox}

    \caption{\textbf{Qualitative comparison on number theory reasoning.} Text highlighted in red denotes redundant verification and verbal parsing, while bold text identifies essential reasoning steps. We use ``\textbf{[$\dots$]}'' as a placeholder for brevity. \textbf{Top:} Prompt. \textbf{Middle:} The Baseline trace correctly identifies the property ($p^2$) early on but falls into extensive, redundant verification loops checking composite numbers and re-listing primes (highlighted in red). \textbf{Bottom:} The Budget Forced trace directly applies the prime-square property and computes the result without unnecessary hesitation or syntactic noise.}
    \label{fig:cot_example_1}
\end{figure}

We report in Figures \ref{fig:cot_example_1} and \ref{fig:cot_example_3} additional qualitative comparisons between the reasoning traces of the unconstrained baseline against our budget-forced model. 

\begin{figure*}[!ht]
    \centering
    
    \begin{tcolorbox}[colback=white, colframe=black, title=\textbf{Input Prompt}]
    \small
    \textbf{Problem:} What is the 2003rd term of the sequence of odd numbers 1, 3, 5, 7, $\dots$?
    \end{tcolorbox}
    \vspace{0.1cm}
    
    \begin{tcolorbox}[colback=gray!5, colframe=gray!40, title=\textbf{Baseline Response (Standard CoT) - 3434 tokens}]
    \small 
    \bloat{Okay, so I need to find the 2003rd term \textbf{[$\dots$]} Let me think. First, I remember that the sequence is \textbf{[$\dots$]}} \bridge{The nth term is $2n - 1$.}
    
    \bloat{Wait, let me check that. If n=1, then 1 \textbf{[$\dots$]} n=2 gives 3 \textbf{[$\dots$]} So that formula checks out.} 
    \bloat{Alternatively, maybe I can think of the sequence as starting at 1 and adding 2 \textbf{[$\dots$]} arithmetic sequence formula $1 + (n-1)2$ \textbf{[$\dots$]} expands to $2n-1$.}
    
    \bloat{Maybe someone might think the nth term is $n^2$ \textbf{[$\dots$]} or $2n$ \textbf{[$\dots$]} let me check term 5 \textbf{[$\dots$]}. Correct.}
    \bloat{Therefore, no matter how I approach it \textbf{[$\dots$]} Wait, but just to be thorough, let me compute $2 \times 2003$ first \textbf{[$\dots$]}}
    
    \bridge{Substituting $n=2003$ gives $2(2003) - 1 = 4005$.}
    \end{tcolorbox}
    \vspace{0.1cm} 
    
    \begin{tcolorbox}[colback=blue!5, colframe=blue!40, title=\textbf{Budget Forced CoT - 829 tokens}]
    \small
    First, observe that the sequence of odd numbers is given by the formula:
    \bridge{$a_n = 2n - 1$.}
    
    To find the 2003rd term, we substitute $n = 2003$ into the formula:
    \bridge{$a_{2003} = 2(2003) - 1 = 4006 - 1 = 4005$.}
    
    Therefore, the final answer is \bridge{4005.}
    \end{tcolorbox}

    \caption{\textbf{Qualitative comparison on pattern recognition.} \textbf{Middle:} The Baseline trace correctly identifies the formula $2n-1$ initially but spends nearly 1000 tokens validating it against alternative arithmetic formulas ($a+(n-1)d$) and hypothetical user errors (confusing term number with value, testing $n^2$, etc.). \textbf{Bottom:} The Budget Forced trace directly retrieves the formula and computes the specific term requested.}
    \label{fig:cot_example_3}
\end{figure*}

\end{document}